\documentclass[accepted]{uai2026}

\usepackage[american]{babel}

\usepackage{natbib}
\usepackage{mathtools}
\usepackage{booktabs}
\usepackage{tikz}

\usepackage{amsthm}

\usepackage{amsmath,amsfonts,bm}

\def\eqref#1{equation~\ref{#1}}

\def\1{\bm{1}}

\DeclareMathAlphabet{\mathsfit}{\encodingdefault}{\sfdefault}{m}{sl}
\SetMathAlphabet{\mathsfit}{bold}{\encodingdefault}{\sfdefault}{bx}{n}

\def\gB{{\mathcal{B}}}

\def\gN{{\mathcal{N}}}

\def\sV{{\mathbb{V}}}

\newcommand{\E}{\mathbb{E}}
\newcommand{\Ls}{\mathcal{L}}

\usepackage{graphicx}
\usepackage{subfig}
\usepackage{float}
\usepackage{caption}
\usepackage{algorithm}
\usepackage{algpseudocode}
\usepackage[noabbrev,nameinlink,capitalise]{cleveref}
\hypersetup{hidelinks}

\newcommand{\ggd}{\mathrm{GGD}}
\newcommand{\mse}{\mathrm{MSE}}

\theoremstyle{plain}
\newtheorem{thm}{Theorem}

\newtheorem{pro}{Proposition}
\theoremstyle{definition}

\theoremstyle{remark}
\newtheorem{rem}{Remark}

\crefname{lem}{Lemma}{Lemmas}
\crefname{pro}{Proposition}{Propositions}
\crefname{dfn}{Definition}{Definitions}
\crefname{rem}{Remark}{Remarks} 
\crefname{thm}{Theorem}{Theorems}
\crefname{equation}{}{}

\title{Generalized Gaussian Temporal Difference Error\\for Uncertainty-aware Reinforcement Learning}

\author[1]{\href{mailto:se-yeon.kim@kt.com}{Seyeon Kim}{}\thanks{These authors contributed equally to this work.}\thanks{Work done at Qraft Technologies}}
\author[2]{\href{mailto:joonhun.lee@qraftec.com}{Joonhun Lee}{}$^{*}$}
\author[2]{\href{mailto:namhoon.cho@qraftec.com}{Namhoon Cho}}
\author[2]{\href{mailto:sungjun.han@qraftec.com}{Sungjun Han}}
\author[2]{\href{mailto:wooseop.hwang@qraftec.com}{Wooseop Hwang}}

\affil[1]{
    Data Team\\
    KT\\
    Republic of Korea
}
\affil[2]{
    AI Research Team\\
    Qraft Technologies\\
    Republic of Korea
}

\begin{document}
\maketitle

\begin{abstract}
	Conventional uncertainty-aware temporal difference (TD) learning often models TD errors as zero-mean Gaussian. This assumption can miss the heavy-tailed and heteroscedastic residuals induced by bootstrapping and exploration. We introduce a state-conditioned shape head based on the \emph{Generalized Gaussian Distribution (GGD)} and use a numerically modified GGD loss as an online surrogate for nonstationary TD residuals. We distinguish two mathematical facts that are sometimes conflated: the exact GGD likelihood is normalized for every $\beta>0$, whereas its exponential density kernel is positive definite for $\beta\in(0,2]$. We use the learned shape to construct a simple monotone weighting heuristic and propose \emph{Batch Inverse Error Variance (BIEV)} regularization from the variance and sample excess kurtosis of ensemble TD errors. Across continuous and discrete control benchmarks, the shape-aware variants improve over Gaussian-based variants in several settings, although the gains are task- and regime-dependent.
\end{abstract}


\section{Introduction}\label{sec:intro}

Deep reinforcement learning (RL) has demonstrated promising potential across various real-world applications, e.g., finance~\citep{moody1998reinforcement,byun2023practical,sun2023reinforcement}, and autonomous driving~\citep{kahn2017uncertainty,emamifar2023uncertainty,hoel2023ensemble}.
Effectively quantifying uncertainty in high-dimensional RL environments enhances decision robustness and sample efficiency, particularly in unseen or ambiguous situations~\citep{lockwood2022review}.

Temporal difference (TD) learning, a fundamental component of many RL algorithms, facilitates value function estimation and policy derivation through iterative updates~\citep{sutton1988learning}.
Traditionally, TD updates assume a Gaussian error distribution, relying on $L_2$ loss for maximum likelihood estimation, or MLE.
However, TD errors are structurally non-Gaussian~\citep{flennerhag2020temporal,janz2019successor}: unlike supervised residuals defined against a fixed target, TD errors are recursively coupled with the evolving value function through bootstrapping~\citep{sutton1988learning}, inheriting non-stationarity and state-dependent heteroscedasticity.
This feedback loop is further amplified by exploration and off-policy updates, giving rise to infrequent but high-impact deviations even when the reward distribution is light-tailed~\citep{osband2017gaussian,agarwal2021deep}.
Beyond variance, error distributions are further characterized by their \emph{shape}~\citep{decarlo1997meaning,milton2017estimation}: kurtosis, an influential scale-independent moment, governs tail thickness independently of scale.
Yet, conventional deep RL methods focus solely on the scale parameter and overlook shape, conflating informative rare errors with irreducible noise.

To address this, we adopt the \emph{generalized Gaussian distribution (GGD)}, a flexible symmetric family encompassing Gaussian, Laplacian, and uniform distributions via a single shape parameter~\citep{box2011bayesian}.
The added shape output supplies a tail-sensitive signal beyond a Gaussian scale head, although fixing $\alpha=1$ deliberately couples shape and dispersion rather than disentangling them.
The implementation uses a GGD-inspired surrogate and normalizes shape weights across the critic ensemble for each transition.
\emph{Batch inverse error variance (BIEV)} separately uses the variance and sample excess kurtosis of ensemble TD errors; both components are evaluated as lightweight heuristics rather than calibrated uncertainty estimators.

\paragraph{Contributions.}\label{par:operationalization}
The GGD identities, stochastic-ordering result, and MSE-best variance estimator used below are established results from probability and statistics.
Our contribution is their RL-side operationalization: a state-conditioned $\beta$ head evaluated per transition, a simple monotone shape-weighting heuristic, and a BIEV term constructed from ensemble TD-error moments.
The implemented loss is a numerically modified online surrogate for bootstrapped residuals, not an offline maximum-likelihood consistency result.

The key contributions of our work are as follows:

\begin{enumerate}
	\item \textbf{Empirical investigations} (\cref{sssec:empirical}):
	      We conduct empirical analyses of TD error distributions, revealing substantial deviations from Gaussianity, particularly in terms of tailedness.
	      These findings underscore the limitations of conventional Gaussian assumptions.
	\item \textbf{Theoretical exploration} (\cref{sssec:theoretical}):
	      We separate the validity of the GGD likelihood for all $\beta>0$ from the positive-definite-kernel property for $\beta\in(0,2]$ and state how these established results relate to our surrogate.
	      The aggregate $\beta$ trajectories remain in the leptokurtic range in the reported experiments (\cref{fig:beta-head}).
	\item \textbf{Aleatoric uncertainty mitigation} (\cref{ssec:ggd}):
	      We investigate the implications of the distribution shape on estimation and mitigation of aleatoric uncertainty.
	      GGD error modeling provides a closed-form variance expression; for fixed $\alpha$ it decreases with $\beta$ over the leptokurtic interval used in our analysis.
	      We use the associated stochastic-ordering result as motivation for a monotone weighting heuristic and evaluate its direction by ablation.
	\item \textbf{Epistemic uncertainty mitigation} (\cref{ssec:biev}):
	      We introduce batch inverse error variance weighting, adapted from batch inverse variance~\citep{mai2022sample}, using the across-critic variance and sample excess kurtosis of TD errors.
	      We treat the resulting finite-ensemble estimator as a regularization heuristic rather than an exact uncertainty correction.
	\item \textbf{Experimental evaluations} (\cref{sec:exp}):
	      We evaluate the method with policy-gradient algorithms and report improvements over Gaussian-based baselines within the scope of the tested benchmarks.
\end{enumerate}

\paragraph{Scope.}\label{par:scope-section1}
The GGD-inspired loss transfers to value-based RL, but the shape-based risk weighting is designed for stochastic policy-gradient algorithms with separate actor networks; \cref{apdx:qlearn} evaluates this distinction.


\section{Background} \label{sec:bg}

We follow the standard Markov decision process (MDP) formulation~\citep{sutton2018reinforcement}, where an agent interacts with the environment via policy $\pi$, receives rewards, and updates value estimates through temporal difference (TD) learning.
Model-free deep RL algorithms~\citep{mnih2015human,schulman2017proximal,haarnoja2018soft} train neural networks to approximate the value function by minimizing the TD error $\delta_t = T_t - Q_t$, where $T_t = r_t + \gamma Q_{t+1}$ denotes the bootstrapped target.
Conventionally, the mean squared error (MSE) loss corresponds to a zero-mean Gaussian error model under maximum likelihood estimation (MLE).


\subsection{Uncertainty} \label{ssec:unc}

Uncertainty in neural networks decomposes into two sources: aleatoric and epistemic~\citep{der2009aleatory,kendall2017uncertainties,depeweg2018decomposition,valdenegro2022deeper}.
Aleatoric uncertainty stems from inherent environmental stochasticity and is fundamentally irreducible, whereas epistemic uncertainty arises from limited data and can be reduced through further exploration.

This distinction is crucial in RL: effective exploration targets regions of high epistemic uncertainty while avoiding areas dominated by aleatoric noise.
Moreover, quantifying aleatoric uncertainty enables risk-sensitive decision-making in stochastic environments~\citep{dabney2018distributional,vlastelica2021risk,seitzer2022pitfalls}.

\paragraph{Aleatoric and Epistemic Uncertainty.}
Variance networks $Q^\sigma$ model zero-mean heteroscedastic TD errors, $\delta_t \mid s_t,a_t\sim\gN(0,{Q^\sigma_t}^2)$, via the Gaussian negative log-likelihood (NLL), $\Ls_\text{GD-NLL} = \sum_t ( (\delta_t/Q^\sigma_t)^2 + \log{Q^\sigma_t}^2 )$~\citep{bishop1994mixture,nix1994estimating,mai2022sample}.
This per-sample likelihood captures \emph{aleatoric uncertainty} arising from irreducible environmental stochasticity.
Here $Q^\mu_t$ denotes a critic's mean value prediction at $(s_t,a_t)$, not the mean of $\delta_t$, which is zero under the Gaussian model; $Q^\sigma_t$ is the per-sample aleatoric scale output.
In contrast, $\sV[Q^\mu_t]$ below is the variance across the $K$ critics' mean predictions at the same $(s_t,a_t)$ and serves as an epistemic-disagreement proxy.
For \emph{epistemic uncertainty}, batch inverse variance (BIV) regularization scales error contributions inversely to the cross-critic ensemble variance $\sV[Q^\mu_t]$, ensuring noisy samples contribute less~\citep{mai2022sample}:
\begin{equation} \label{eq:mai-loss}
	\Ls = \sum_t \left( (\delta_t/{Q^\sigma_t})^2 + \log{{Q^\sigma_t}^2} + \lambda \frac{\omega^\text{BIV}_t}{\sum_\tau \omega^\text{BIV}_\tau} \delta_t^2 \right).
\end{equation}
Here, $\omega^\text{BIV}_t = \frac{1}{\gamma^2 \sV[Q^\mu_t] + \xi}$ denotes the BIV weight, with $\xi$ ensuring numerical stability.
The regularization temperature $\lambda$ controls the strength of epistemic regularization: the Gaussian NLL captures aleatoric uncertainty at the sample level via $Q^\sigma$, while BIV regularization mitigates epistemic uncertainty by down-weighting TD errors with high ensemble variance across critics.


\subsection{Tailedness} \label{ssec:tail}

Mainstream machine learning often prioritizes central tendencies, yet tail events can affect learning dynamics~\citep{david1979robust,gather1988maximum}.
This is particularly relevant for MLE under normality assumptions in variance network frameworks.

Under non-Gaussian sampling, normal-theory uncertainty calculations for variance and covariance estimators can be miscalibrated, and their sampling variability depends on fourth moments~\citep{yuan2005effect}.
Heavy tails therefore make small-sample variance estimates noisier even when the point estimator is unbiased or its finite-sample bias is known.
This motivates evaluating a kurtosis-aware weighting heuristic in~\cref{ssec:biev}, not claiming that kurtosis alone removes inverse-variance bias.

\begin{rem}[Kurtosis bias~\citep{burch2014estimating}] \label{rem:tail-variance}
	The sampling variance of a sample-variance estimator depends on the fourth central moment.
	Positive excess kurtosis can therefore make normal-theory confidence intervals too narrow even when the variance point estimator itself is not systematically downward biased.
\end{rem}

\paragraph{Gumbel Error Modeling.}
Recent work applies the Gumbel distribution to estimate $\max Q$ in Bellman updates~\citep{garg2022extreme,hui2023double}, leveraging the extreme value theorem~\citep{fisher1928limiting,mood1950introduction}.
However, TD residuals are not block maxima and are nonstationary; their tails drift with bootstrapping and policy improvement, and Gumbel-like behavior often fades during training~\citep{garg2022extreme}.
Moreover, Gumbel modeling targets one-sided tails and offers limited control over symmetric tailedness.

In contrast, our GGD-inspired approach uses a symmetric, shape-controlled surrogate for heavy-tailed and near-Gaussian regimes.
It can be integrated as a critic-loss modification without introducing an additional target network or rollout beyond those used by the underlying algorithm.

\paragraph{Student-$t$ Error Modeling.}
Another natural candidate for heavy-tailed regression is the Student-$t$ distribution, whose degrees-of-freedom parameter $\nu$ controls the transition from heavy tails, i.e., small $\nu$, to Gaussian, i.e., $\nu\to\infty$.
However, the Student-$t$ family is limited to the leptokurtic-to-Gaussian spectrum and cannot represent platykurtic, i.e., uniform-like, distributions.
The GGD spans leptokurtic exponential-power shapes through Gaussian at $\beta=2$ to a uniform limit as $\beta\to\infty$, whereas Student-$t$ provides polynomial tails.
Neither family strictly contains the other.
We nevertheless acknowledge Student-$t$ regression as an additional missing heavy-tailed error-modeling baseline; a direct empirical comparison is left to future work.


\subsection{Related Work} \label{ssec:related}

\paragraph{Distributional Reinforcement Learning.}
Distributional RL methods model the full return distribution via dedicated architectures, e.g., categorical atoms~\citep{bellemare2017distributional}, quantile regression~\citep{dabney2018distributional}, and implicit quantile functions~\citep{dabney2018implicit}.
Our approach is complementary: we model the \emph{error} distribution during TD learning with a lightweight head compatible with any actor-critic algorithm.
Whereas return-distributional methods use many categorical or quantile atoms, our surrogate uses a single scalar shape output for each one-step TD error.
The two views operate at different levels and can in principle be combined; we do not evaluate such stacking here.

\paragraph{Uncertainty Estimation in RL.}
Ensemble methods~\citep{lakshminarayanan2017simple,osband2016deep,lee2021sunrise} quantify epistemic uncertainty via disagreement, while variance networks~\citep{nix1994estimating,mai2022sample} capture aleatoric uncertainty with a heteroscedastic head.
We augment the variance-network paradigm with a GGD-inspired shape surrogate and use ensemble TD-error moments for BIEV regularization.

\paragraph{Heavy-Tailed Phenomena in Learning.}
Heavy tails arise in gradient noise~\citep{simsekli2019tail,gurbuzbalaban2021heavy}, reward distributions~\citep{cayci2024provably,zhuang2021no}, and TD errors~\citep{flennerhag2020temporal,janz2019successor}.
We examine a GGD-inspired parametric shape output whose aggregate trajectory evolves during training.


\begin{figure*}[t]
	\centering
	\subfloat[Ant-v4]{\includegraphics[width=.65\textwidth]{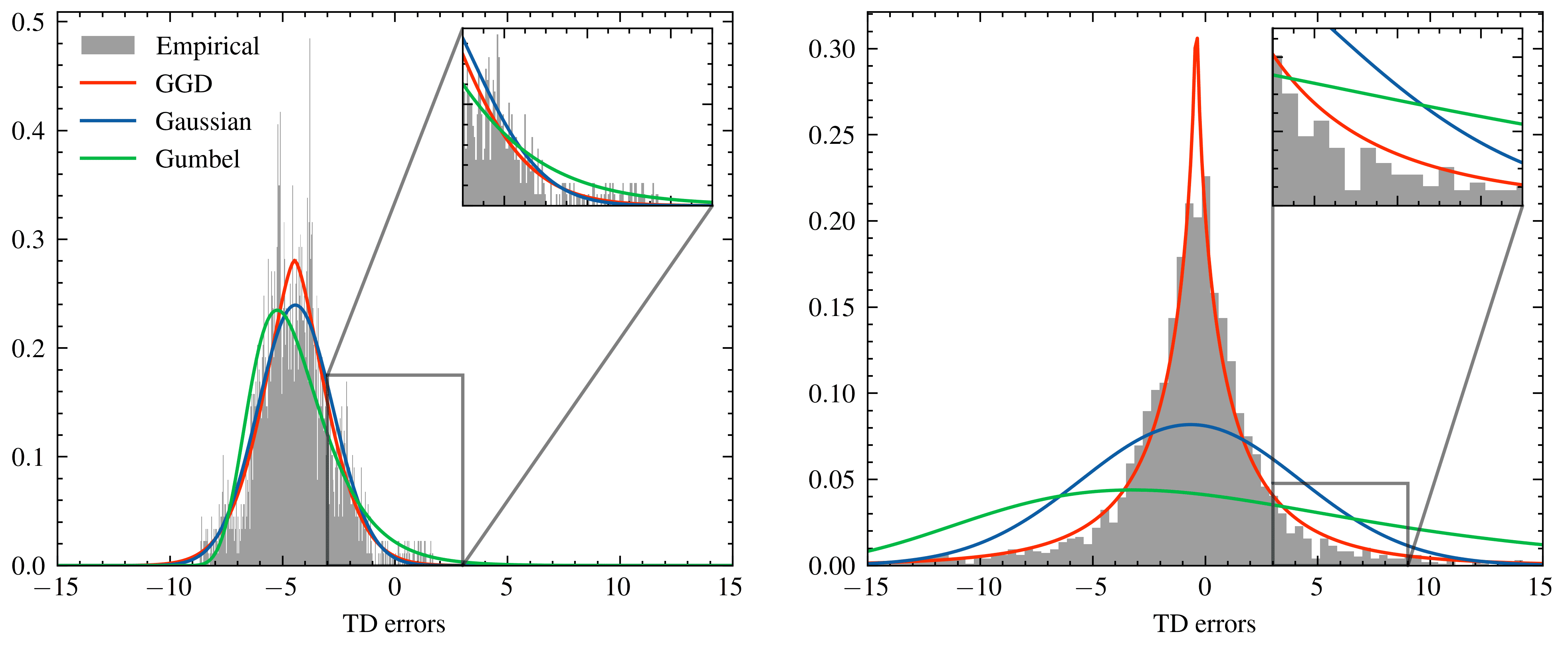}} \\
	\subfloat[Hopper-v4]{\includegraphics[width=.65\textwidth]{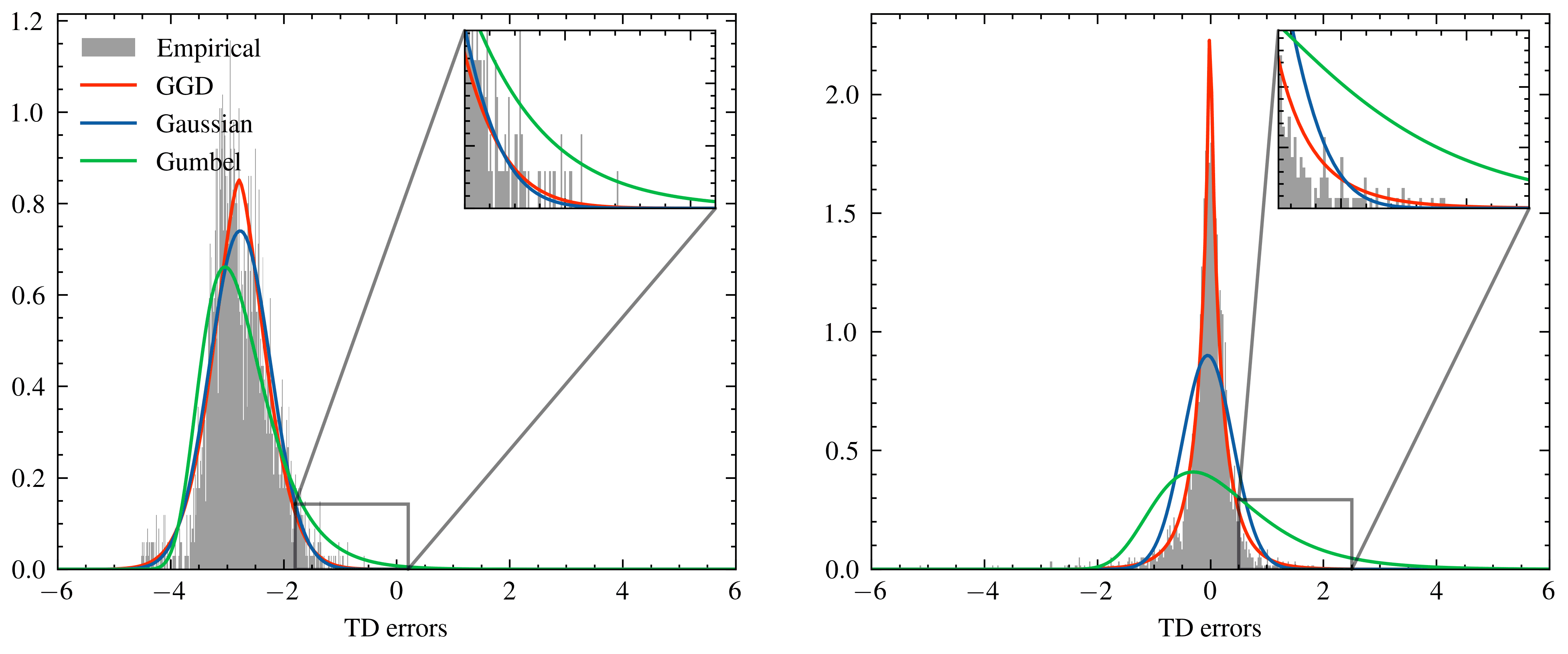}}
	\caption{
		TD error plots of Soft Actor-Critic (SAC) at initial and final evaluations (left to right) with fitted probability density functions (PDFs) using SciPy~\citep{virtanen2020scipy}.
		Additional plots for other environments and Proximal Policy Optimization (PPO) are in~\cref{apdx:ext:tde}.
		The histograms pool errors over many state--action pairs and therefore show marginal error distributions, not the per-transition conditional shape output used in training.
		The fitted curves are visualization overlays only; training does not fit a single global distribution to these histograms.
	} \label{fig:tde-sac}
\end{figure*}

\begin{figure*}[t]
	\centering
	\subfloat[Ant-v4]{\includegraphics[width=.31\textwidth]{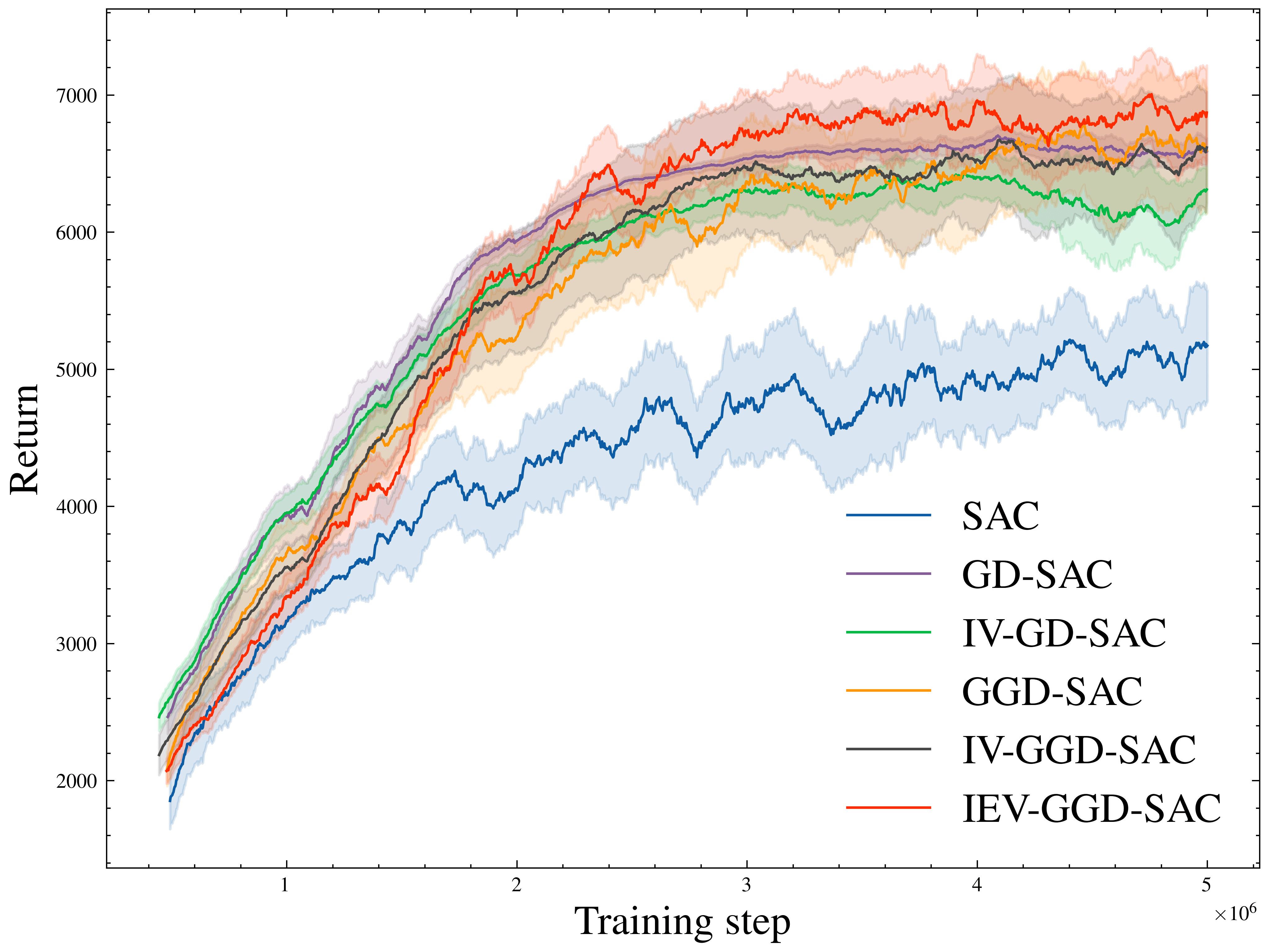}}
	\subfloat[Hopper-v4]{\includegraphics[width=.31\textwidth]{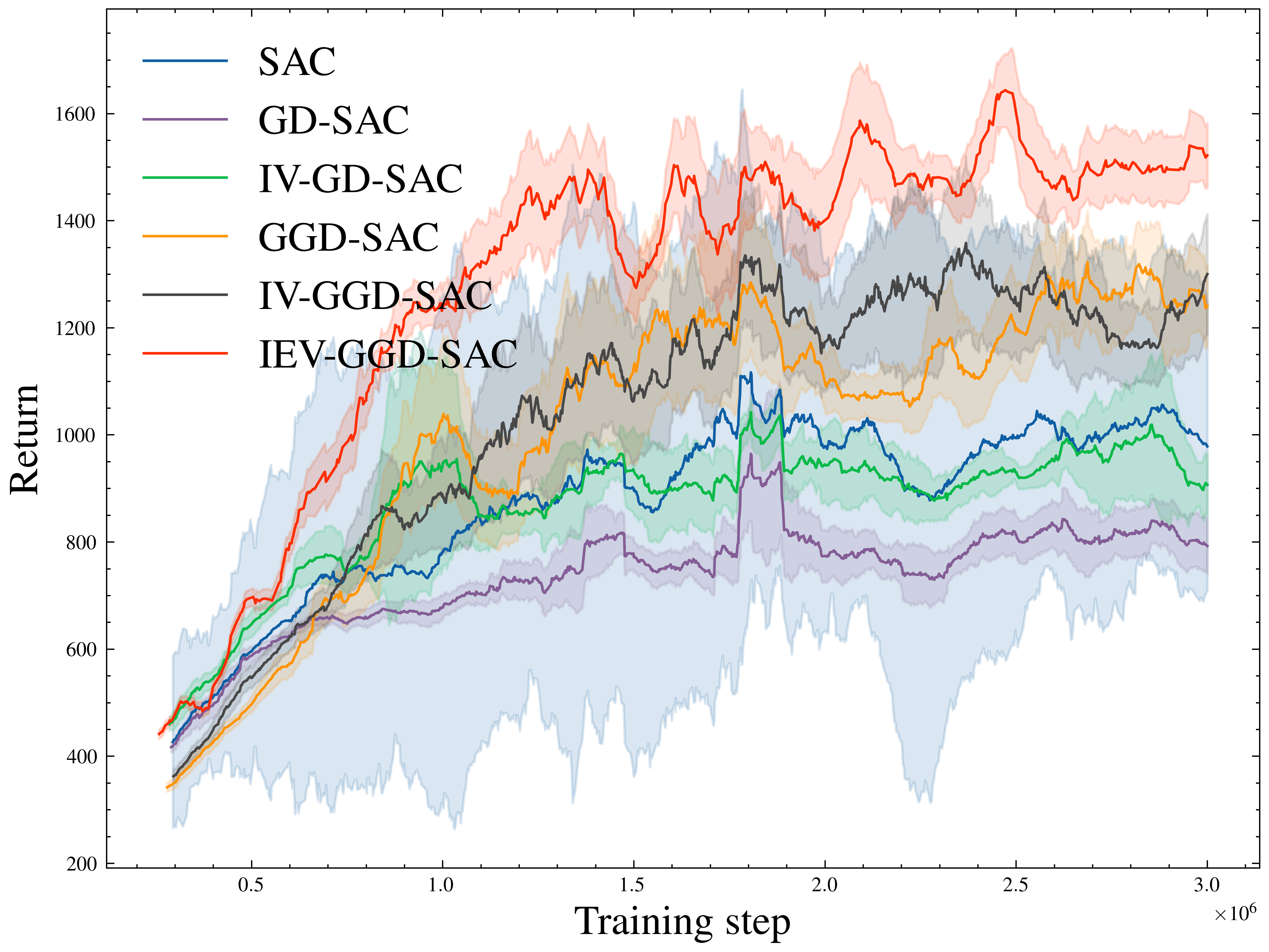}}
	\subfloat[Walker2D-v4]{\includegraphics[width=.31\textwidth]{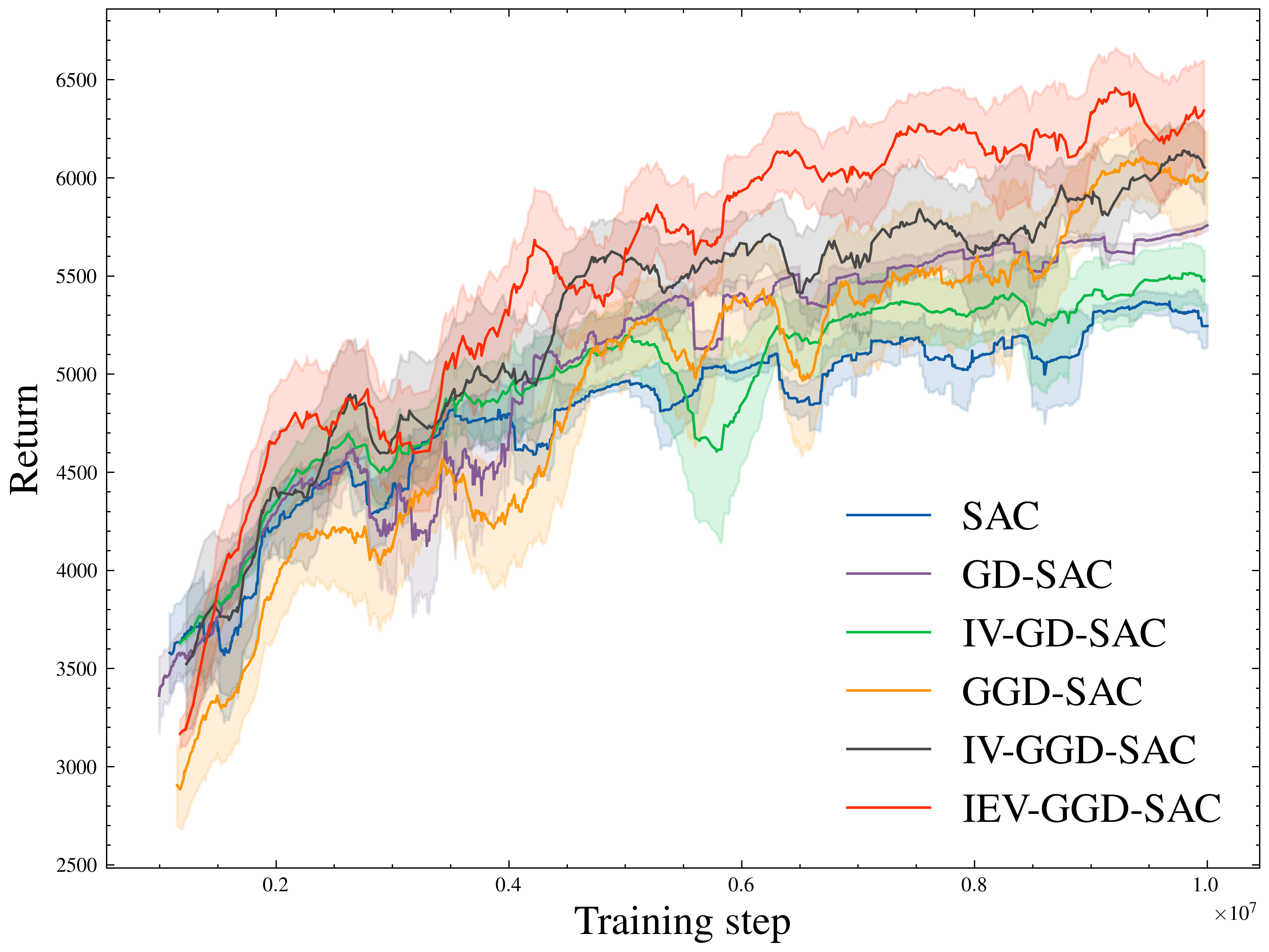}}
	\caption{
		Sample-efficiency curves of SAC on selected MuJoCo environments.
		Prefixes denote applied methods: `GD-' uses a Gaussian NLL and variance head, `GGD-' the GGD-inspired surrogate and shape head, `IV-' batch inverse variance, and `IEV-' BIEV regularization.
		Additional results appear in~\cref{apdx:ext:sac}.
	} \label{fig:sac-mujoco}
\end{figure*}

\begin{figure*}[t]
	\centering
	\subfloat[Ant-v4]{\includegraphics[width=.31\textwidth]{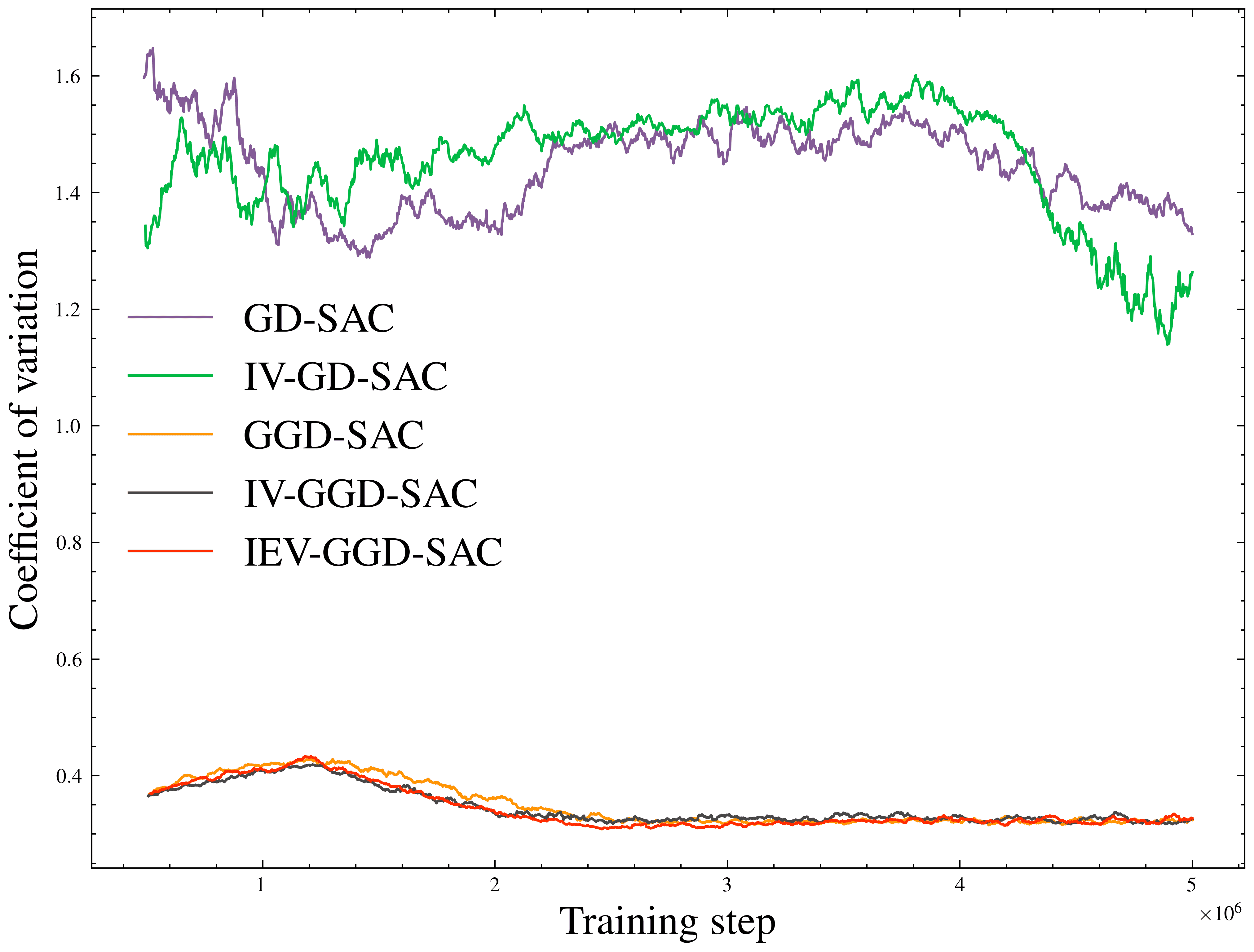}}
	\subfloat[Hopper-v4]{\includegraphics[width=.31\textwidth]{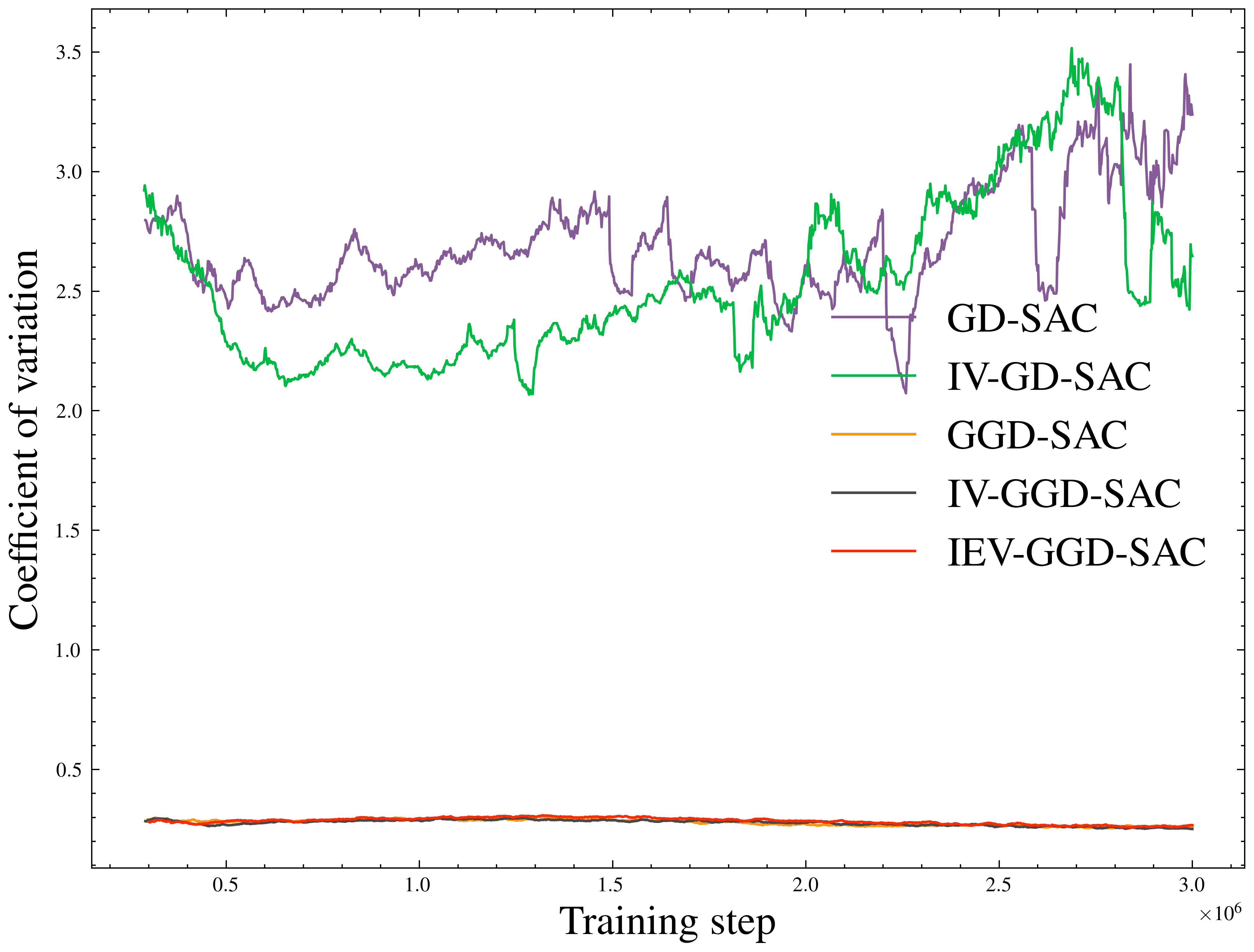}}
	\subfloat[Walker2D-v4]{\includegraphics[width=.31\textwidth]{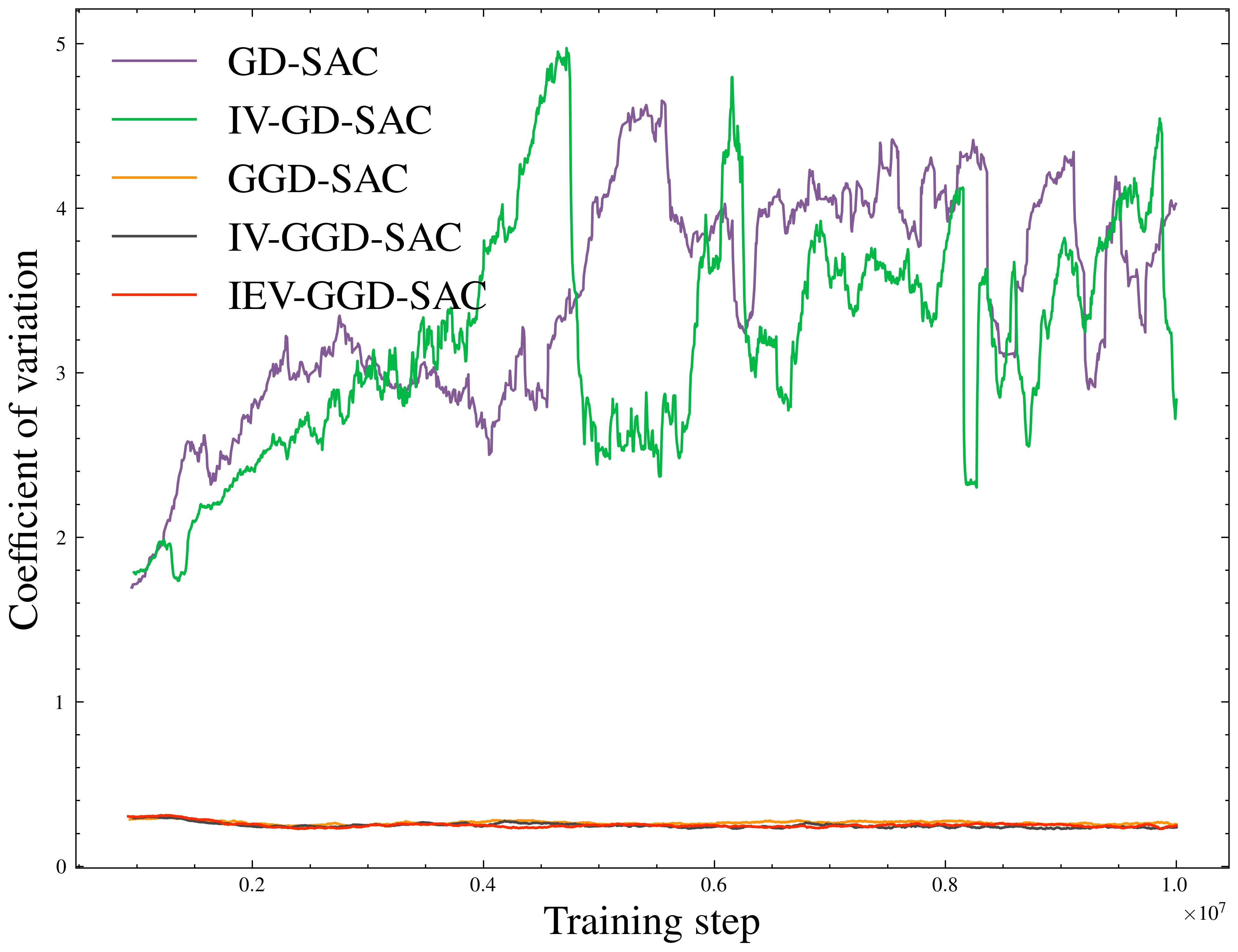}}
	\caption{
		Coefficients of variation of parameter estimates for SAC variants.
		Additional results appear in~\cref{apdx:ext:param}.
	} \label{fig:param}
\end{figure*}

\begin{figure*}[t]
	\centering
	\subfloat[CartPole-v1]{\includegraphics[width=.31\textwidth]{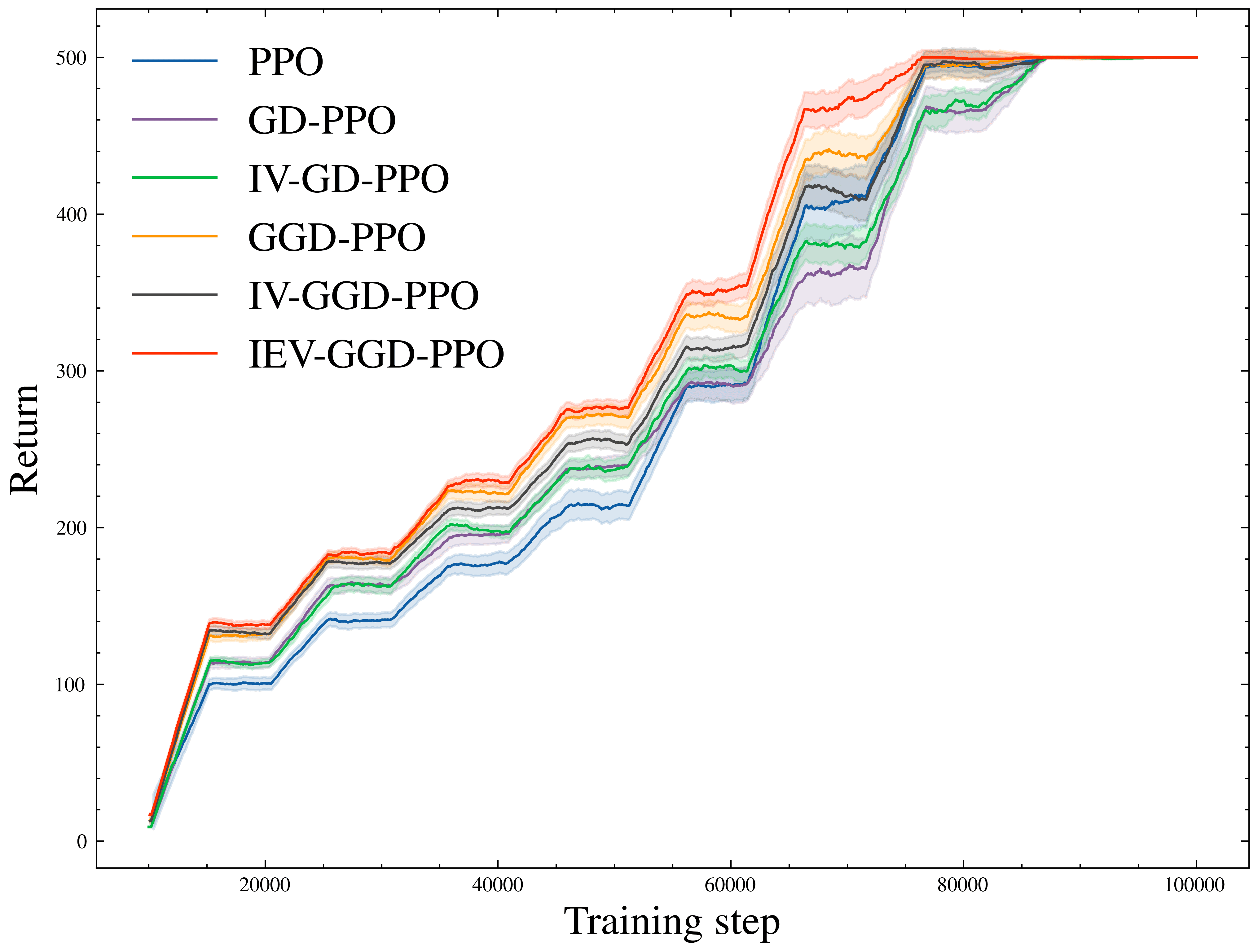}}
	\subfloat[LunarLander-v2]{\includegraphics[width=.31\textwidth]{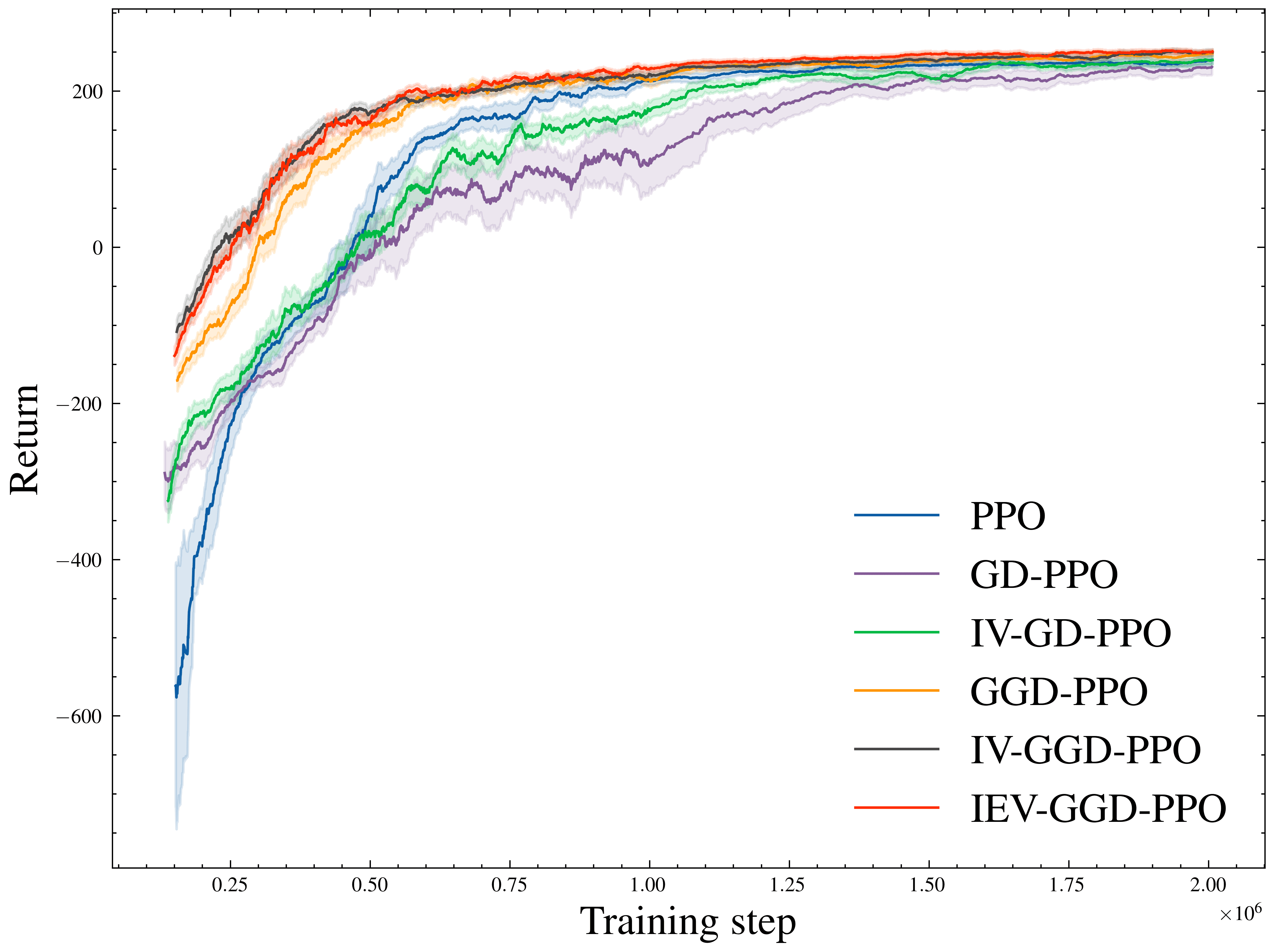}}
	\subfloat[HalfCheetah-v4]{\includegraphics[width=.31\textwidth]{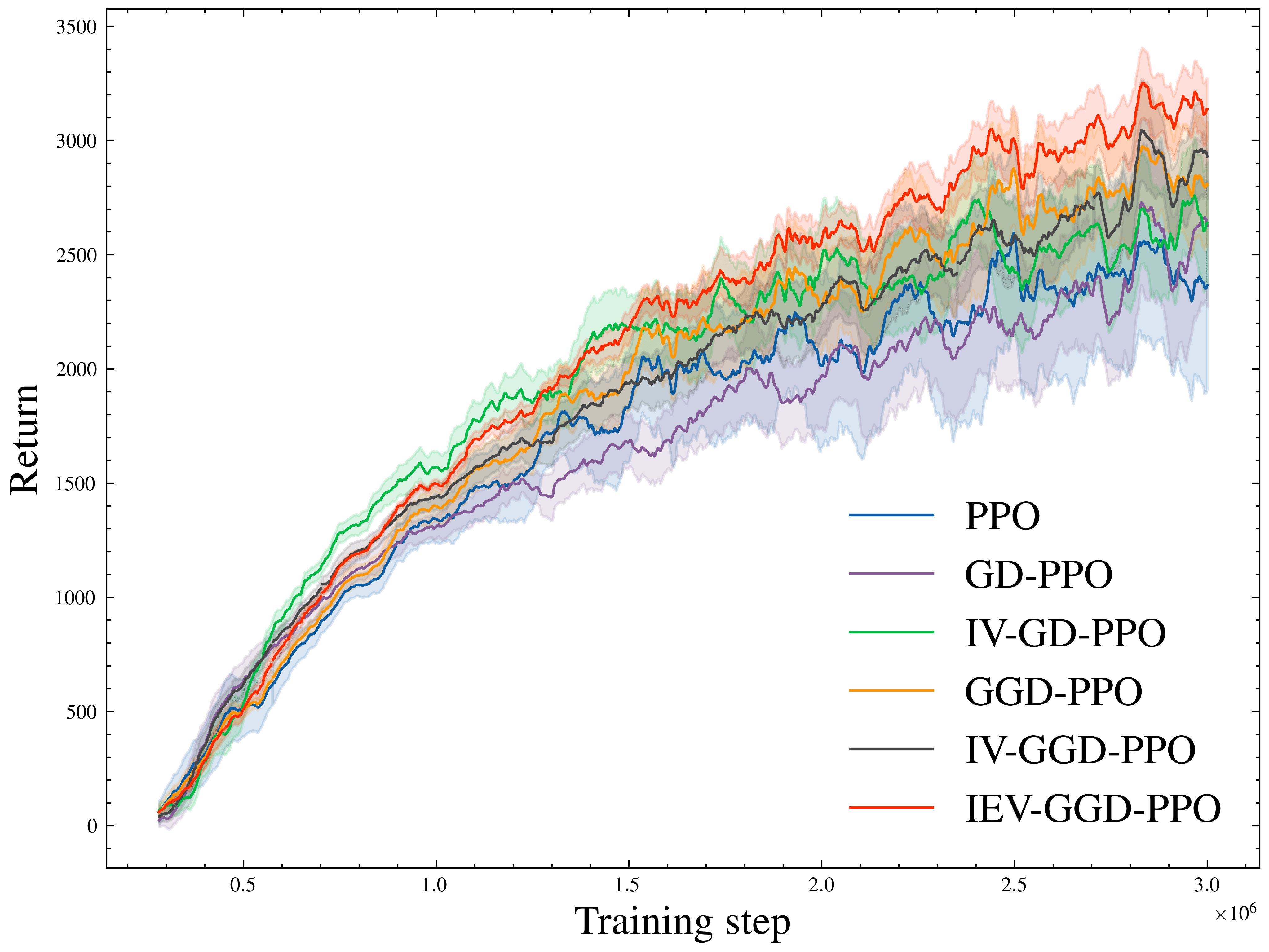}}
	\caption{
		Sample-efficiency curves of PPO on various control environments.
	} \label{fig:ppo}
\end{figure*}

\begin{figure*}[t]
	\centering
	\subfloat[Ant-v4]{\includegraphics[width=.31\textwidth]{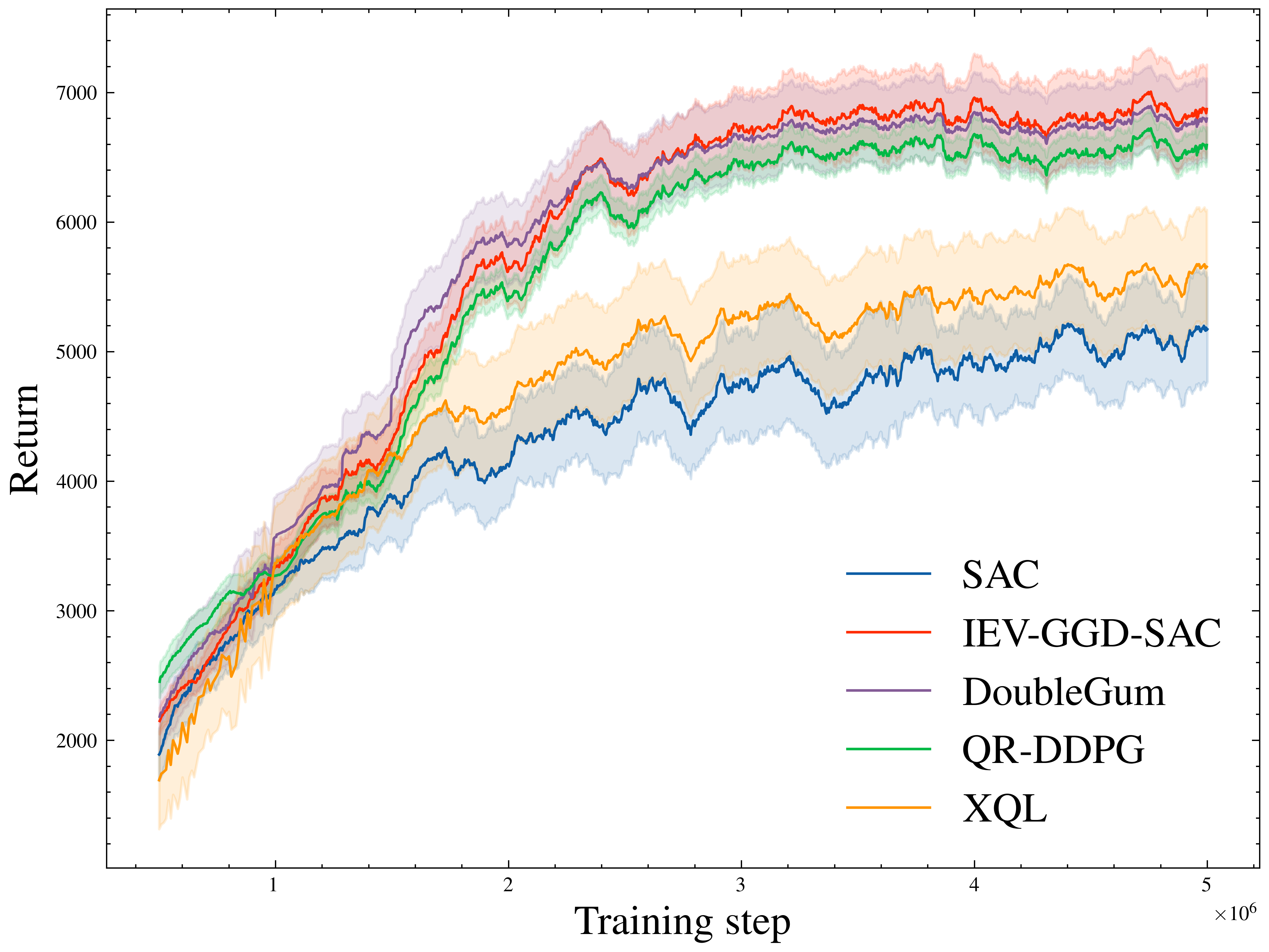}}
	\subfloat[HalfCheetah-v4]{\includegraphics[width=.31\textwidth]{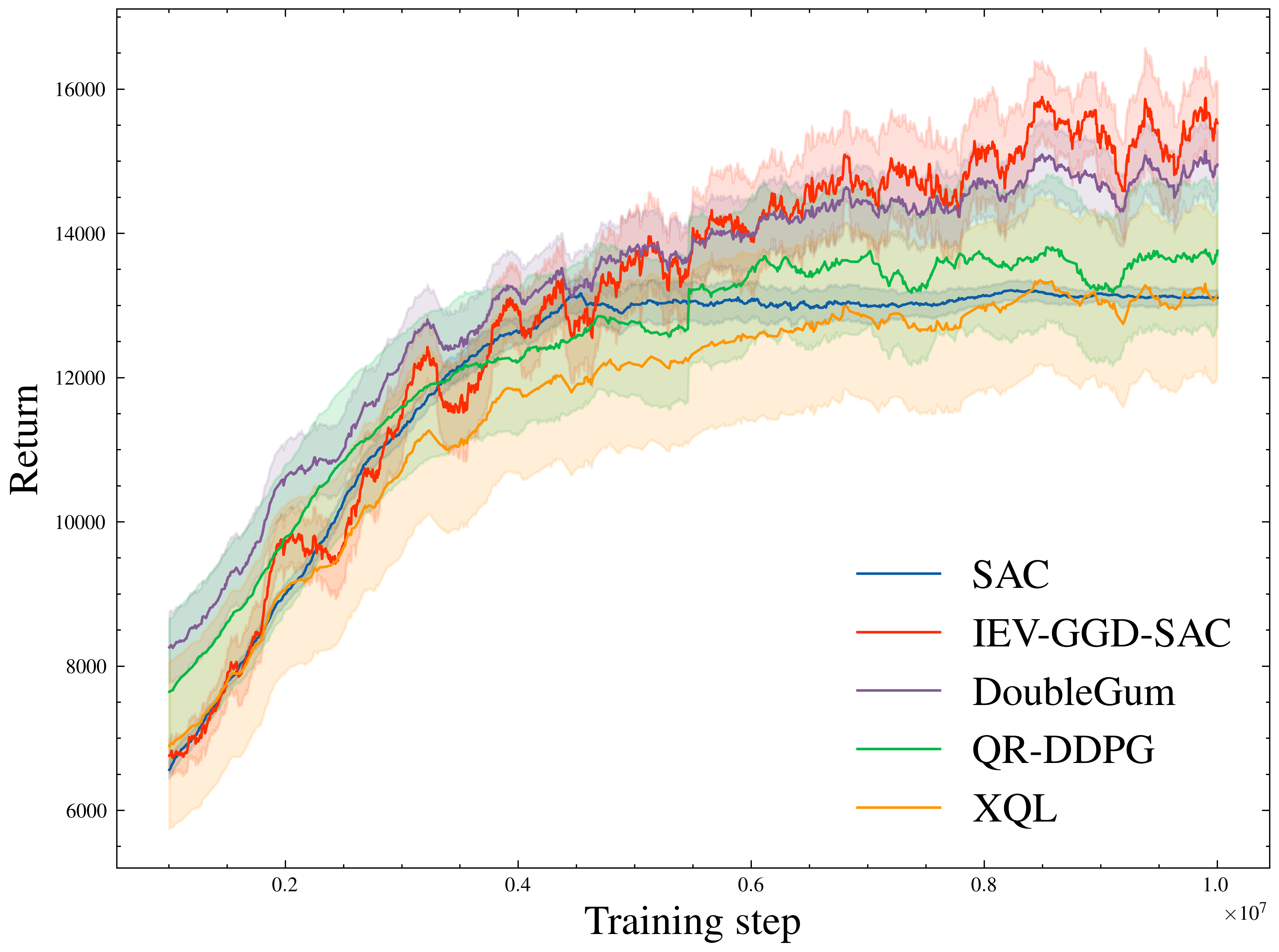}}
	\subfloat[Hopper-v4]{\includegraphics[width=.31\textwidth]{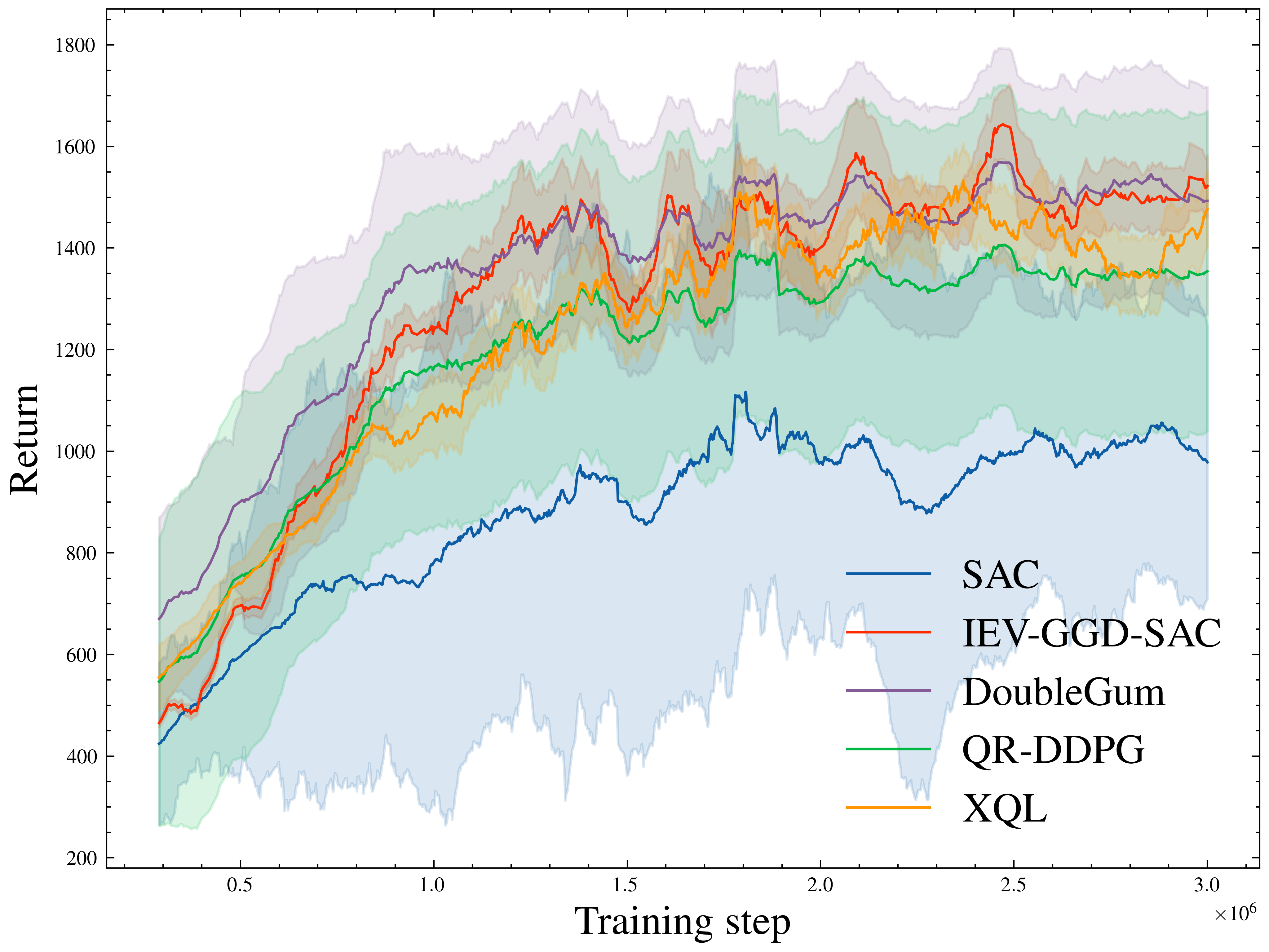}}
	\caption{
		Descriptive comparison with Gumbel-based (DoubleGum) and quantile-based (QR-DDPG) distributional RL methods on MuJoCo environments.
		Additional results appear in~\cref{apdx:ext:comp}.
	} \label{fig:comp}
\end{figure*}


\section{Methods} \label{sec:method}

\subsection{Generalized Gaussian Error Modeling} \label{ssec:ggd}

We model heteroscedastic TD errors using a zero-mean symmetric GGD~\citep{zeckhauser1970linear,chai2019using,giacalone2020combined,upadhyay2021robustness}:
\begin{equation*}
	\delta \sim \ggd(0,\alpha,\beta) = \frac{\beta}{2 \alpha \Gamma(1/\beta)} e^{-\left( |\delta| / \alpha \right)^{\beta}},
\end{equation*}
where $\alpha$ and $\beta$ denote the scale and shape parameters, respectively.
This formulation captures a wide range of error behaviors and provides a unified framework connecting Gaussian to uniform distributions as $\beta$ increases~\citep{giller2005generalized,nadarajah2005generalized,novey2009complex,dytso2018analytical}.

The shape parameter $\beta$ governs the distribution's tailedness.
The excess kurtosis $\kappa$, defined as Pearson's kurtosis minus three, is solely a function of $\beta$~\citep{decarlo1997meaning}:
\begin{equation*}
	\kappa = \frac{\Gamma(5/\beta)\Gamma(1/\beta)}{\Gamma(3/\beta)^2} - 3.
\end{equation*}
When $\beta<2$, the distribution is leptokurtic ($\kappa>0$), indicating heavier tails than the Gaussian.
With only one additional parameter, GGD captures tail behavior that location-scale families cannot express.

\begin{rem} \label{rem:beta-head}
	We estimate only $\beta$ and fix $\alpha=1$ as a one-parameter design choice, not as an identifiability result.
	Because $\sigma^2=\alpha^2\Gamma(3/\beta)/\Gamma(1/\beta)$~\citep{dytso2018analytical}, this restriction makes $\beta$ control both tail thickness and dispersion rather than allowing independent scale adaptation.
	It reduces overhead and removes scale--shape co-adaptation; the corresponding ablation in~\cref{apdx:abl:alpha} evaluates this trade-off.
\end{rem}

For $\alpha=1$, $\sigma^2(\beta)=\Gamma(3/\beta)/\Gamma(1/\beta)$ decreases with $\beta$ on the leptokurtic interval $\beta\in(0,2]$ considered below; it is not globally decreasing for all $\beta>0$.
The observed degradation from jointly learning $(\alpha,\beta)$ is specific to our parameterization and benchmarks and does not establish that the two-parameter GGD is intrinsically unidentifiable.

\subsubsection{Empirical Evidence} \label{sssec:empirical}


\cref{fig:tde-sac} shows pooled TD errors departing from Gaussianity and appearing heavier-tailed at later evaluations.
We hypothesize that this leptokurtic behavior is amplified during exploration, where occasional large TD errors arise even as epistemic uncertainty gradually decreases with experience~\citep{kendall2017uncertainties}.
This evolving interplay between exploration-induced outliers and irreducible environmental stochasticity is consistent with the emergence of heavy-tailed error distributions.
This shift is also consistent with the fading Gumbel-like behavior discussed in~\cref{ssec:tail}.

\paragraph{Marginal versus Conditioned Shape.}\label{par:marginal-vs-persample}
The histograms in~\cref{fig:tde-sac,apdx:ext:tde} pool TD errors over many state--action pairs and motivate non-Gaussian modeling at the marginal level.
Training instead evaluates a state-conditioned network output $Q^\beta_k(s_t,a_t)$ for each transition and critic; it does not estimate one global shape parameter from the plotted histograms.

\subsubsection{Theoretical Analysis} \label{sssec:theoretical}

We first separate an established property of the GGD density kernel from the validity of the likelihood itself.

\begin{thm}[Positive definiteness of the GGD density kernel~\citep{bochner1937stable,ushakov2011selected,dytso2018analytical}] \label{thm:ggd-nll}
	For $\beta\in(0,2]$, the function $k_\beta(x)=\exp(-|x|^\beta)$ is positive definite.
	Separately, the GGD density in~\cref{ssec:ggd} is normalized and its exact NLL is finite for every $\beta>0$ and finite residual $\delta$.
\end{thm}

Positive definiteness here is a property of the density kernel as a function, not a claim that the NLL Hessian is positive definite or that optimization is stable.
Normalization follows directly from integration and does not require $\beta\le2$.
See~\cref{apdx:proof:ggd-nll} for the proof and parameter-domain discussion.

\paragraph{GGD Loss as an Online TD Surrogate.}\label{par:online-surrogate}
Bootstrapped targets evolve with the critic and policy, so we use a numerically modified GGD loss as an online surrogate rather than claim offline MLE consistency for a fixed residual distribution.
The learned shape trajectories remain bounded and task-dependent (\cref{fig:beta-head}), and their aggregate coefficients of variation are generally lower than those of the Gaussian scale head (\cref{fig:param}); these observations support numerical use of the surrogate but are not a consistency proof.

\paragraph{Scope.}
The restriction $\beta\in(0,2]$ applies only to the positive-definite-kernel statement in~\cref{thm:ggd-nll}.
The GGD remains a valid density for $\beta>2$, but neither the theorem nor our experiments provide a general optimization guarantee in that regime.

\paragraph{FSD versus SSD.}\label{par:fsd-ssd}
First-order stochastic dominance (FSD) compares CDFs pointwise and represents monotone preferences; it does not usefully order distinct centered symmetric error distributions.
Second-order stochastic dominance (SSD) compares integrated CDFs and represents monotone concave preferences, including aversion to dispersion.
The precise established ordering requires the scale convention stated next.

\begin{thm}[Second-order stochastic dominance~\citep{dytso2018analytical}] \label{thm:ssd}
	Let $X_i\sim\ggd(0,2^{1/\beta_i}\alpha,\beta_i)$, where $\alpha>0$ and $\beta_i>0$.
	If $\beta_1\leq\beta_2$, then $X_2$ exhibits second-order stochastic dominance over $X_1$.
	This dominance implies, for all $x$,
	\begin{equation} \label{eq:ssd}
		\int_{-\infty}^x \left[F_{X_1}(t)-F_{X_2}(t)\right] dt\geq 0,
	\end{equation}
	where $F$ denotes the cumulative distribution function.
\end{thm}

The factor $2^{1/\beta_i}$ is part of the parameterization used by the cited ordering result.
Our implementation instead fixes the GGD scale to $\alpha=1$, so \cref{thm:ssd} motivates a monotone preference in $\beta$ but does not uniquely derive our weighting rule.
See~\cref{apdx:proof:ssd} for the parameterization mapping.

Throughout, $\beta$ denotes the abstract GGD shape parameter, $Q^\beta_k(s,a)$ the $k$th critic's network output before positivity enforcement, and $\hat\beta_{t,k}=\operatorname{softplus}(Q^\beta_k(s_t,a_t))$ its realized estimate.
We use the parameter-free monotone rule
\begin{equation*}
	\omega^{\mathrm{RA}}_{t,k}=\frac{\hat\beta_{t,k}}{\sum_{j=1}^{K}\hat\beta_{t,j}},
\end{equation*}
which normalizes weights across critics for each transition.

\begin{rem} \label{rem:aleatoric-unc}
	The training of the critic is more directly influenced by aleatoric uncertainty, since only the critic loss is a direct function of the state, action, and reward, with the actor being downstream of the critic in uncertainty propagation.
	GGD error modeling enables us to quantify aleatoric uncertainty in closed form, i.e., $\sigma^2=\alpha^2\Gamma(3/\beta)/\Gamma(1/\beta)$~\citep{upadhyay2021robustness}.
	For fixed $\alpha=1$ and $\beta\in(0,2]$, this variance decreases as $\beta$ increases.
	We therefore give larger surrogate weight to estimates with larger $\beta$; this is a design heuristic under our fixed-scale parameterization, not a consequence uniquely implied by SSD.
\end{rem}

As a negative control, \cref{apdx:abl:raw,fig:abl-raw} reverses the monotonicity.
The inverse rule produces periodic non-monotone performance jumps, but no exploration metric was measured.
This comparison supports the chosen direction in our ablation, while alternative monotone mappings remain untested.


\subsection{Batch Inverse Error Variance Regularization} \label{ssec:biev}

For transition $t$ and critic $k$, let $\delta_{t,k}$ denote the TD error.
BIV uses disagreement among the critics' value predictions, whereas BIEV summarizes the distribution of the $K$ TD errors themselves.
Define the population-normalized empirical variance
\begin{equation*}
	v_t=\frac{1}{K}\sum_{k=1}^{K}(\delta_{t,k}-\bar\delta_t)^2
\end{equation*}
and let $\hat\kappa_t$ be the bias-corrected sample excess kurtosis across those errors.
Our implementation applies the MSE-best-estimator-inspired factor
\begin{equation*}
	\widetilde s_t^2=\left(\frac{\hat\kappa_t}{K}+\frac{K+1}{K-1}\right)^{-1}v_t
\end{equation*}
and forms
\begin{equation} \label{eq:biev-weight}
	\omega^\text{BIEV}_t = \frac{1}{\widetilde{s}^2_t + \xi}.
\end{equation}
The stabilizer $\xi$ is selected by numerical search to target an effective batch size of $\min\{|\gB|-1,m\}$, subject to optimizer tolerance, following BIV; we set $m=16$.
Inversion remains nonlinear, so this construction is a regularization surrogate rather than an unbiased inverse-variance estimator.

The multiplicative factor is motivated by the following established result~\citep{kleffe1985some,searls1990note,wencheko2009estimation}.

\begin{pro}[MBBE of variance~\citep{searls1990note,wencheko2009estimation}] \label{pro:mbbe}
	Let $S^2=(n-1)^{-1}\sum_{i=1}^n(X_i-\bar X)^2$ be the unbiased sample variance of i.i.d. observations with variance $\sigma^2$ and excess kurtosis $\kappa$.
	Among estimators $cS^2$ with constant $c$, the mean-squared-error (MSE) minimizer is
	\begin{equation*}
		c^*S^2 = \left( \frac{\kappa}{n} + \frac{n+1}{n-1} \right)^{-1} S^2.
	\end{equation*}
	Its relative efficiency with respect to $S^2$ is
	\begin{equation*}
		\mathrm{RE}_n = \frac{\sV[S^2]}{\mse(c^*S^2)} = 1 + \frac{\kappa}{n} + \frac{2}{n-1}.
	\end{equation*}
\end{pro}

Our code uses $v_t$ with divisor $K$, rather than the unbiased $S^2$ in~\cref{pro:mbbe}, and substitutes the sample estimate $\hat\kappa_t$ for the unknown population kurtosis.
With only $K=5$ errors, the fourth-moment estimate is noisy.
We therefore call $\widetilde s_t^2$ MBBE-inspired and treat BIEV as a heuristic, not an exact application of the proposition.

\paragraph{BIV versus BIEV.}\label{par:biv-vs-biev}
BIV~\citep{mai2022sample} uses the unbiased across-critic variance of mean value predictions (scaled by $\gamma^2$ in our implementation).
BIEV instead uses the population-normalized variance and sample excess kurtosis of the critics' TD errors; it does not obtain kurtosis from the $\beta$ head.
Both methods add an effective-batch-size stabilizer and normalize inverse weights across the minibatch.

\paragraph{Empirical Positioning.}\label{par:biev-insurance}
BIEV is not presented as a uniform performance gain.
In our ablations, it is often near BIV and sometimes improves asymptotic performance (\cref{apdx:abl:biev}); the comparison is empirical and does not establish a heavy-tail correction guarantee.

For the GGD variants evaluated in the released implementation, BIEV weights absolute TD errors:
\begin{equation*}
	\Ls_{\mathrm{BIEV}} = \sum_t{
		\frac{1}{|\gB|}\frac{\omega^{\mathrm{BIEV}}_t}{\sum_{\tau} \omega^{\mathrm{BIEV}}_{\tau}}
		\sum_{k=1}^{K}|\delta_{t,k}|
	}.
\end{equation*}
The complete critic objective combines the risk-weighted GGD surrogate with BIEV regularization:
\begin{equation} \label{eq:loss}
	\Ls  = \Ls^\text{RA}_\text{GGD-sur} + \lambda \Ls_\text{BIEV}.
\end{equation}
The complete critic update procedure is summarized in~\cref{apdx:impl}.

\subsection{Implementation and Numerical Stability} \label{ssec:impl-stab}

For numerical stability, we apply softplus to the $\beta$ outputs, evaluate normalization terms in log space, floor $\widetilde s_t^2$ at $\epsilon=10^{-6}$, and select $\xi$ by numerical search to target $\min\{|\gB|-1,16\}$ effective samples, subject to optimizer tolerance.
The released implementation does not clip $\hat\kappa_t$ to $[0,10]$.
The $\beta$ head adds $O(H)$ parameters per critic, and BIEV computes $O(BK)$ minibatch statistics.
Refer to~\cref{apdx:impl} for detailed numerical considerations.


\section{Experiments} \label{sec:exp}

We evaluate our method on MuJoCo~\citep{todorov2012mujoco} and discrete control environments from Gymnasium~\citep{towers2023gymnasium}, augmenting discrete environments with uniform dynamics perturbations.
All compared variants were configured to use the same registered environment perturbations.
Clean-environment results are reported in~\cref{fig:abl-orig-env}; there the effects are smaller and mixed, so we do not claim that the perturbation study establishes a universal advantage.

We select baselines covering diverse RL paradigms: Soft Actor-Critic (SAC)~\citep{haarnoja2018soft}, an off-policy $Q$-based algorithm, and Proximal Policy Optimization (PPO)~\citep{schulman2017proximal}, an on-policy $V$-based method.
We focus on variance-network baselines for computational efficiency; uncertainty prediction requires one additional output head.
Comparisons with representative return-distributional RL alternatives, including DoubleGum and QR-DDPG, appear in~\cref{fig:comp}, with full results in~\cref{apdx:ext:comp}.

We implement all algorithms in PyTorch using Stable-Baselines3~\citep{raffin2021stable}; algorithm-specific settings are specified in the released configuration files, while unlisted constructor arguments retain their implementation defaults.
Both PPO and SAC variants use $K=5$ critics, target effective-batch-size parameter $m=16$, and $\lambda=0.1$; $\xi$ is selected numerically as described in~\cref{ssec:impl-stab}.
Additional details appear in~\cref{apdx:exp}.

\paragraph{Reproducibility.}\label{par:reproducibility}
The implementation, dependencies, and training configurations are available at \url{https://github.com/ait-lab/ggtde}; environment details appear in~\cref{apdx:exp}.
We report neither aggregate GPU-hours nor seed counts because no auditable run-level accounting or immutable run manifest is available.


\cref{fig:sac-mujoco} presents SAC performance across MuJoCo environments.
Under the released aggregation, the $\beta$-head variants improve several plotted central curves, including HalfCheetah-v4 and Hopper-v4, while other settings are mixed; BIEV is often near BIV and improves some asymptotic curves, without uniform dominance.



\cref{fig:param} reports the coefficient of variation (CV), i.e., $\sqrt{\sV[X]}/\E[X]$, for variance- and $\beta$-head estimates.
The aggregate $\beta$ trajectories generally have lower CV than the Gaussian scale-head trajectories in the plotted tasks, providing descriptive evidence of smoother estimates under the evaluated setup rather than proof of universally better conditioning.



\cref{fig:ppo} shows PPO results across MuJoCo and discrete control environments.
The $\beta$ head and BIEV improve several plotted curves, with smaller or mixed differences elsewhere; this supports applicability to $V$-based TD learning but does not establish uniform gains.



\cref{fig:comp} provides a descriptive comparison with Gumbel-based DoubleGum and quantile-based QR-DDPG~\citep{garg2022extreme,hui2023double} on Ant, Hopper, and HalfCheetah.
Performance is broadly comparable in some environments and differs in others; because architectures and evaluation pipelines are not fully matched, this is not a controlled state-of-the-art ranking.
The GGD surrogate is implemented in DQN, PPO, and SAC and adds one scalar $\beta$ output per critic, whereas return-distributional methods use richer representations; this is an architectural contrast rather than a demonstrated accuracy-to-complexity optimum.


Ablations in~\cref{apdx:ablation} examine each component.
The monotone $\beta$ weighting improves early learning, whereas its inverse is unstable; this supports the chosen direction but not the linear form as optimal.
BIEV is often near BIV, and the tested temperature range $\lambda\in[0.05,0.5]$ is not highly sensitive in these environments.
Adding an $\alpha$ head reduces sample efficiency, supporting the simpler one-parameter design for the reported experiments only.
\cref{apdx:qlearn} separates the components: the GGD-inspired critic loss transfers directly, whereas shape-based risk weighting can conflict with greedy exploration and remains scoped to the stochastic policy-gradient setting evaluated here.


\section{Discussion} \label{sec:disc}

Our results motivate treating TD-error shape as an additional modeling signal rather than assuming a Gaussian scale model throughout training.
The proposed surrogate changes only the critic output and loss, but the observed performance and uncertainty effects remain benchmark-dependent.

Shape modeling may be particularly relevant when TD errors are leptokurtic, for example during exploration, sparse rewards, or aggressive bootstrapping.
The learned aggregate $\beta$ trajectories in~\cref{fig:beta-head} remain below the Gaussian value $2$ but are environment-dependent and can exceed $1.2$; the plots therefore support a leptokurtic focus without establishing a universal numerical bound.

\paragraph{Further Investigation.}
Integrating GGD into maximum-entropy RL is a promising direction: Gaussians maximize entropy under a second-moment constraint, while GGDs maximize entropy under a $p$-th absolute moment constraint~\citep{mackay2003information}.
Leveraging higher-order moments may yield new perspectives for entropy-regularized objectives and policy smoothing.
A regret analysis under heavy-tailed TD errors is also natural, given the role of reward-driven aleatoric noise and existing results for heavy-tailed rewards~\citep{cayci2024provably,zhuang2021no}.
Although our main experiments focus on policy gradients, \cref{apdx:qlearn} indicates that the GGD-inspired critic loss, but not necessarily the risk-weighting rule, can be evaluated in $Q$-learning.
Generalized extreme-value models may also clarify tail phenomena.

\paragraph{Relevant Applications.}
Potential applications include robust and risk-sensitive RL in which rare TD errors are operationally important.
However, our simulation benchmarks do not establish calibration, safety, or reliability in real systems; those claims require task-specific validation.

\paragraph{Limitations.}
The GGD assumes symmetric errors; asymmetric or multimodal TD errors (e.g., in multi-goal settings) require extensions such as the skewed GGD.
Fixing $\alpha=1$ conflates scale and shape, and the implemented GGD loss is a modified surrogate rather than the exact NLL.
BIEV applies an MBBE-inspired factor to a population-normalized ensemble variance and estimates excess kurtosis from only $K=5$ TD errors; the released implementation does not clip that estimate, and the empirical gain over BIV is regime-dependent.
The method inherits ensemble cost scaling linearly with $K$ and has not been compared directly with Student-$t$ regression.
\cref{thm:ggd-nll} concerns positive definiteness of the density kernel for $\beta\in(0,2]$; it is not an optimization guarantee.
Finally, real-world transfer, distribution shift, and safety-critical use remain untested.


\begin{acknowledgements}
	We are grateful for the helpful advice provided by Dr. Heather Battey in the Department of Mathematics, Imperial College London.
	Her insights and feedback have been invaluable to the development and improvement of this work.
\end{acknowledgements}

\bibliography{references}

\newpage \onecolumn
\raggedbottom

\renewcommand\thefigure{\Alph{figure}}
\setcounter{figure}{0}

\begin{center}
	\Large\bfseries
	Supplementary Material for\\
	Generalized Gaussian Temporal Difference Error
\end{center}
\appendix

\section{Extended Results} \label{apdx:extended}

\subsection{Temporal Difference Error Plots} \label{apdx:ext:tde}

\begin{figure}[H]
	\centering
	\subfloat[HalfCheetah-v4]{\includegraphics[width=.65\textwidth]{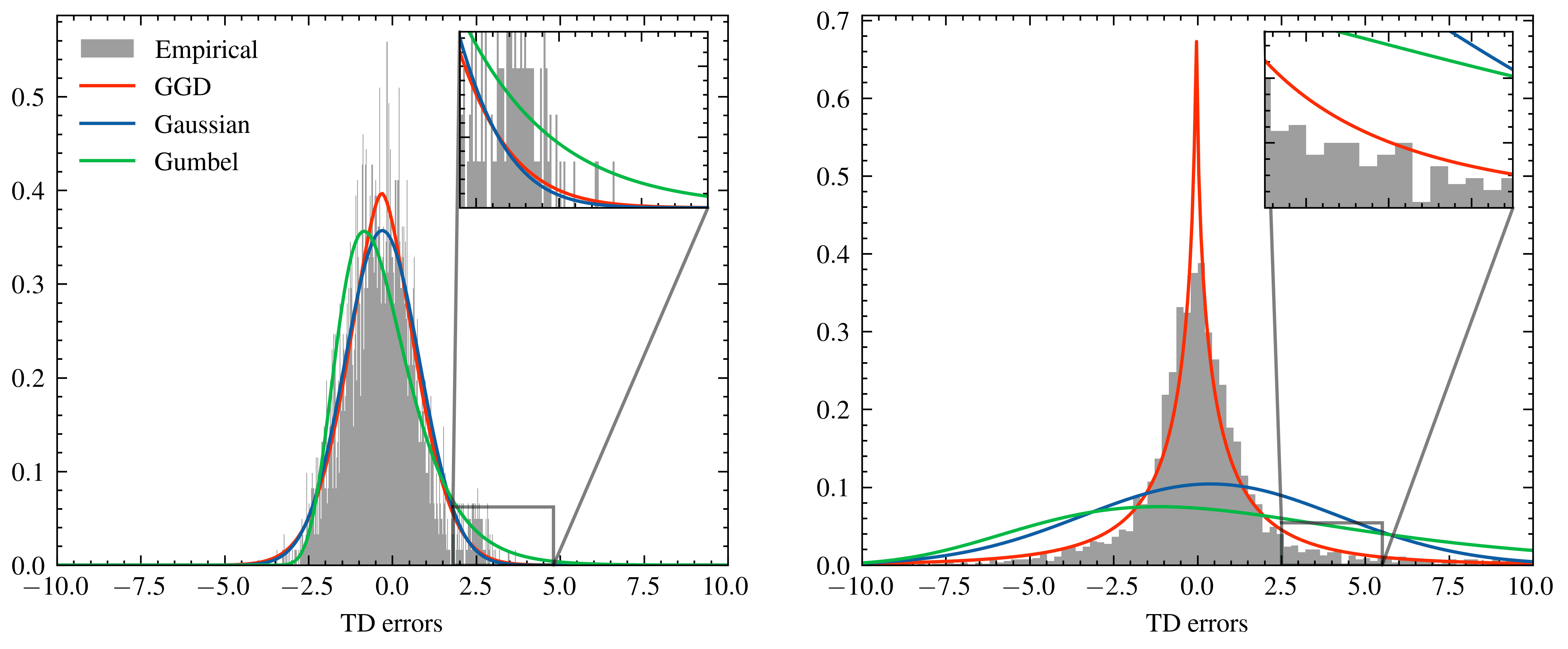}} \\
	\subfloat[Humanoid-v4]{\includegraphics[width=.65\textwidth]{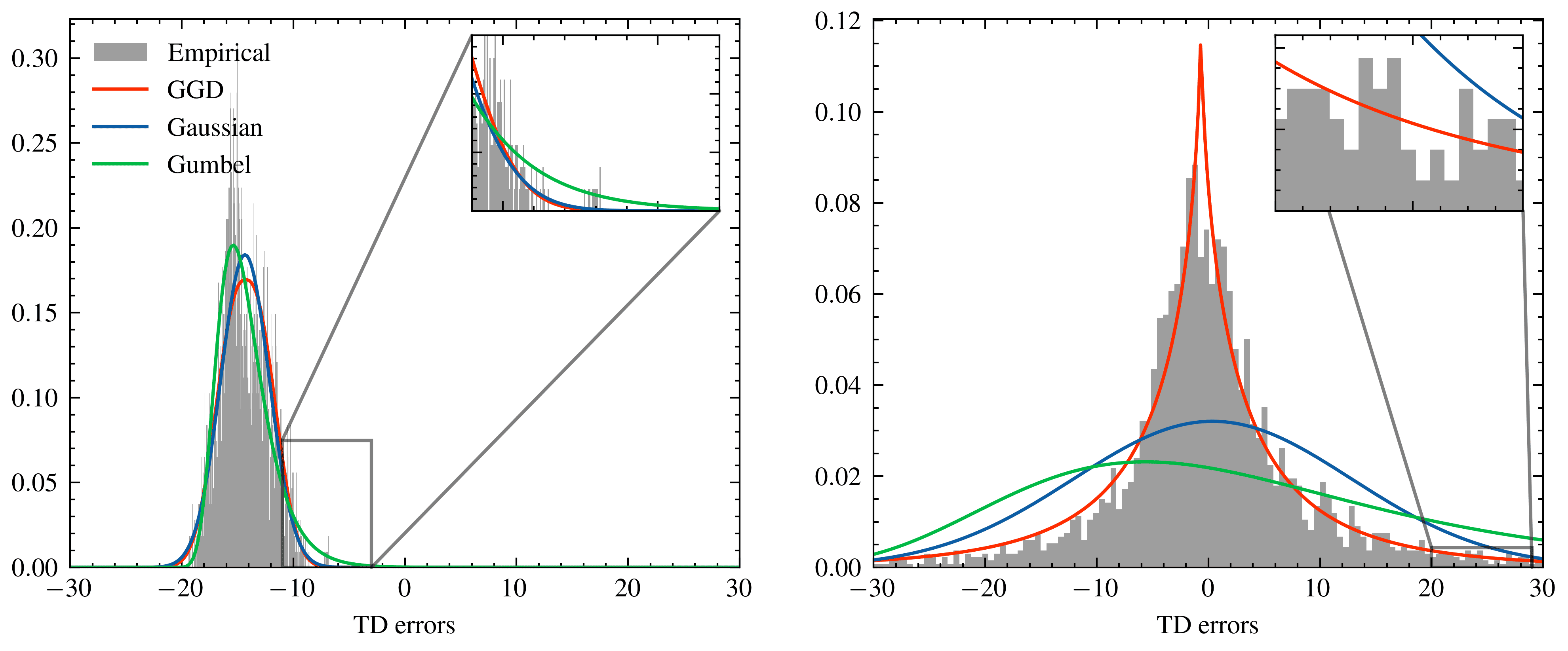}} \\
	\subfloat[Walker2D-v4]{\includegraphics[width=.65\textwidth]{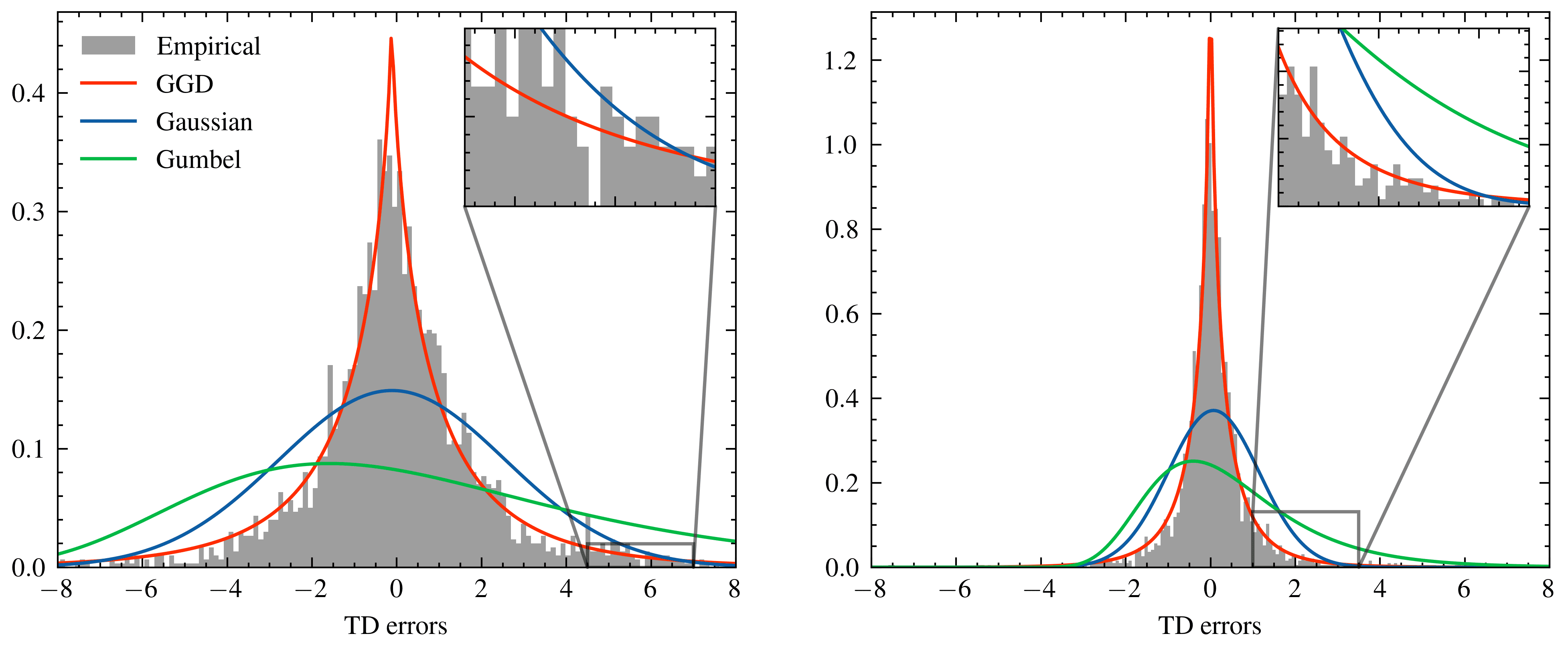}}
	\caption{
		TD error distributions of SAC at the initial and final evaluations (left to right) on additional MuJoCo environments.
	} \label{fig:tde-sac2}
\end{figure}

\begin{figure}[H]
	\centering
	\subfloat[HalfCheetah-v4]{\includegraphics[width=.65\textwidth]{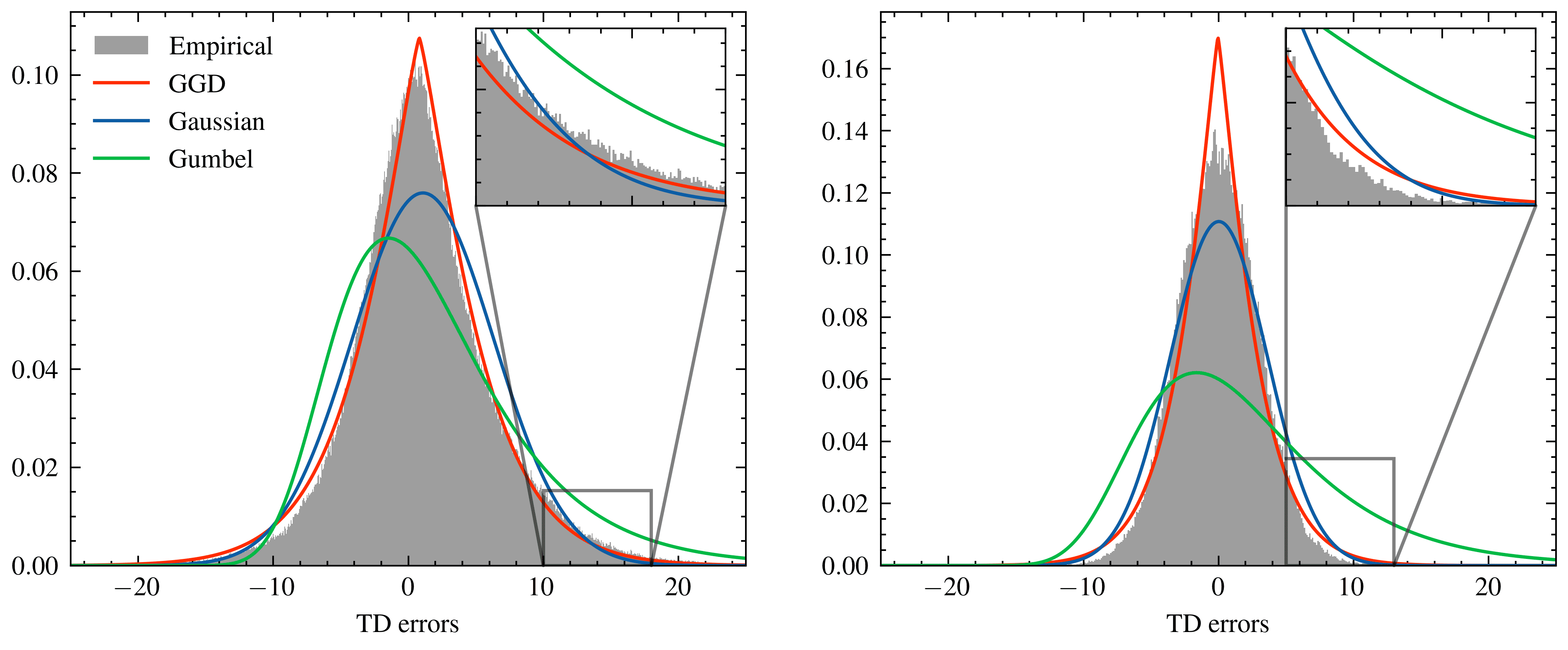}} \\
	\subfloat[Walker2D-v4]{\includegraphics[width=.65\textwidth]{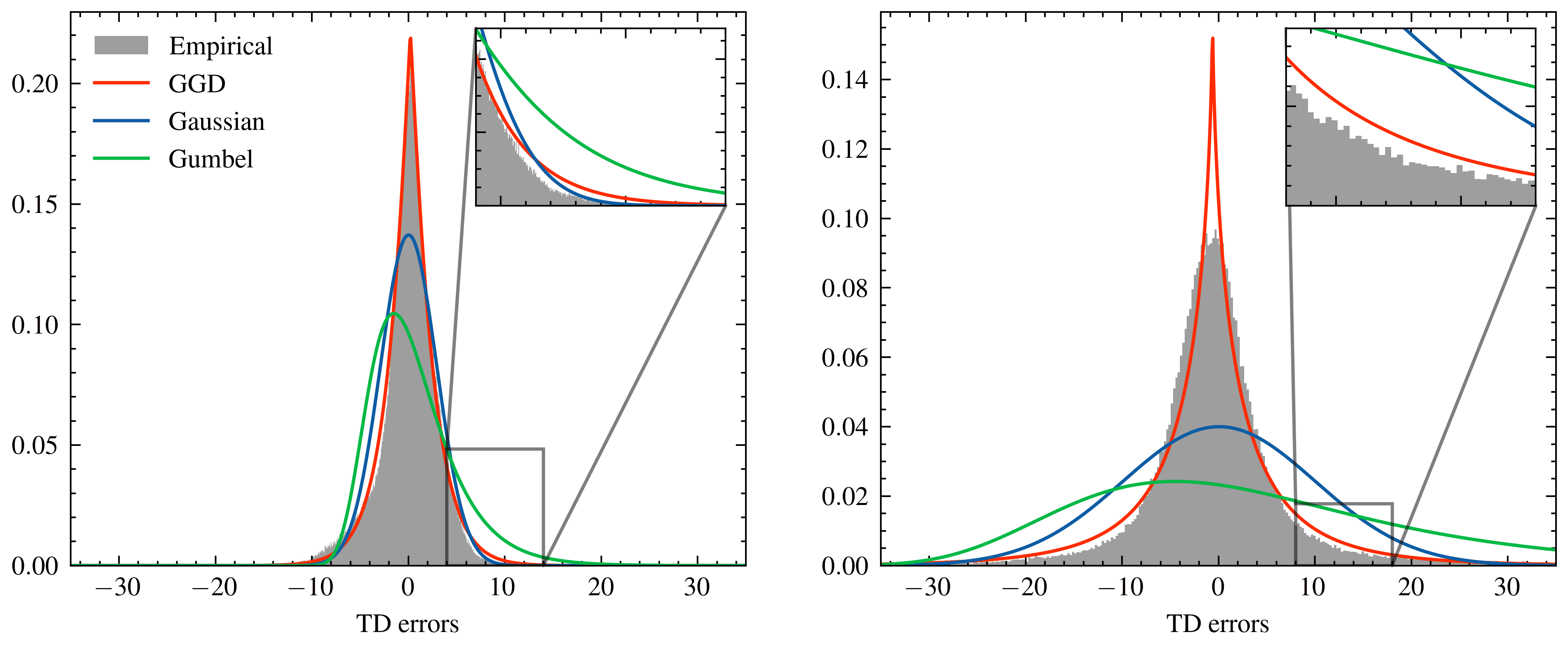}} \\
	\subfloat[LunarLander-v2]{\includegraphics[width=.65\textwidth]{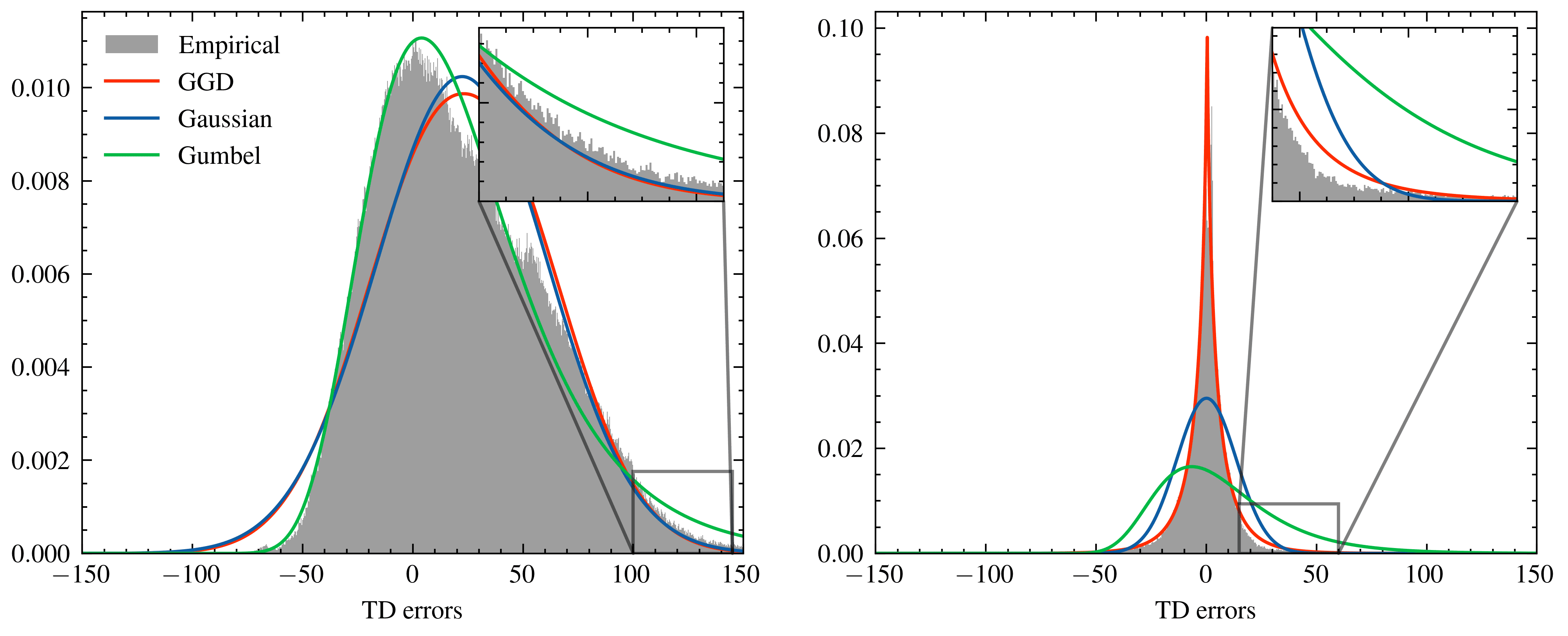}} \\
	\subfloat[MountainCar-v0]{\includegraphics[width=.65\textwidth]{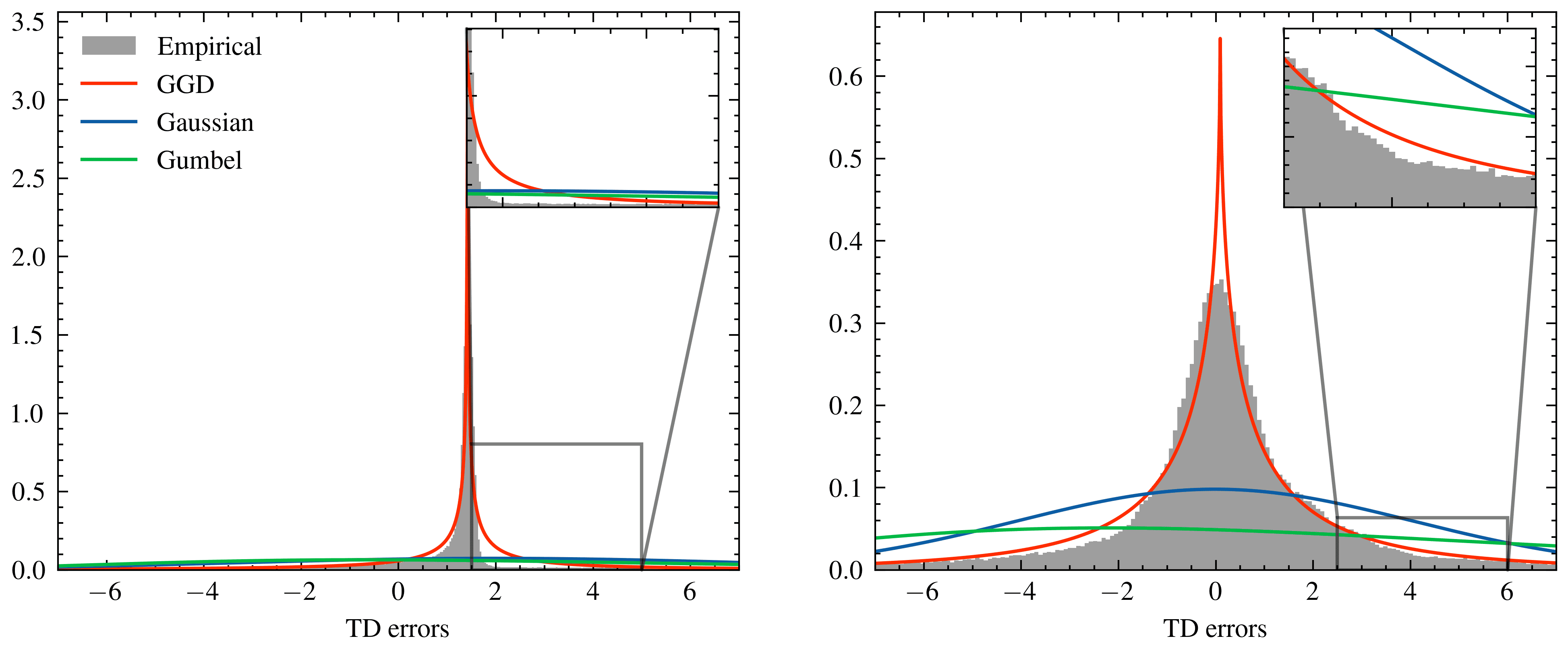}}
	\caption{
		TD error distributions of PPO at the initial and final evaluations (left to right) on additional control environments.
	} \label{fig:tde-ppo}
\end{figure}

We present the distributions of TD errors sampled at the initial and final evaluation steps, depicted in~\cref{fig:tde-sac,fig:tde-sac2} for SAC, and~\cref{fig:tde-ppo} for PPO, which highlights the heavy tailedness of TD errors and the tendency to converge toward a heavy tail throughout training.
This finding also emphasizes how aleatoric uncertainty affects their distribution, as elaborated in~\cref{ssec:ggd}.
Interestingly, both state-action values $Q$ and state values $V$ demonstrate similar characteristics in their TD error distributions.

\subsection{SAC on Other Environments} \label{apdx:ext:sac}

\begin{figure}[H]
	\centering
	\subfloat[HalfCheetah-v4]{\includegraphics[width=.31\textwidth]{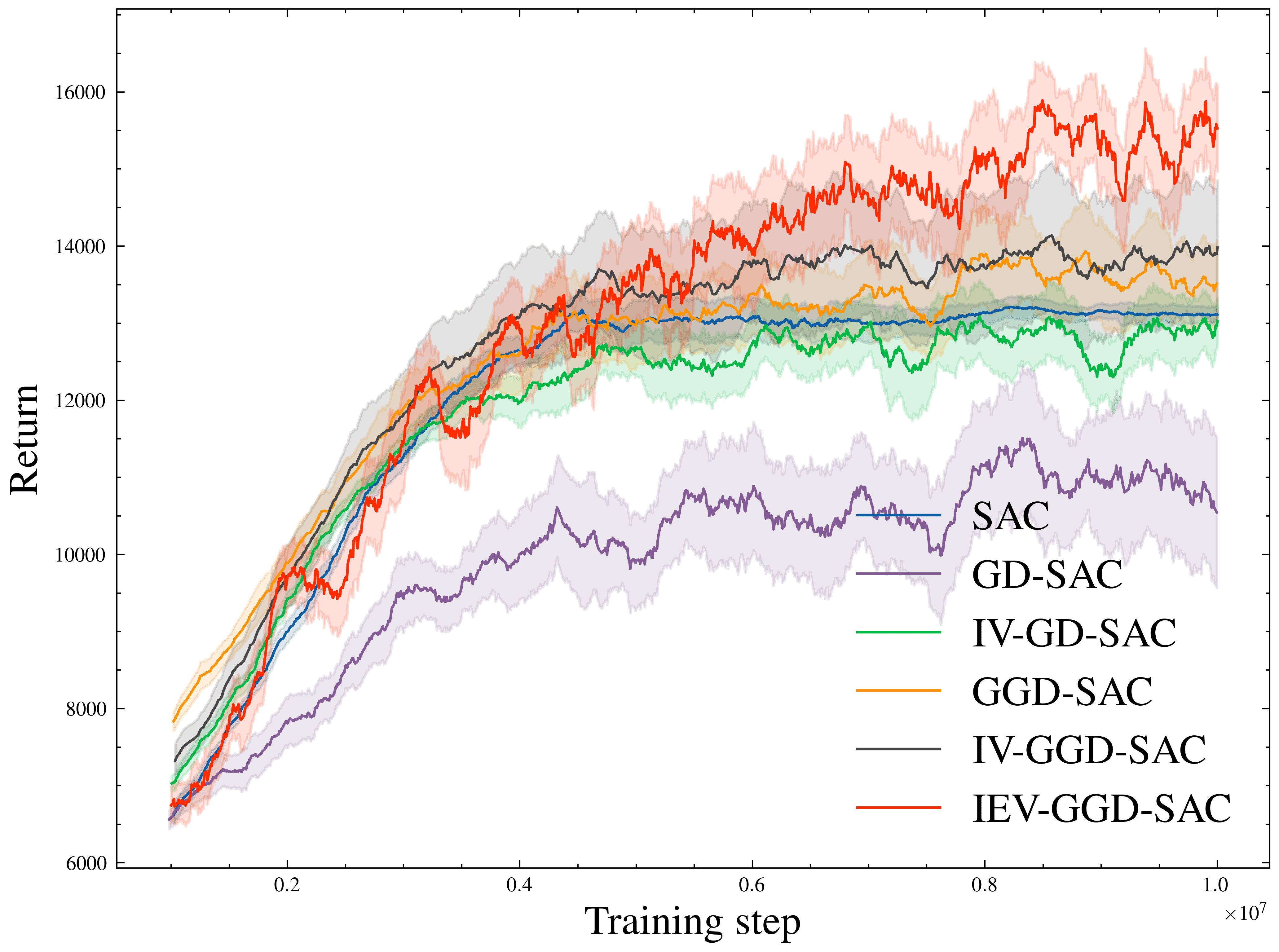}}
	\subfloat[Humanoid-v4]{\includegraphics[width=.31\textwidth]{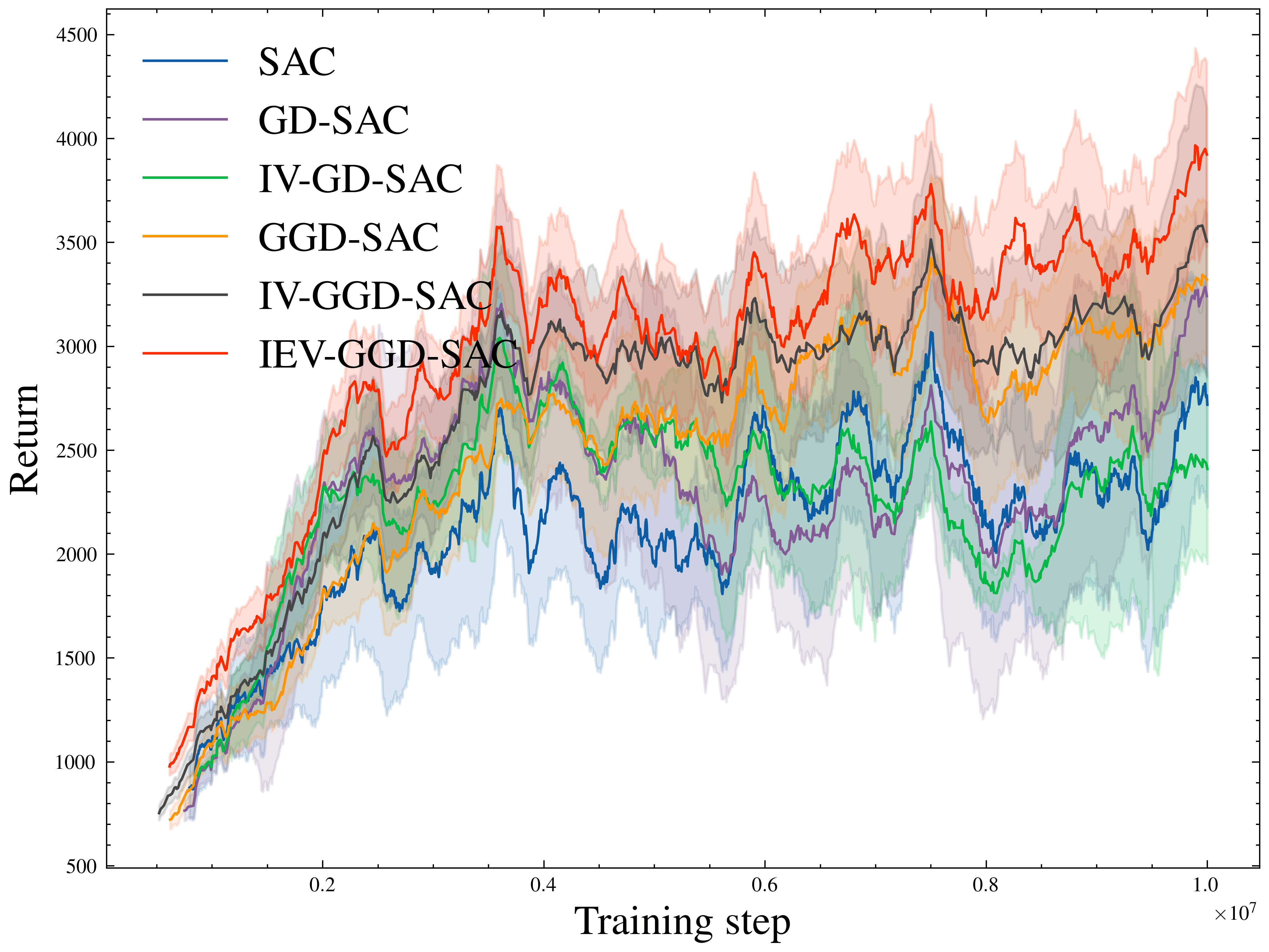}}
	\caption{
		Sample-efficiency curves of SAC on the remaining MuJoCo environments.
	} \label{fig:sac-extended}
\end{figure}


\subsection{Coefficients of Variation of Parameter Estimation} \label{apdx:ext:param}

\begin{figure}[H]
	\centering
	\subfloat[HalfCheetah-v4]{\includegraphics[width=.31\textwidth]{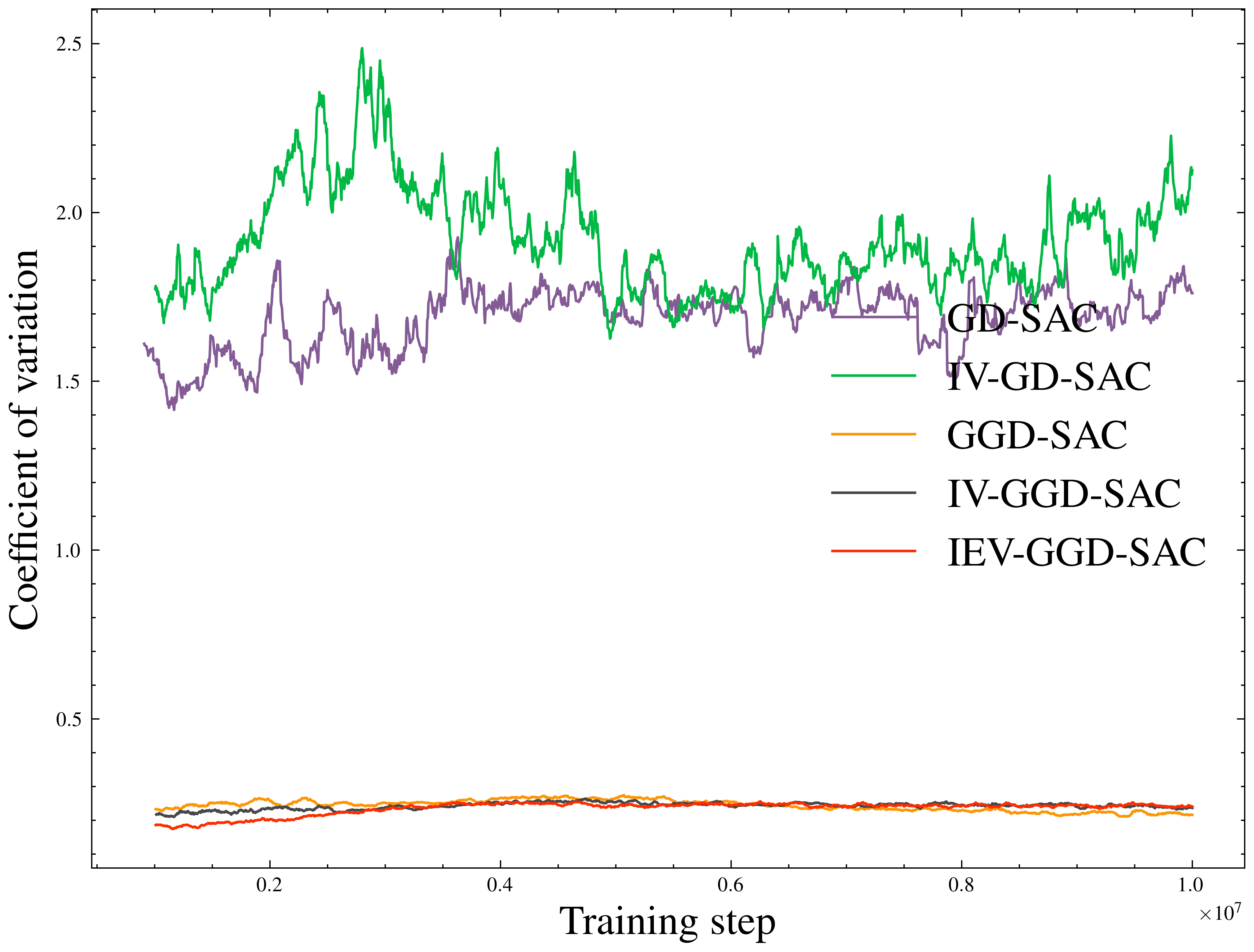}}
	\subfloat[Humanoid-v4]{\includegraphics[width=.31\textwidth]{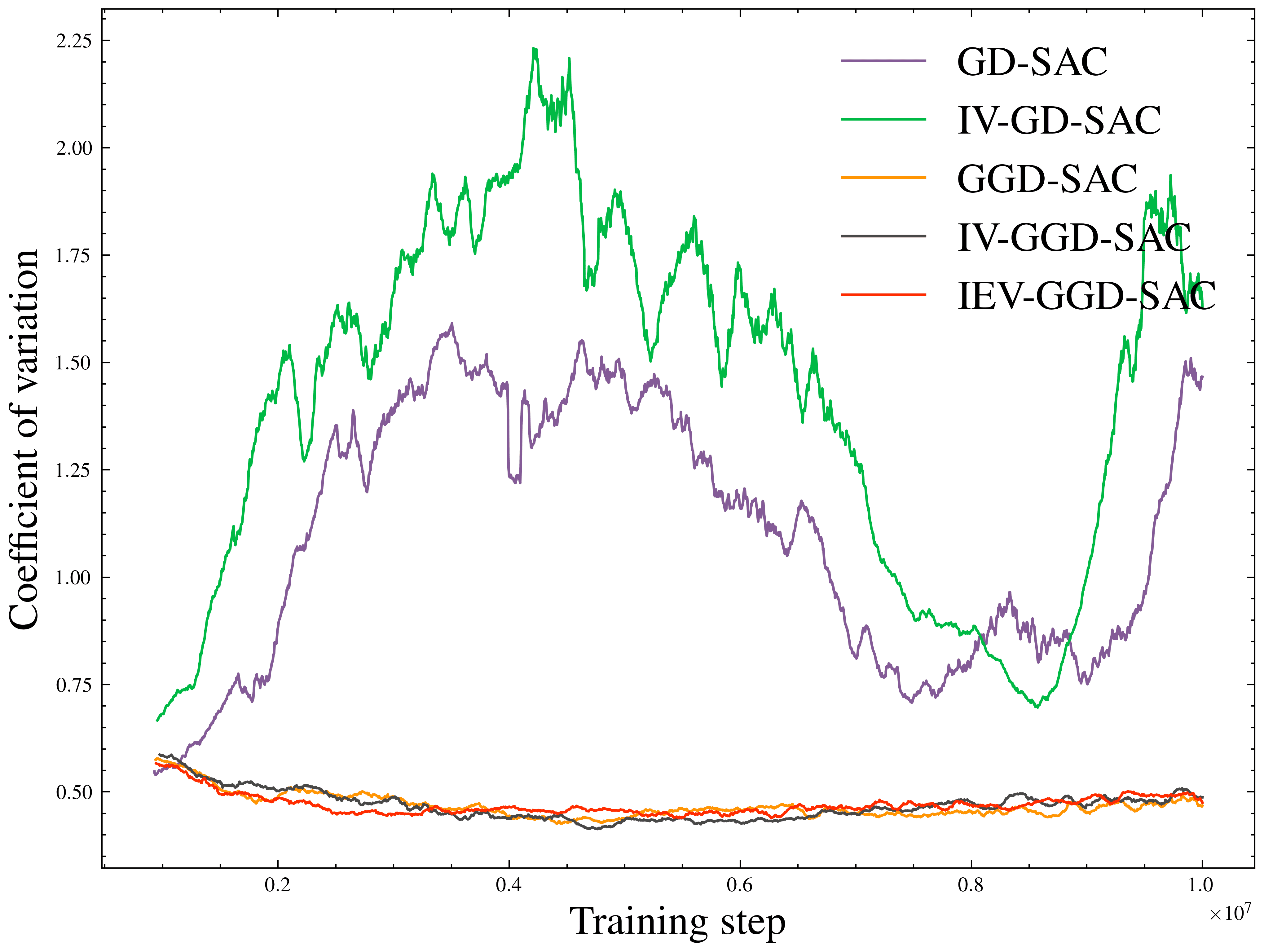}}
	\caption{
		Coefficients of variation of parameter estimates for SAC variants.
	} \label{fig:param-extended}
\end{figure}


\subsection{PPO on Other Environments} \label{apdx:ext:ppo-other}

\begin{figure}[H]
	\centering
	\subfloat[MountainCar-v0]{\includegraphics[width=.31\textwidth]{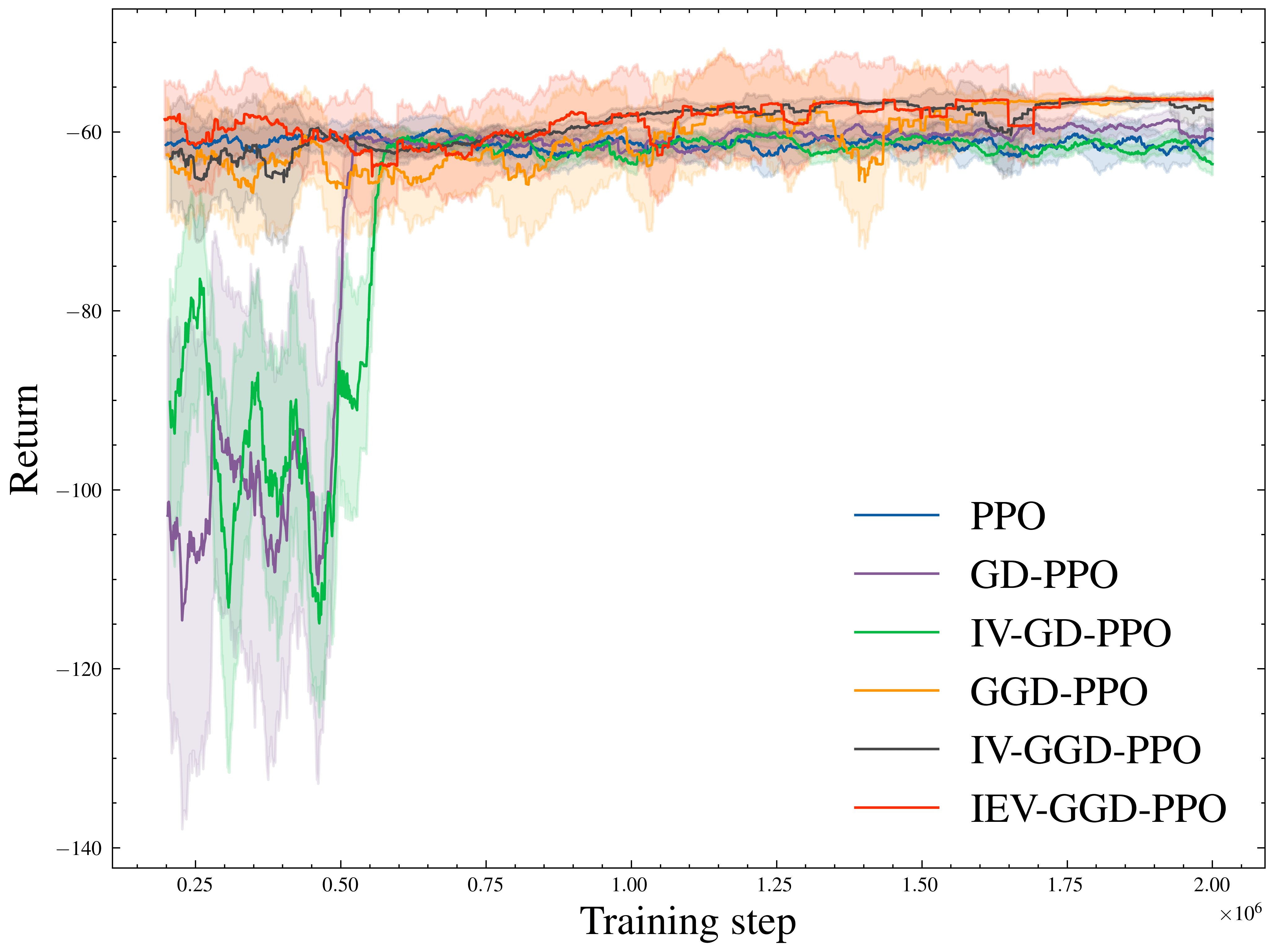}}
	\subfloat[Ant-v4]{\includegraphics[width=.31\textwidth]{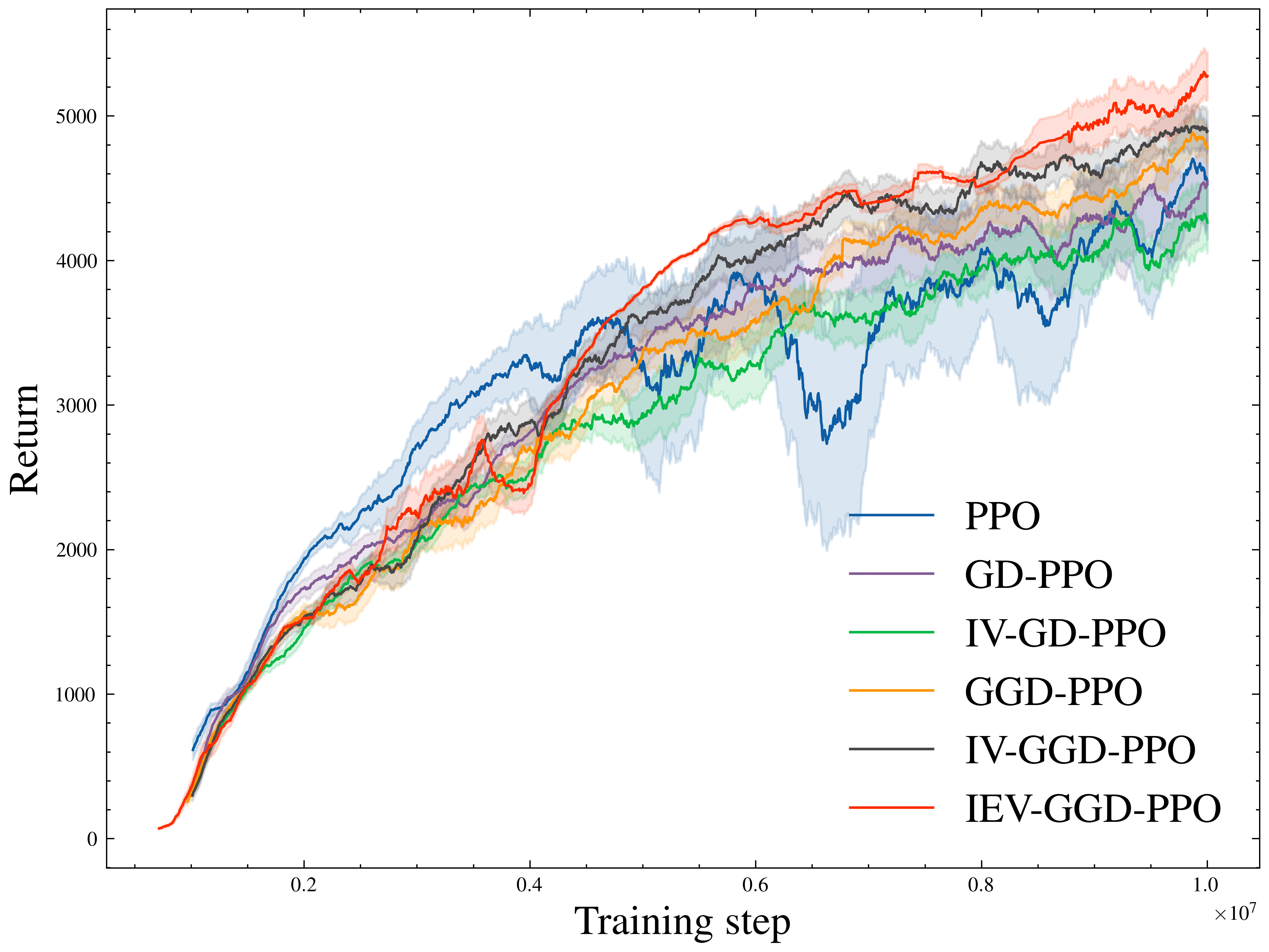}}
	\subfloat[Humanoid-v4]{\includegraphics[width=.31\textwidth]{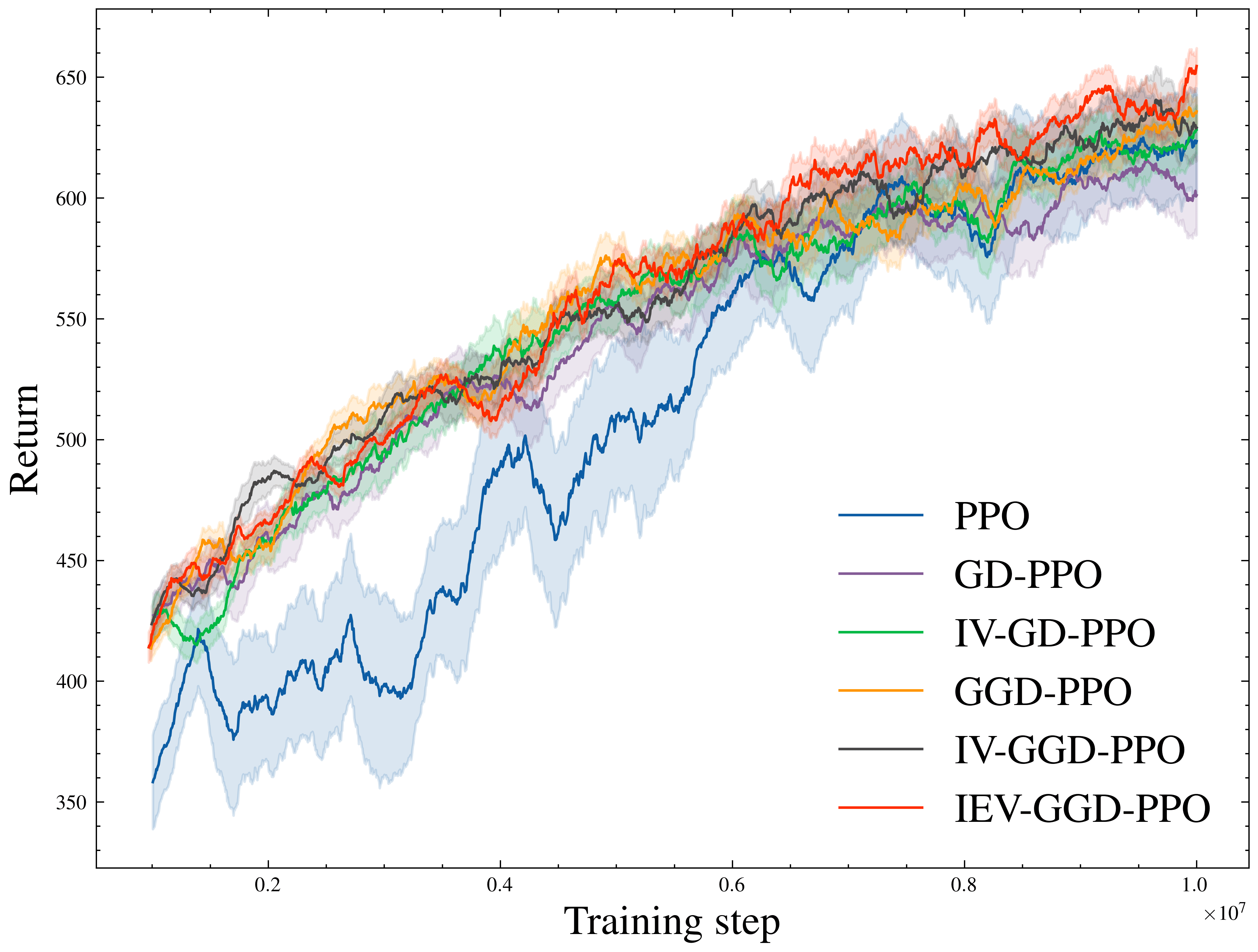}} \\
	\subfloat[Hopper-v4]{\includegraphics[width=.31\textwidth]{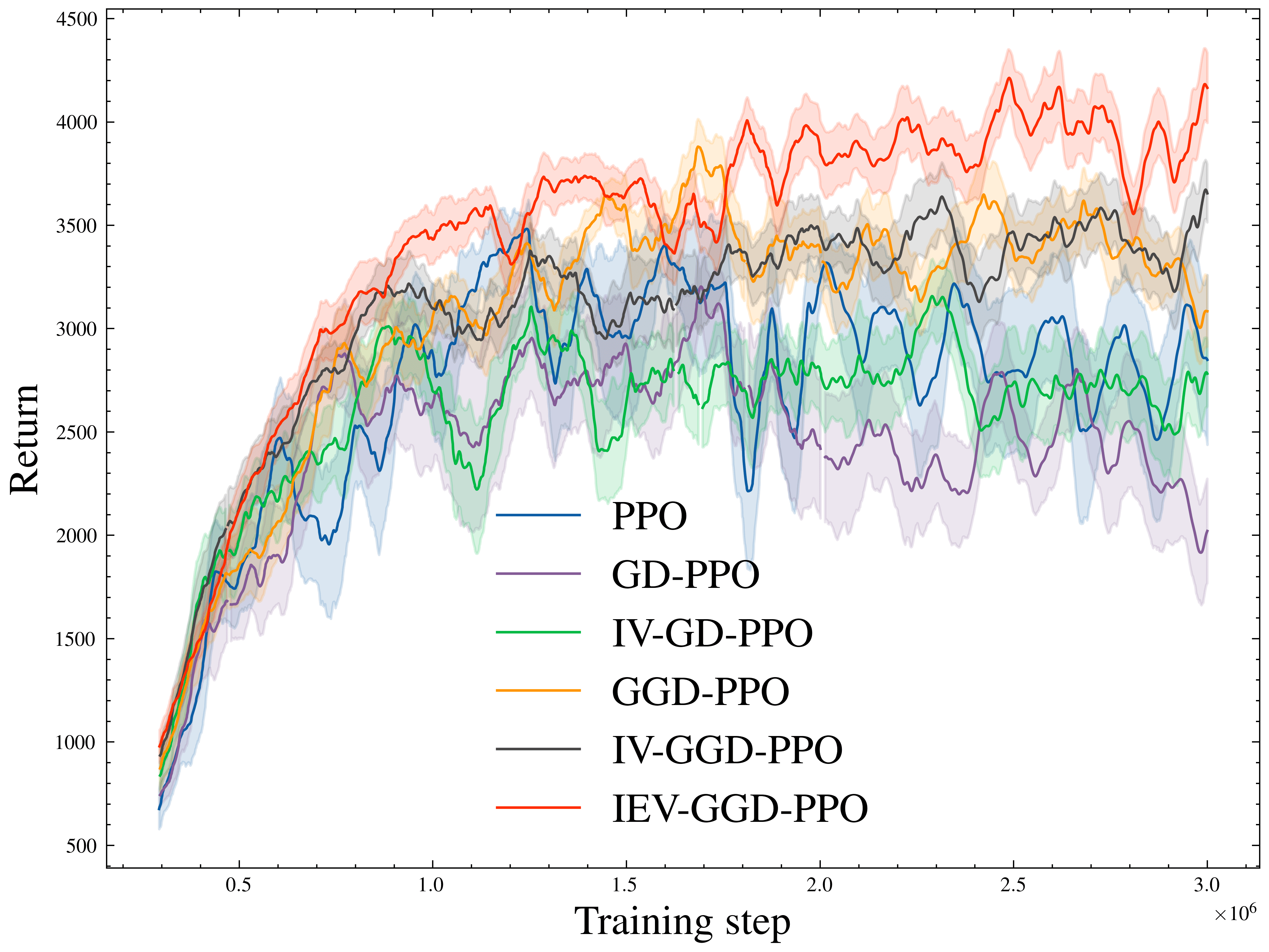}}
	\subfloat[Walker2D-v4]{\includegraphics[width=.31\textwidth]{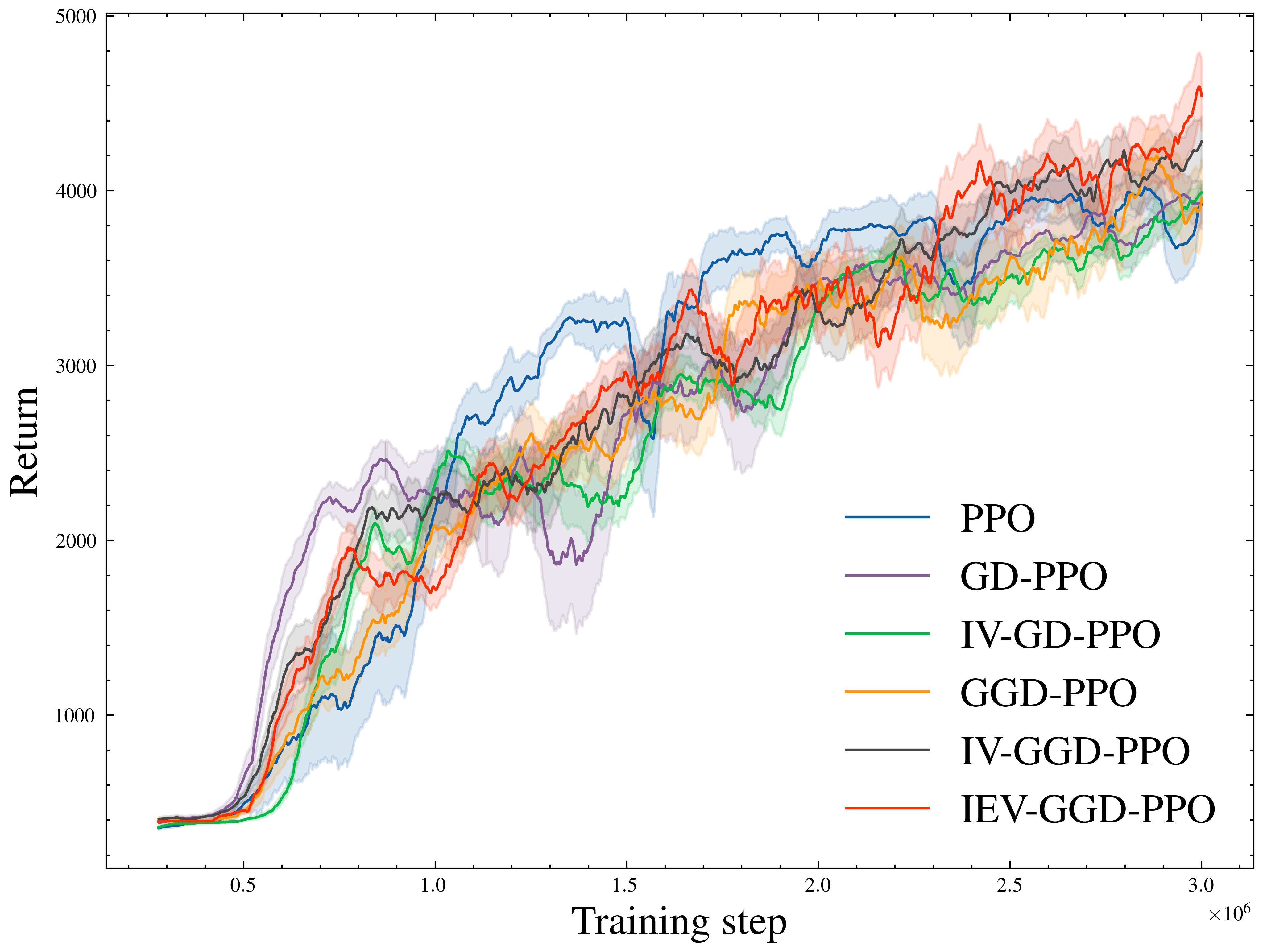}}
	\caption{
		Sample efficiency curves of PPO on remaining environments.
	} \label{fig:ppo-extended}
\end{figure}


\subsection{Parameter Estimation} \label{apdx:ext:beta-head}

\begin{figure}[H]
	\centering
	\subfloat[Ant-v4]{\includegraphics[width=.31\textwidth]{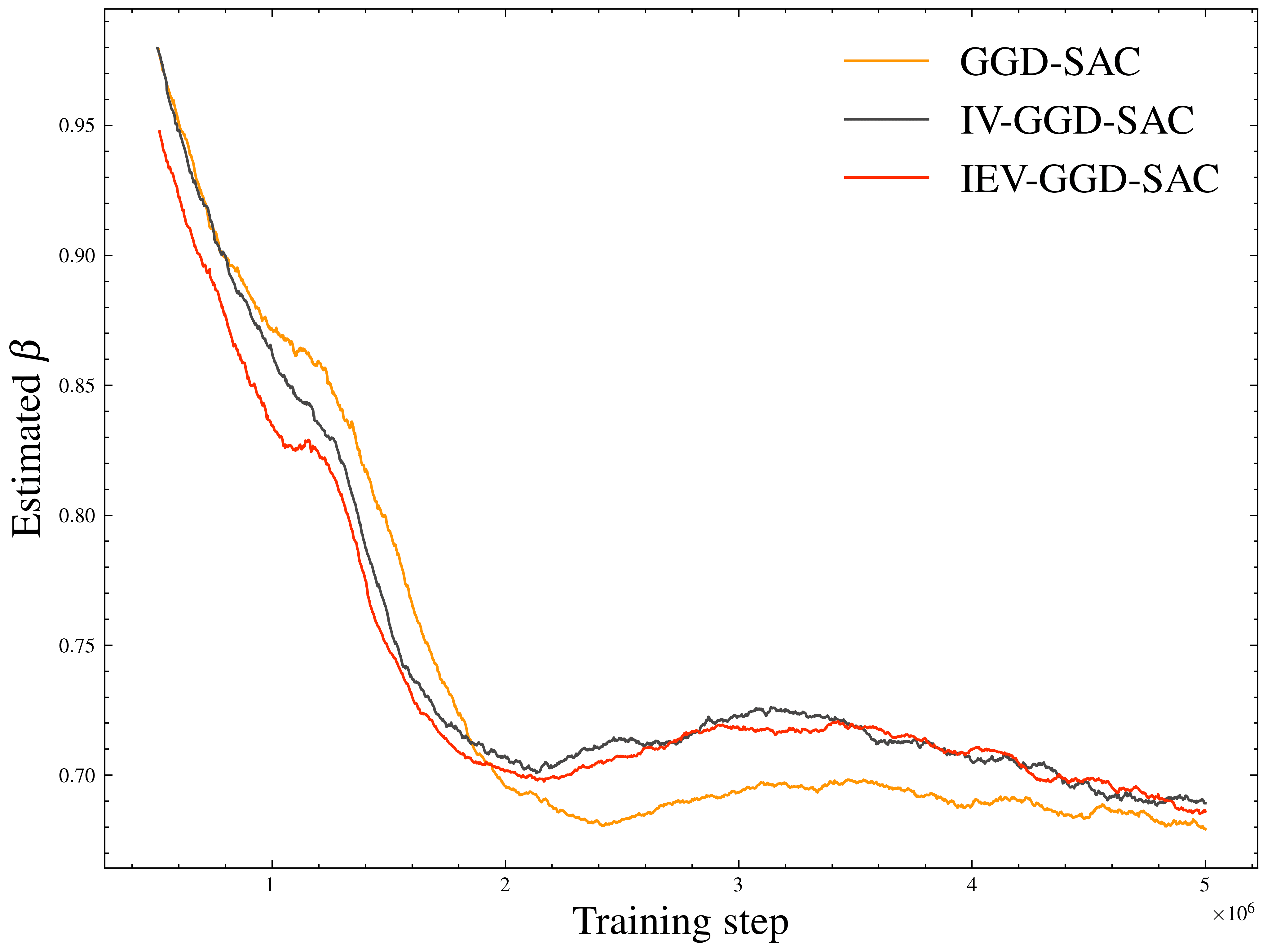}}
	\subfloat[HalfCheetah-v4]{\includegraphics[width=.31\textwidth]{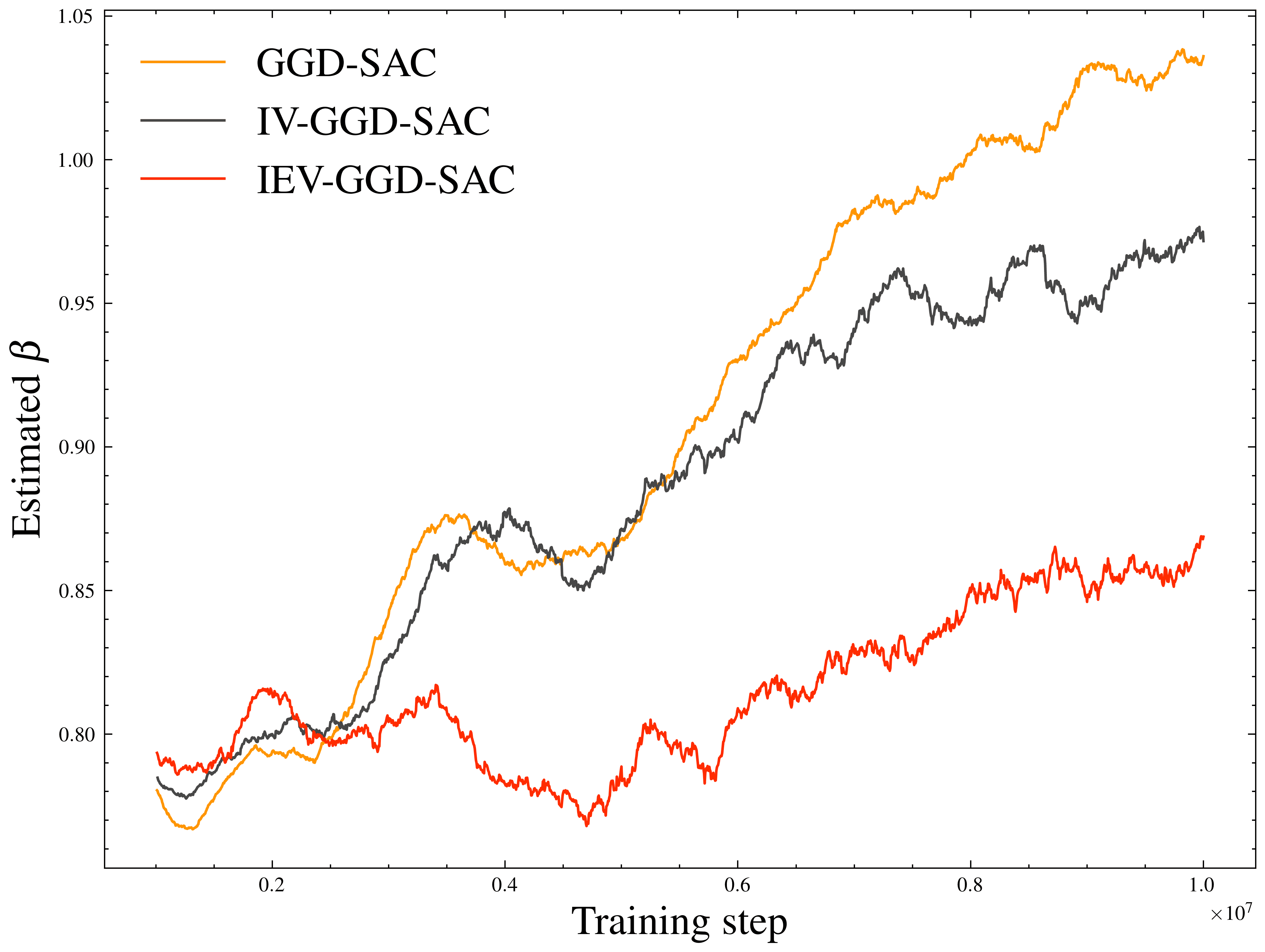}}
	\subfloat[Hopper-v4]{\includegraphics[width=.31\textwidth]{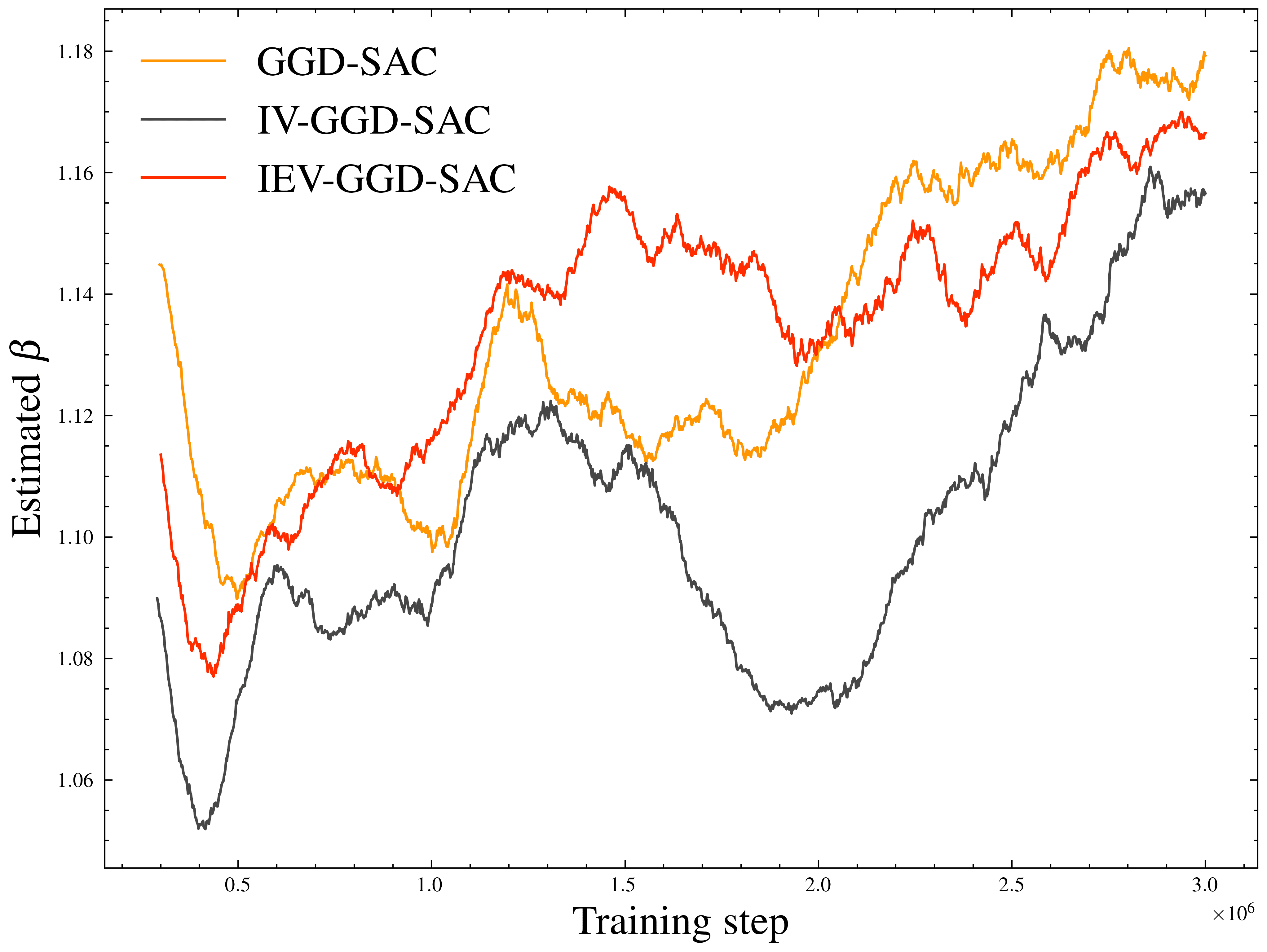}} \\
	\subfloat[Humanoid-v4]{\includegraphics[width=.31\textwidth]{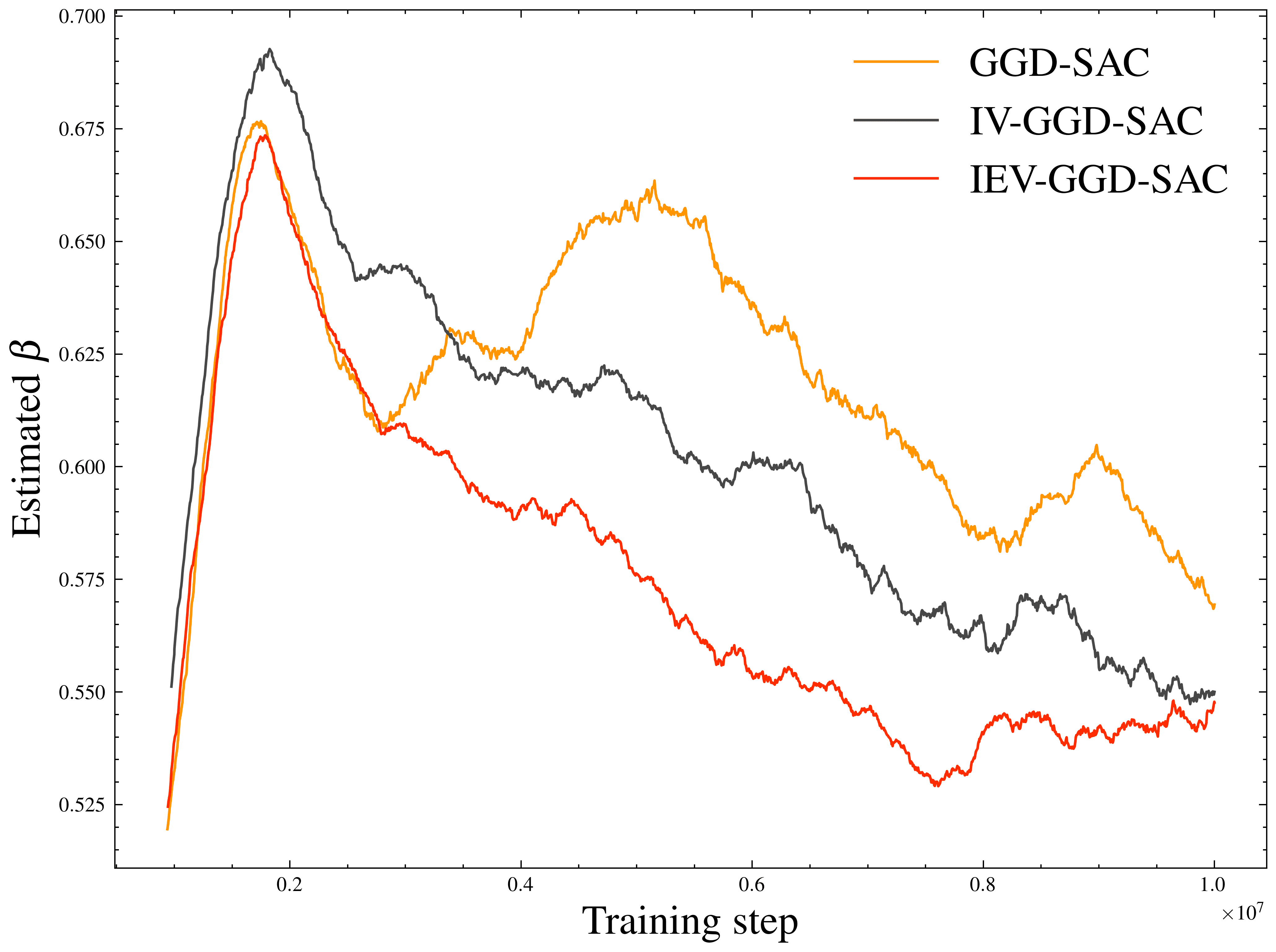}}
	\subfloat[Walker2D-v4]{\includegraphics[width=.31\textwidth]{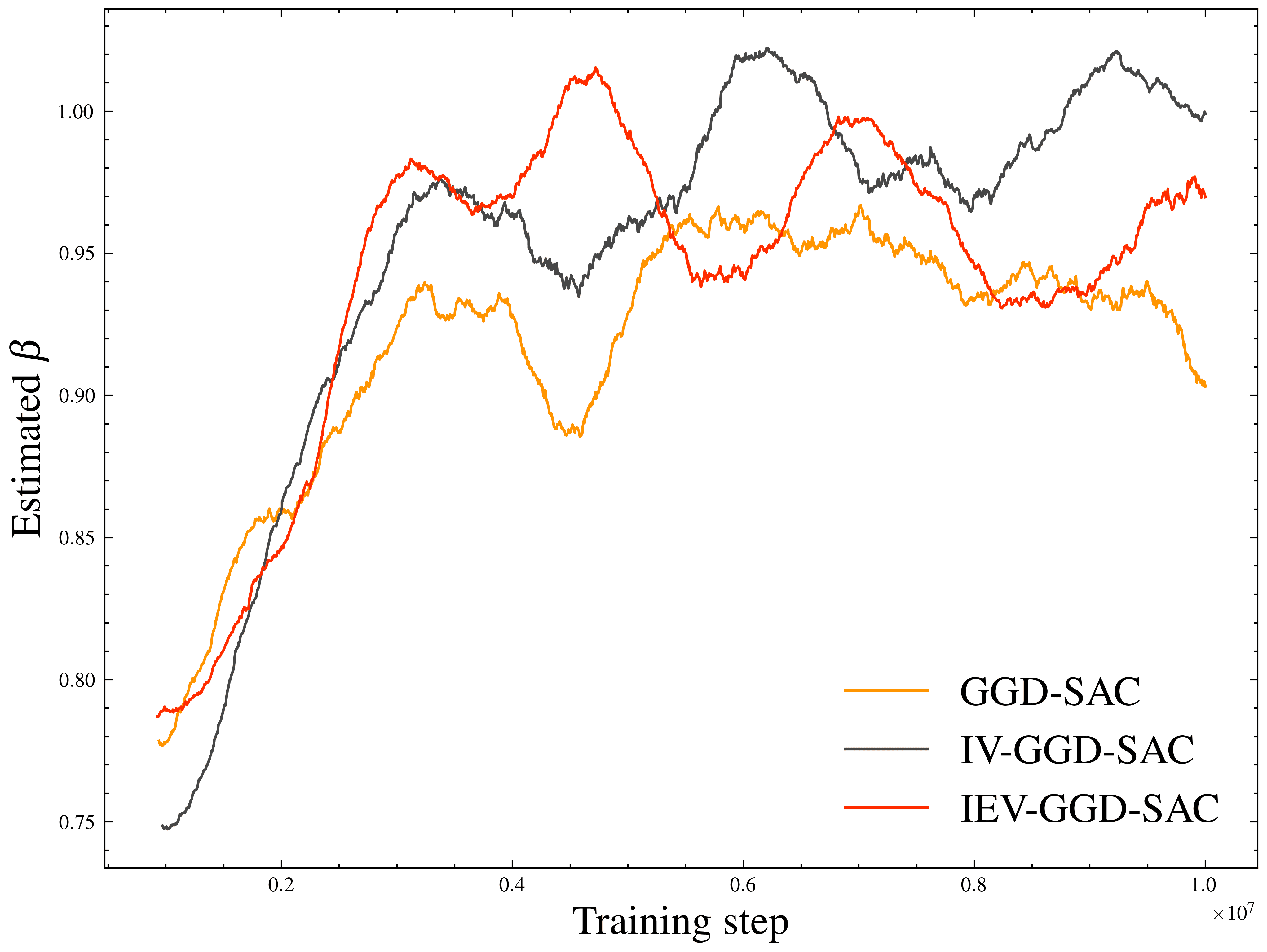}}
	\caption{
		Estimated $\beta$ for SAC variants.
	} \label{fig:beta-head}
\end{figure}

The aggregate shape estimates in~\cref{fig:beta-head} remain below the Gaussian value $\beta=2$ while varying across environments and methods.
This is consistent with, but does not by itself prove, the leptokurtic patterns visible in the pooled TD-error plots in~\cref{fig:tde-sac,fig:tde-sac2,fig:tde-ppo}.

\subsection{Extended Comparison} \label{apdx:ext:comp}

\begin{figure}[H]
	\centering
	\subfloat[Humanoid-v4]{\includegraphics[width=.31\textwidth]{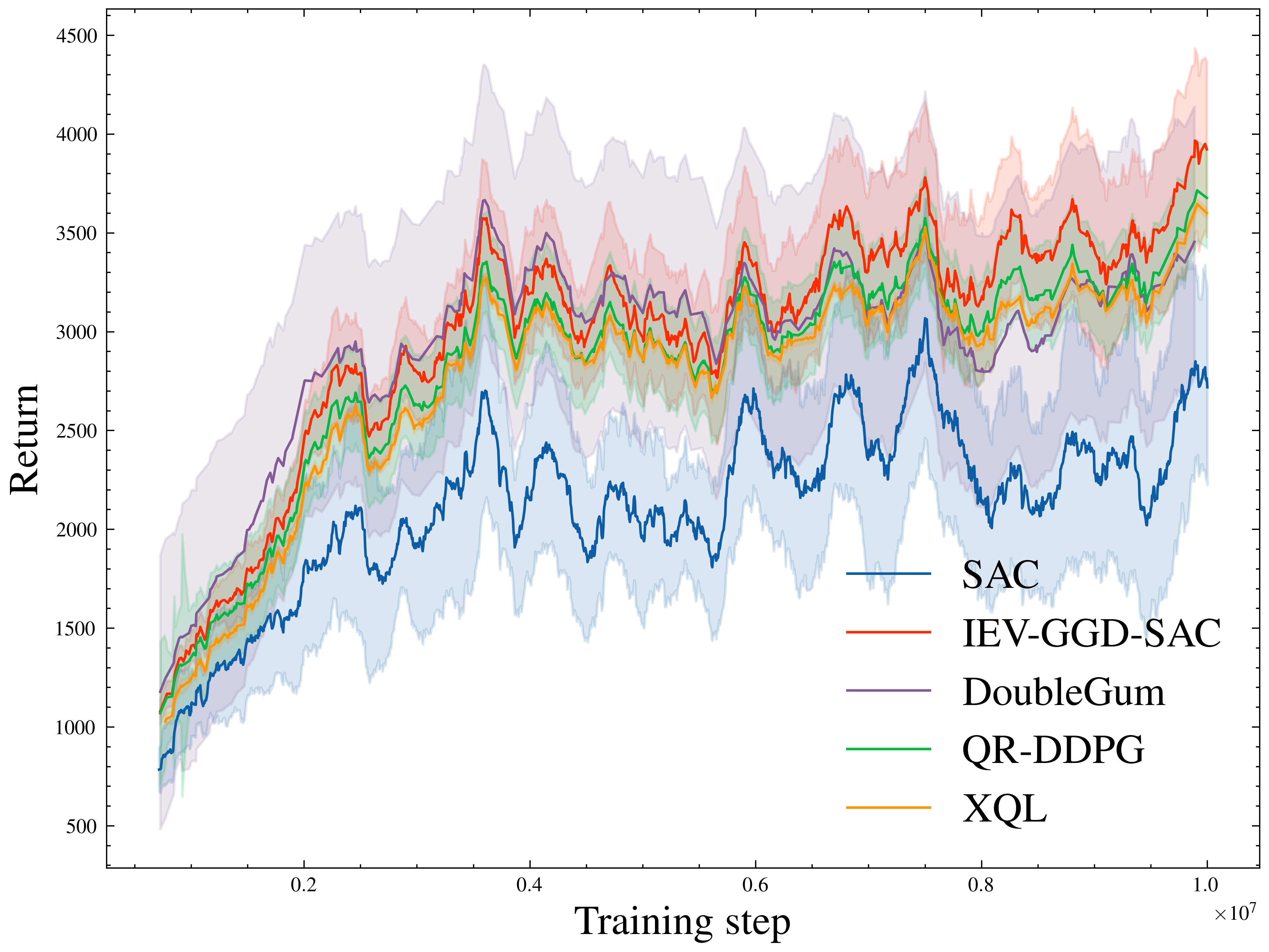}}
	\subfloat[Walker2D-v4]{\includegraphics[width=.31\textwidth]{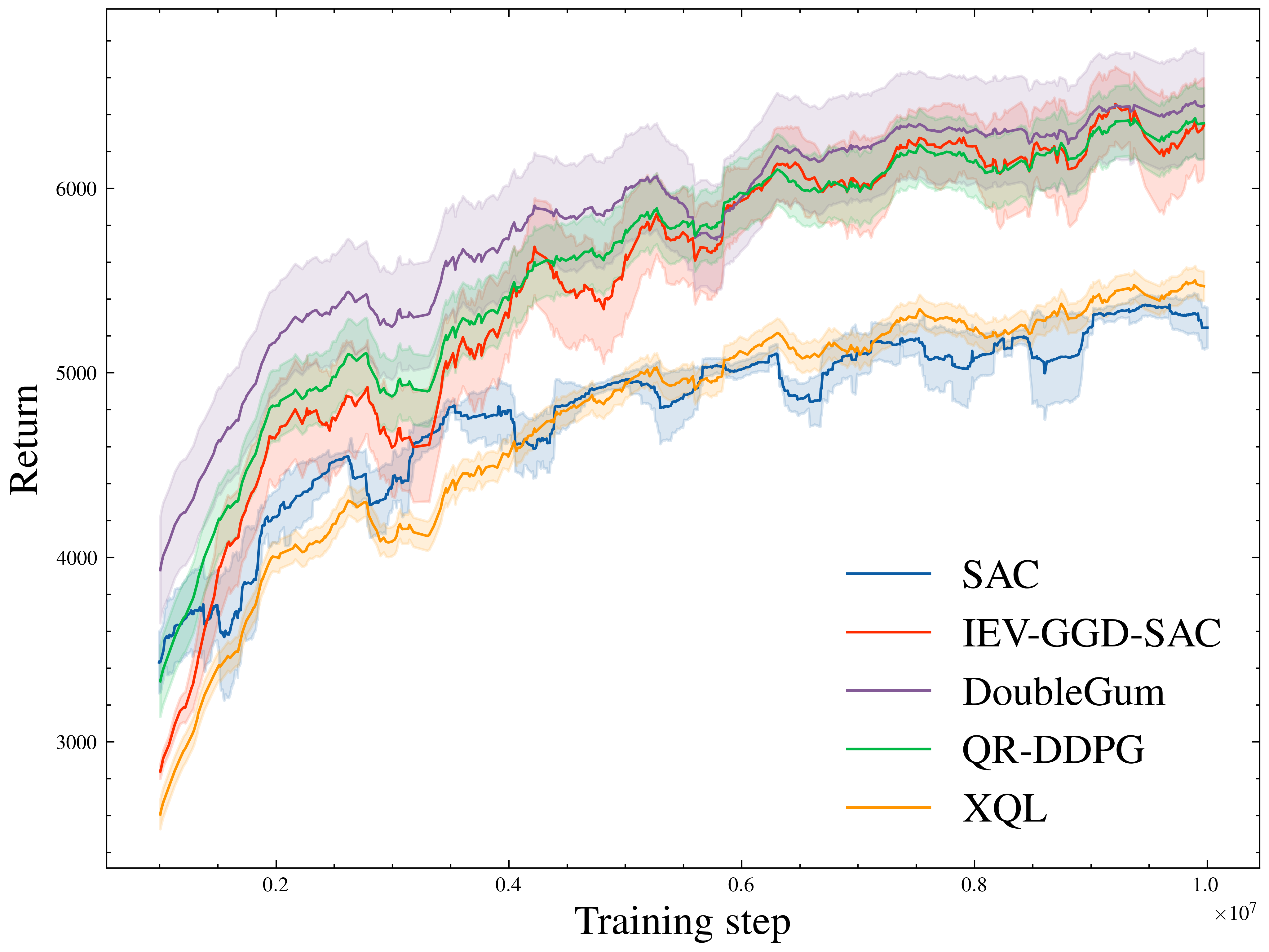}}
	\caption{
		Descriptive comparison with representative alternatives on remaining environments.
	} \label{fig:comp-extended}
\end{figure}

This section extends the comparison in~\cref{fig:comp} to the remaining MuJoCo environments.
\cref{fig:comp-extended} shows that relative performance also varies across the remaining environments.

\section{Proofs} \label{apdx:proof}

\subsection{Of Theorem~\ref{thm:ggd-nll}} \label{apdx:proof:ggd-nll}

\begin{proof}
	For $0<\beta\le2$, $k_\beta(x)=\exp(-|x|^\beta)$ is the characteristic function of a centered symmetric $\beta$-stable law~\citep{bochner1937stable,ushakov2011selected}.
	Every characteristic function is positive definite by Bochner's theorem; positive rescaling and composition with $x/\alpha$ preserve that property.

	Likelihood normalization is a separate calculation valid for every $\beta>0$:
	\begin{align*}
		\int_{-\infty}^{\infty} e^{-(|x|/\alpha)^\beta}\,dx
		 & =2\int_0^\infty e^{-(x/\alpha)^\beta}\,dx \\
		 & =\frac{2\alpha}{\beta}\Gamma(1/\beta).
	\end{align*}
	Multiplication by $\beta/[2\alpha\Gamma(1/\beta)]$ therefore gives a normalized density.
	For finite $\delta$, $\alpha>0$, and $\beta>0$, every term of its exact NLL is finite.
	This normalization statement does not imply convexity or optimization stability.
\end{proof}


\subsection{Of Theorem~\ref{thm:ssd}} \label{apdx:proof:ssd}

\begin{proof}
	Under our notation, $X\sim\ggd(0,2^{1/\beta}\alpha,\beta)$ has density
	\begin{equation*}
		\frac{\beta}{2^{(\beta+1)/\beta}\alpha\Gamma(1/\beta)}
		\exp\!\left(-\frac{|x|^\beta}{2\alpha^\beta}\right).
	\end{equation*}
	This is exactly the zero-location parameterization used in Proposition~3 of \citet{dytso2018analytical}.
	That proposition gives~\cref{eq:ssd} whenever $\beta_1\le\beta_2$, which proves the stated result.
	The mapping also shows why the cited theorem does not directly order our implemented family with the GGD scale fixed to one for every $\beta$.
\end{proof}


\subsection{Of Proposition~\ref{pro:mbbe}} \label{apdx:proof:mbbe}

\begin{proof}
	For the unbiased sample variance $S^2$, the fourth-moment identity gives
	\begin{equation*}
		\sV[S^2]=\frac{\sigma^4}{n}\left(\kappa+\frac{2n}{n-1}\right).
	\end{equation*}
	The MSE of $cS^2$ is
	\begin{equation*}
		\mse(cS^2)=c^2\sV[S^2]+(c-1)^2\sigma^4.
	\end{equation*}
	Differentiating with respect to $c$ and setting the result to zero gives
	\begin{equation*}
		c^*=\frac{\sigma^4}{\sV[S^2]+\sigma^4}
		=\left(\frac{\kappa}{n}+\frac{n+1}{n-1}\right)^{-1}.
	\end{equation*}
	The second derivative $2\sV[S^2]+2\sigma^4$ is positive.
	Substitution of $c^*$ yields
	\begin{equation*}
		\frac{\sV[S^2]}{\mse(c^*S^2)}
		=1+\frac{\kappa}{n}+\frac{2}{n-1}.
	\end{equation*}
\end{proof}


\section{Implementation Details} \label{apdx:impl}

\begin{algorithm}[t]
	\caption{GGD-aware Critic Update} \label{alg:ggtde}
	\begin{algorithmic}[1]
		\Require Batch $\gB$, critics $\{Q_k\}_{k=1}^K$ with $\beta$ outputs, temperature $\lambda$, variance floor $\epsilon$, minimum effective batch size $m$
		\For{each transition $t \in \gB$}
		\For{$k=1,\ldots,K$}
		\State $\delta_{t,k}\gets T_{t,k}-Q_k^\mu(s_t,a_t)$
		\State $\hat\beta_{t,k}\gets\operatorname{softplus}(Q_k^\beta(s_t,a_t))$
		\State $\ell^{\mathrm{sur}}_{t,k}\gets \hat\beta_{t,k}|\delta_{t,k}|-\log\hat\beta_{t,k}+\log\Gamma(1/\hat\beta_{t,k})$
		\EndFor
		\State $\omega^{\mathrm{RA}}_{t,k}\gets\hat\beta_{t,k}/\sum_{j=1}^K\hat\beta_{t,j},\quad k=1,\ldots,K$
		\State $v_t\gets K^{-1}\sum_k(\delta_{t,k}-\bar\delta_t)^2$
		\State $\hat\kappa_t\gets\operatorname{SampleExcessKurtosis}(\{\delta_{t,k}\}_{k=1}^K)$
		\State $\widetilde s_t^2\gets\max\!\left\{\epsilon,\left(\frac{\hat\kappa_t}{K}+\frac{K+1}{K-1}\right)^{-1}v_t\right\}$
		\EndFor
		\State Numerically select $\xi\ge0$ to target effective size $\min\{|\gB|-1,m\}$
		\State $u_t\gets(\widetilde s_t^2+\xi)^{-1}$ and $\bar u_t\gets u_t/\sum_\tau u_\tau$
		\State $\Ls\gets |\gB|^{-1}\!\left(\sum_{t,k}\omega^{\mathrm{RA}}_{t,k}\ell^{\mathrm{sur}}_{t,k}+\lambda\sum_t\bar u_t\sum_k|\delta_{t,k}|\right)$
		\State Update critic parameters via $\nabla\Ls$
	\end{algorithmic}
\end{algorithm}

We apply softplus, a smooth approximation to ReLU~\citep{dugas2000incorporating}, to enforce a positive $\hat\beta$.
With $\alpha=1$, the exact GGD NLL (up to constants) contains $|\delta|^\beta-\log\beta+\log\Gamma(1/\beta)$.
The released implementation replaces the exponent term by a linear product and evaluates
\begin{equation*}
	\ell_\text{GGD-sur}(\delta,\hat\beta)
	=|\delta|\hat\beta-\log\hat\beta+\log\Gamma(1/\hat\beta).
\end{equation*}
This is a different surrogate objective rather than an algebraically equivalent rewrite or a controlled numerical approximation.
Accordingly, \cref{thm:ggd-nll} characterizes the exact GGD density kernel and likelihood, not the optimization properties of this surrogate.

Risk weights are normalized across the $K$ critics for each transition.
BIEV statistics are computed across the same critic dimension, whereas inverse BIEV weights are normalized across the minibatch.
The corrected variance is floored by $\epsilon$, and $\xi$ is selected by numerical search to target the configured effective size subject to optimizer tolerance.
The released implementation uses a bias-corrected sample excess kurtosis without clipping and contains no special $\hat\beta>2.5$ rescaling rule.

The reported GGD+BIV and GGD+BIEV variants use mean absolute error (MAE) in the regularizer; the mean squared error alternative is evaluated in~\cref{apdx:abl:biev}.
The temperature sensitivity study appears in~\cref{apdx:abl:lambda}.


\section{Experimental Details} \label{apdx:exp}

The released implementation, dependency specification, and training configurations are available at \url{https://github.com/ait-lab/ggtde}.

To broaden the empirical scope of our study to discrete control environments, we augment the baseline settings with stochastic perturbations to simulate more realistic and challenging dynamics.
In CartPole-v1 and MountainCar-v0, we introduce uniform noise to manipulate forces or torques.
Furthermore, in LunarLander-v2, wind dynamics are activated.
For further details on how wind affects LunarLander-v2 dynamics, refer to the official documentation\footnote{\url{https://gymnasium.farama.org/environments/box2d/lunar_lander}}.

\begin{figure}[t]
	\centering
	\subfloat[LunarLander-v2]{\includegraphics[width=.65\textwidth]{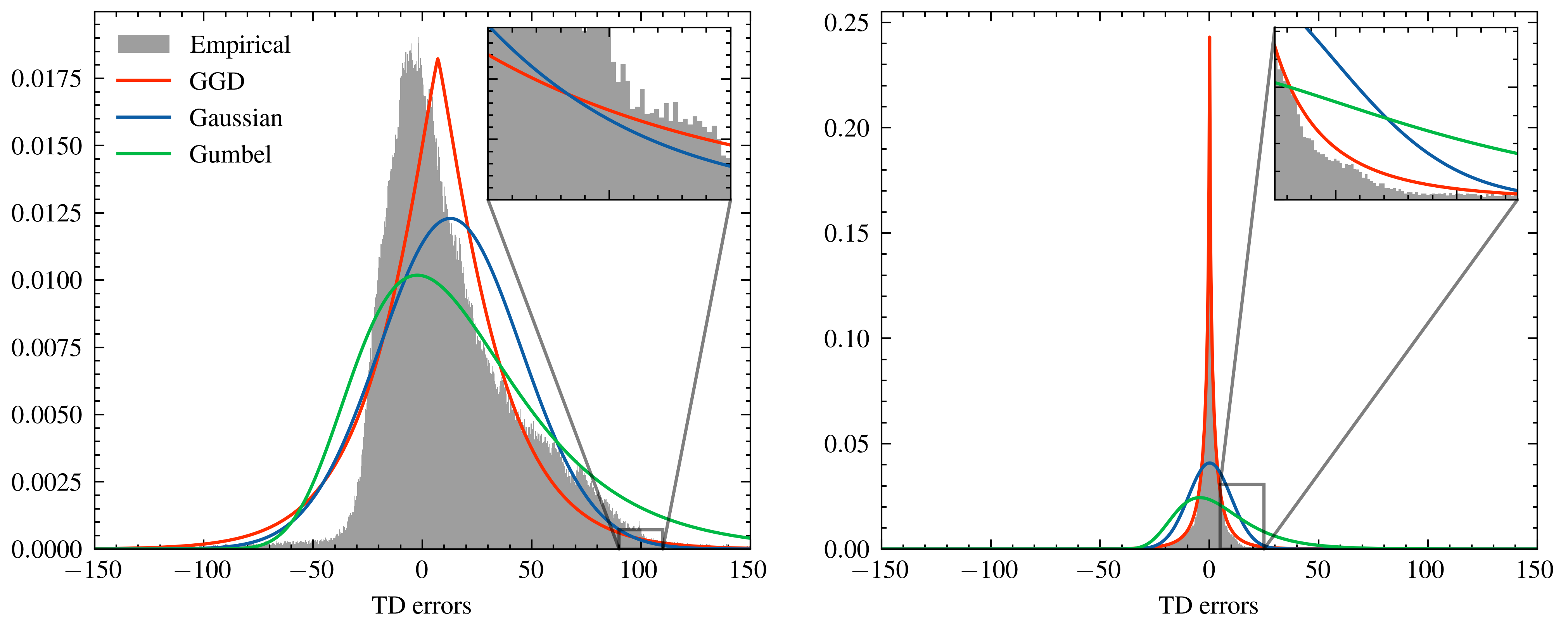}} \\
	\subfloat[MountainCar-v0]{\includegraphics[width=.65\textwidth]{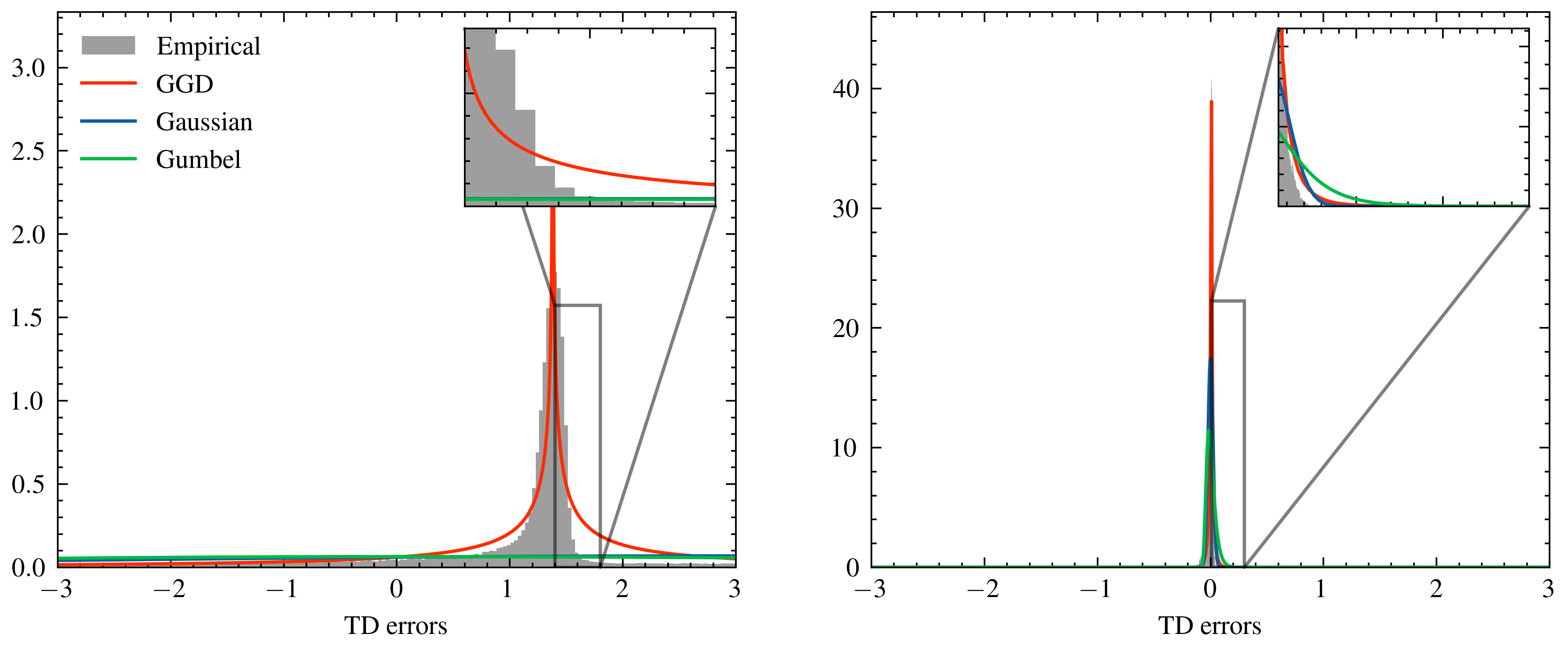}}
	\caption{
		TD error plots of PPO on original environments without additional noise.
	} \label{fig:tde-ppo-orig}
\end{figure}

\begin{figure}[t]
	\centering
	\subfloat[CartPole-v1]{\includegraphics[width=.31\textwidth]{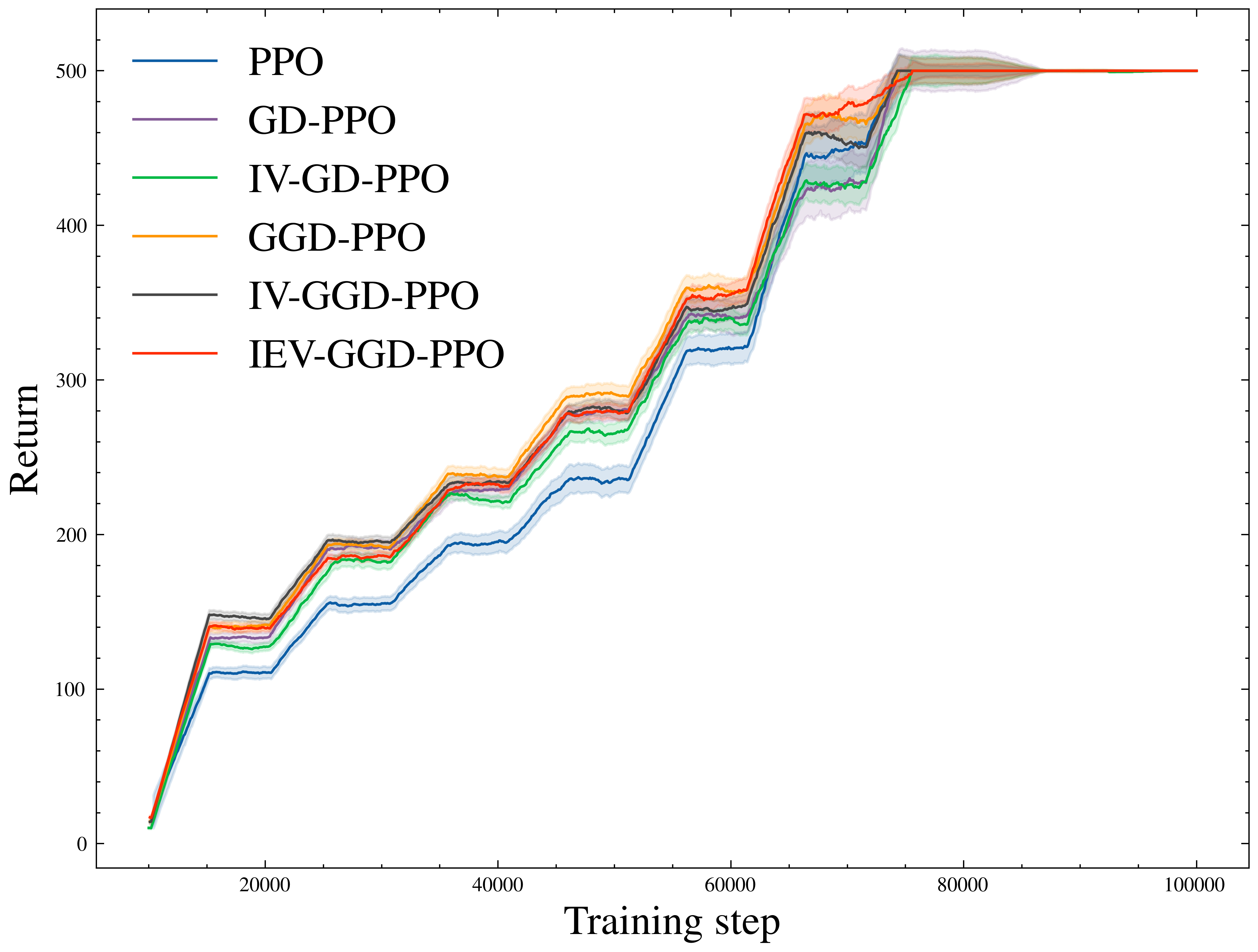}}
	\subfloat[LunarLander-v2]{\includegraphics[width=.31\textwidth]{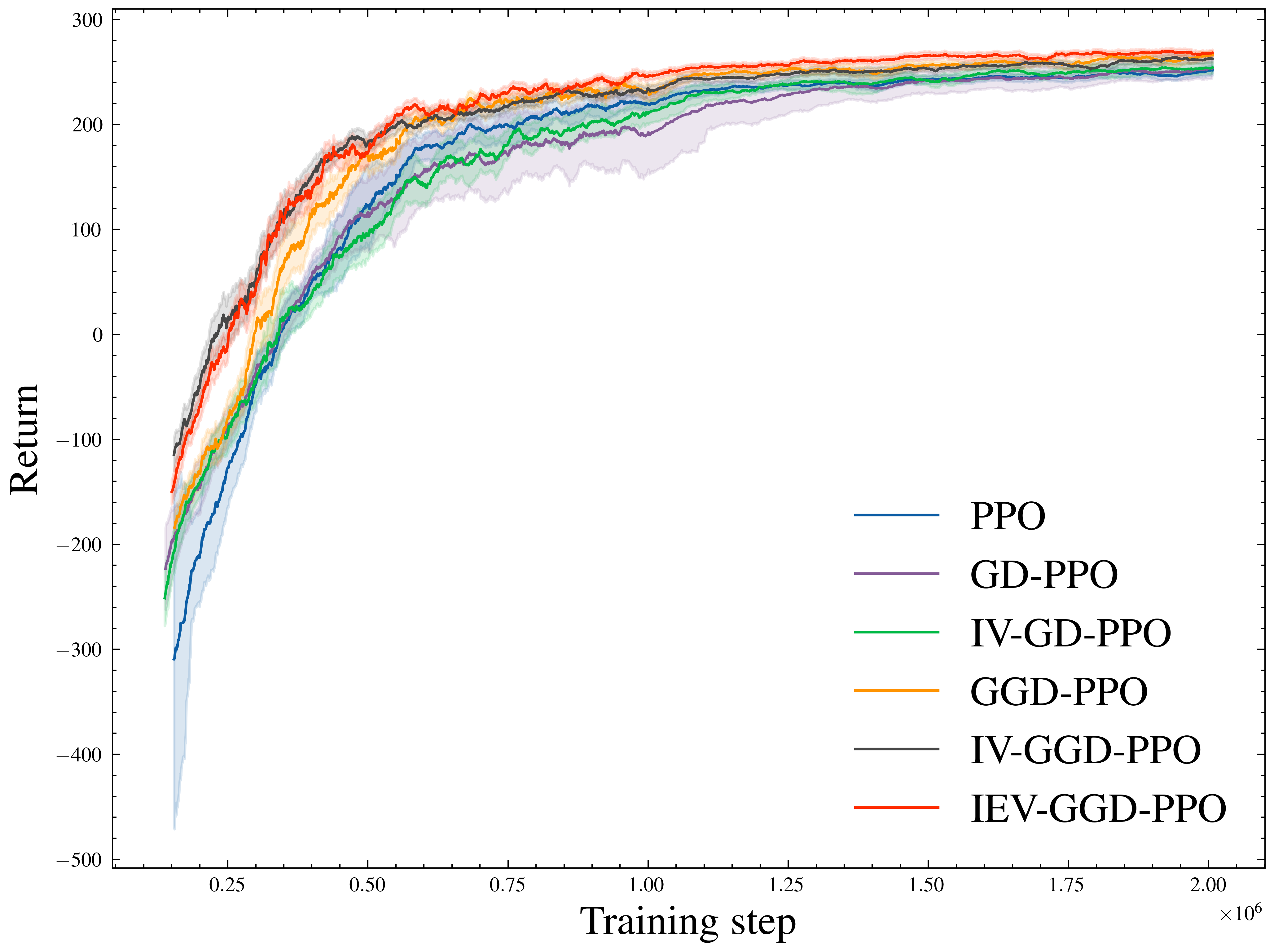}}
	\subfloat[MountainCar-v0]{\includegraphics[width=.31\textwidth]{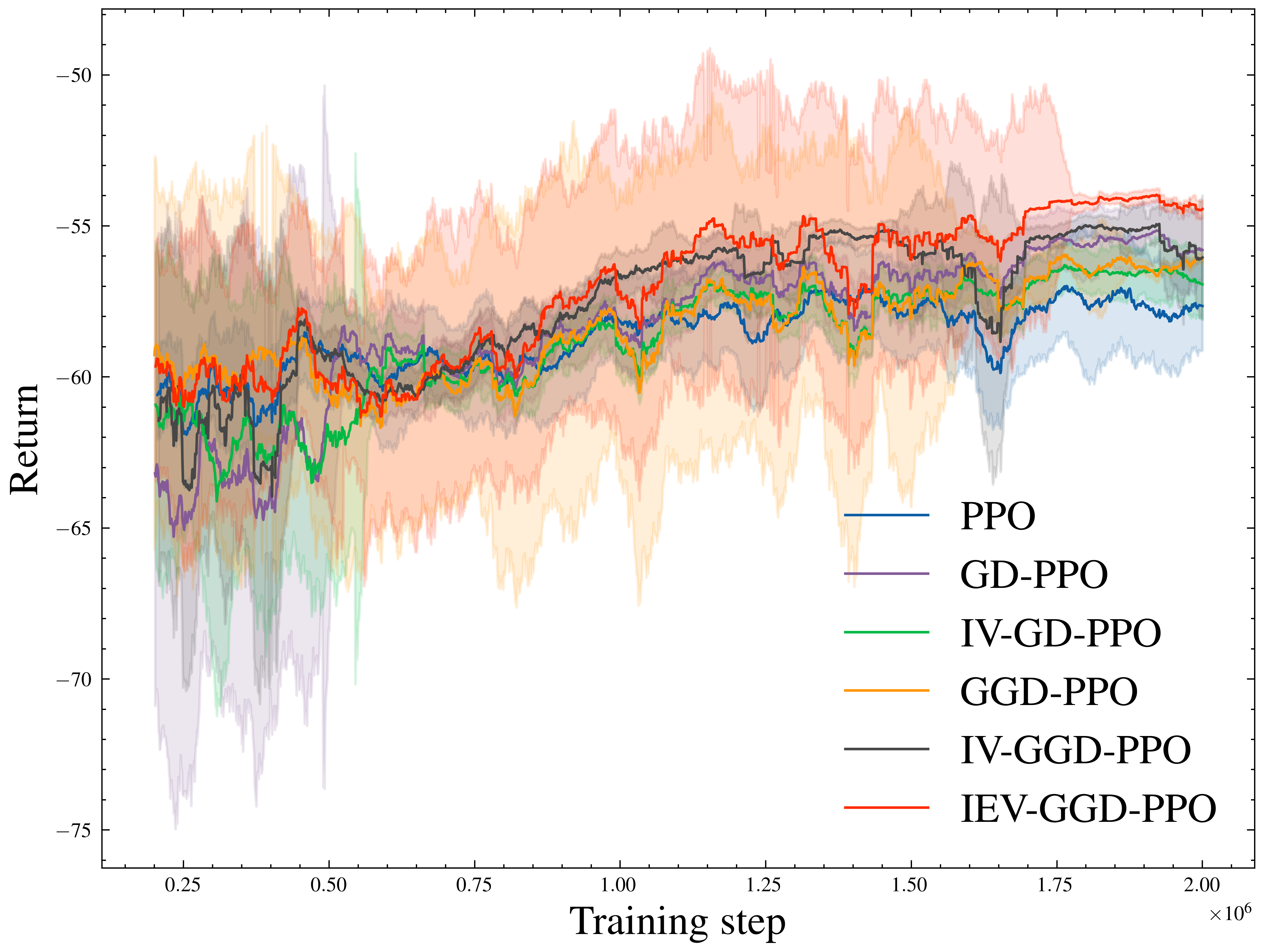}}
	\caption{
		Sample efficiency curves of PPO on original environments without additional noise.
	} \label{fig:abl-orig-env}
\end{figure}

We also report a limited check on the original environments without injected noise.
The pooled TD-error plots in~\cref{fig:tde-ppo-orig} remain visibly non-Gaussian in the reported tasks, while the performance differences in~\cref{fig:abl-orig-env} are smaller and mixed.
These observations do not isolate an interaction between perturbation strength and the GGD-inspired surrogate.

\section{Ablation studies} \label{apdx:ablation}

\begin{figure}[t]
	\centering
	\subfloat[Ant-v4]{\includegraphics[width=.31\textwidth]{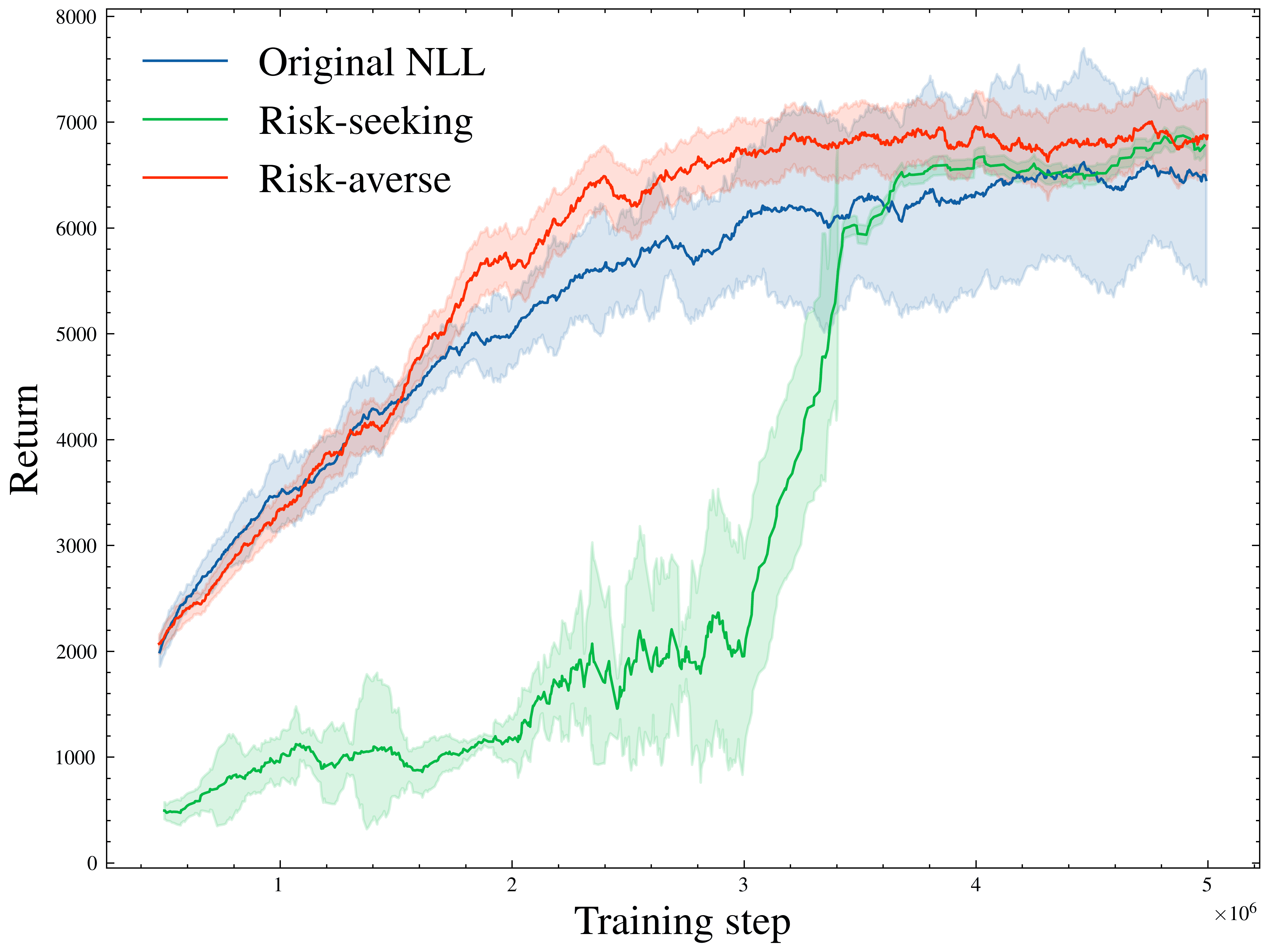}}
	\subfloat[HalfCheetah-v4]{\includegraphics[width=.31\textwidth]{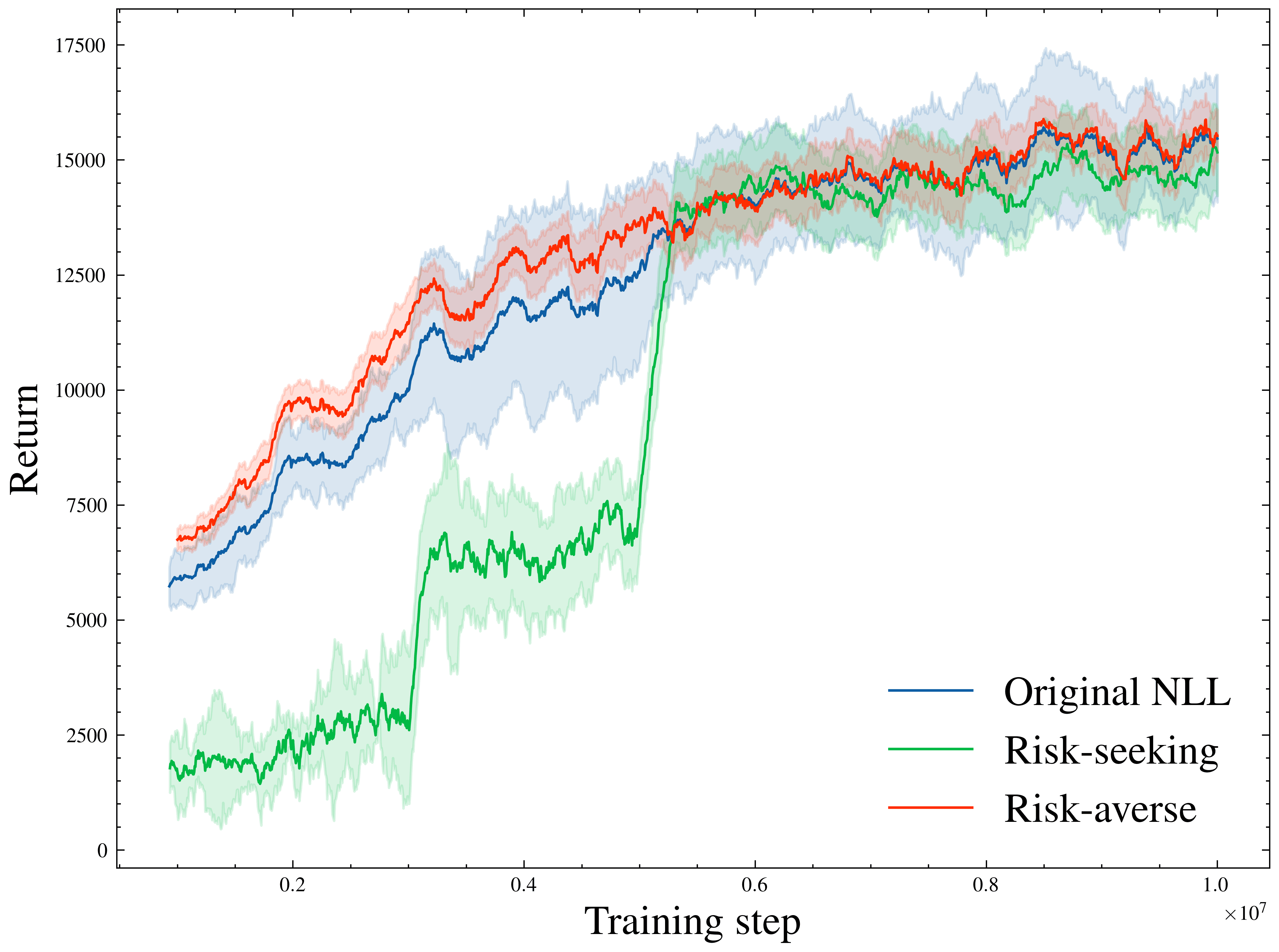}}
	\subfloat[Hopper-v4]{\includegraphics[width=.31\textwidth]{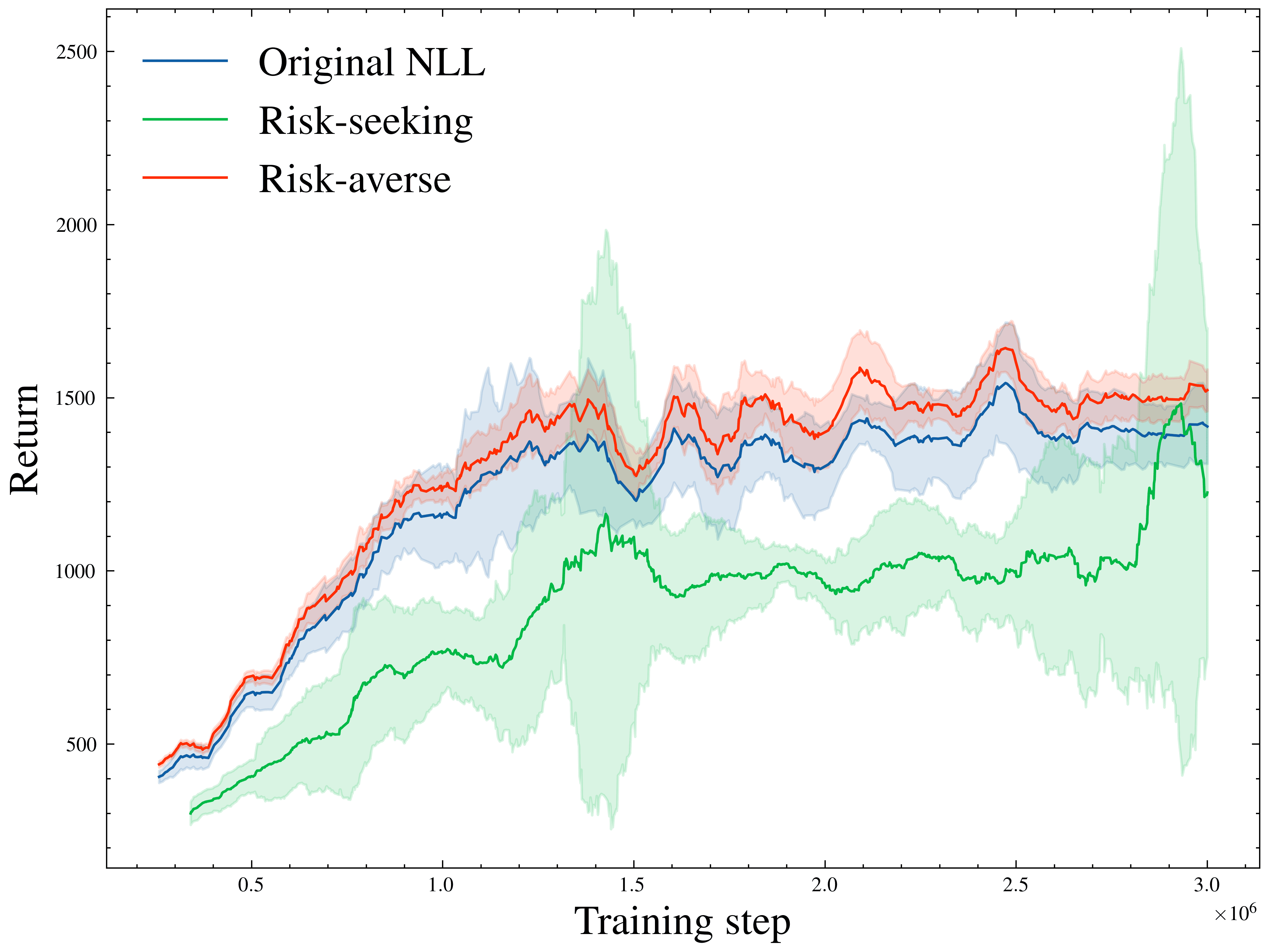}}
	\caption{
		Ablation study on risk-averse weighting.
	} \label{fig:abl-raw}
\end{figure}

\begin{figure}[t]
	\centering
	\subfloat[Ant-v4]{\includegraphics[width=.31\textwidth]{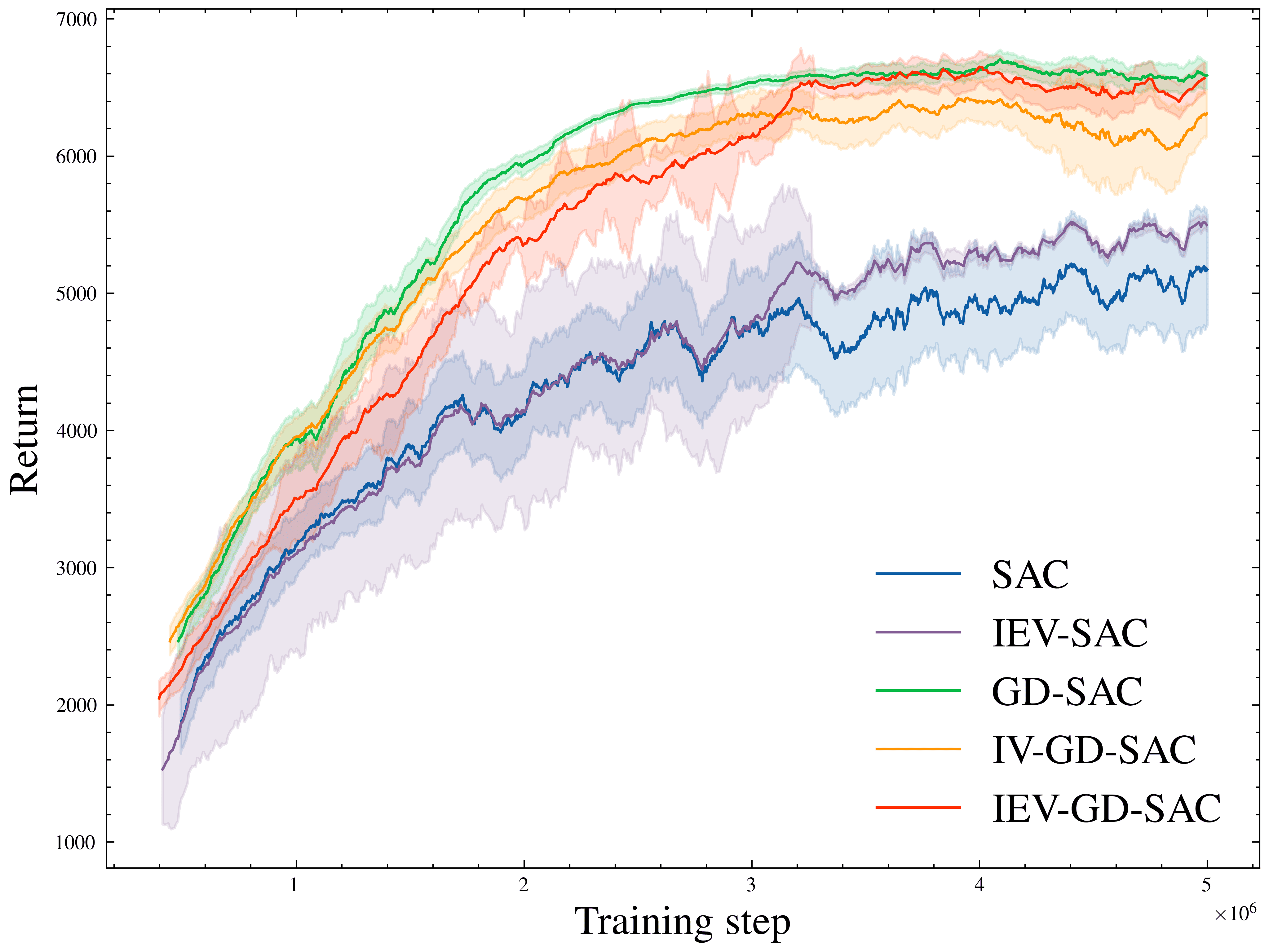}}
	\subfloat[HalfCheetah-v4]{\includegraphics[width=.31\textwidth]{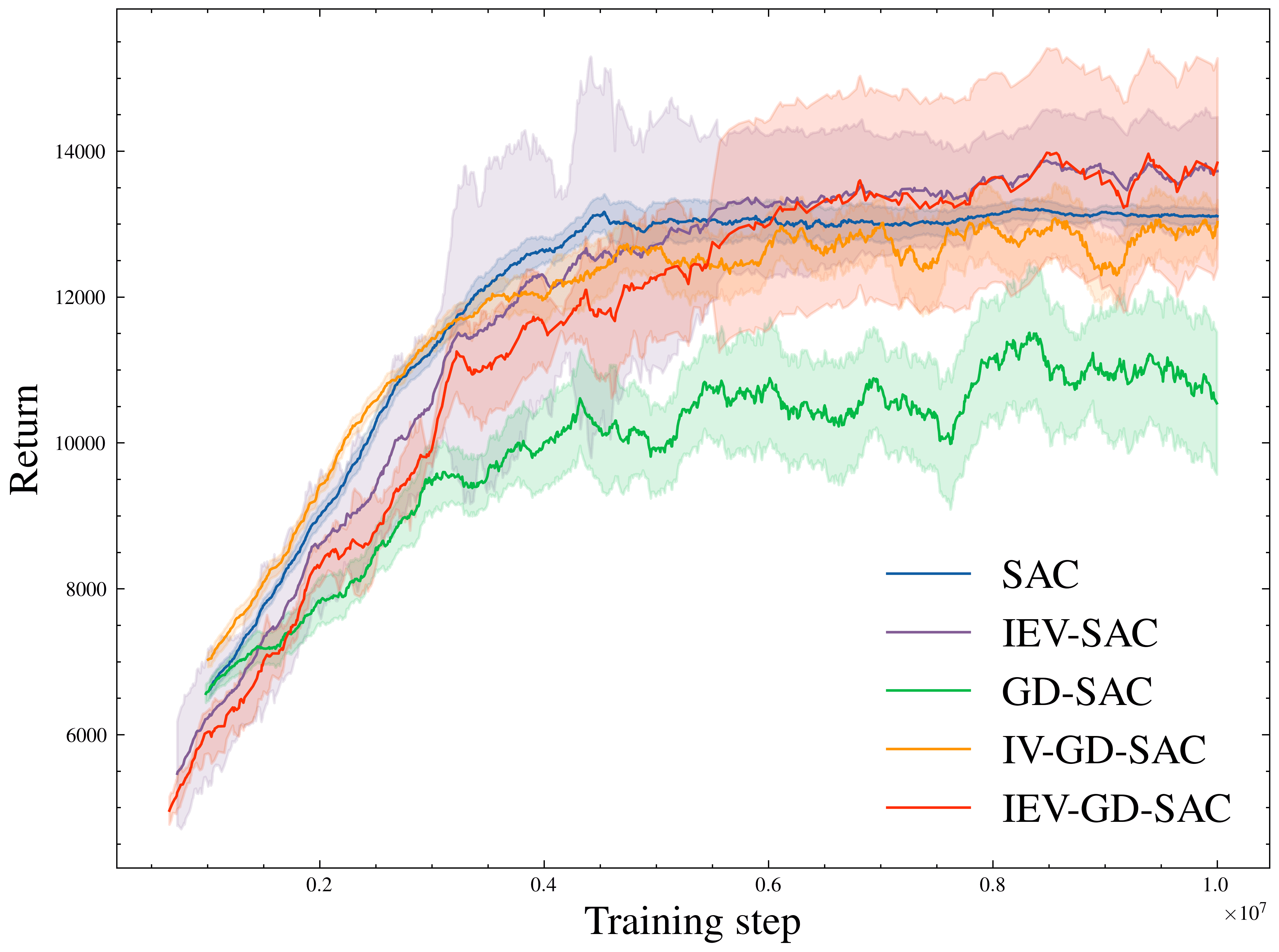}}
	\subfloat[Hopper-v4]{\includegraphics[width=.31\textwidth]{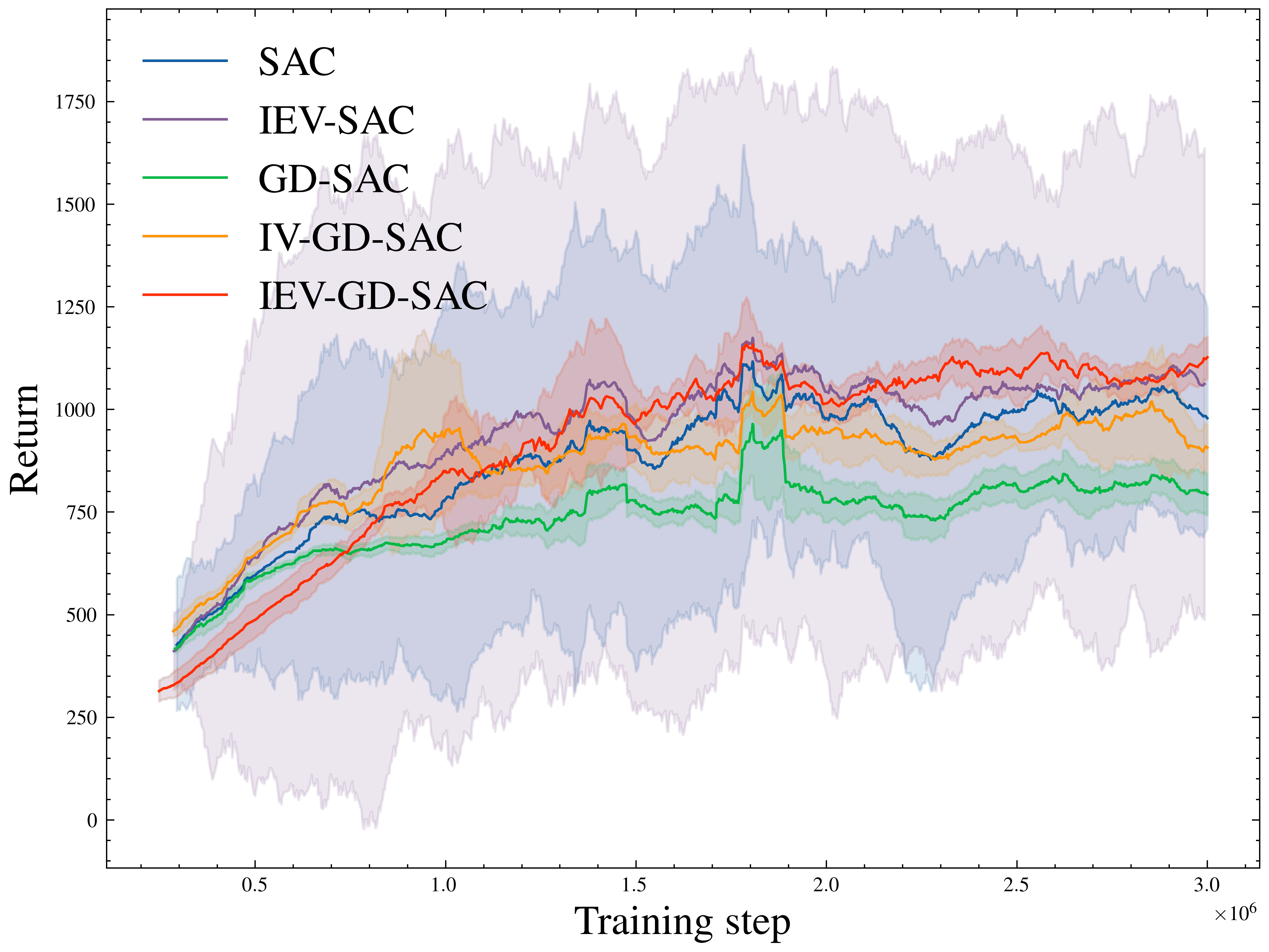}}
	\caption{
		Ablation study on BIEV regularization.
	} \label{fig:abl-biev}
\end{figure}

\begin{figure}[t]
	\centering
	\subfloat[Ant-v4]{\includegraphics[width=.31\textwidth]{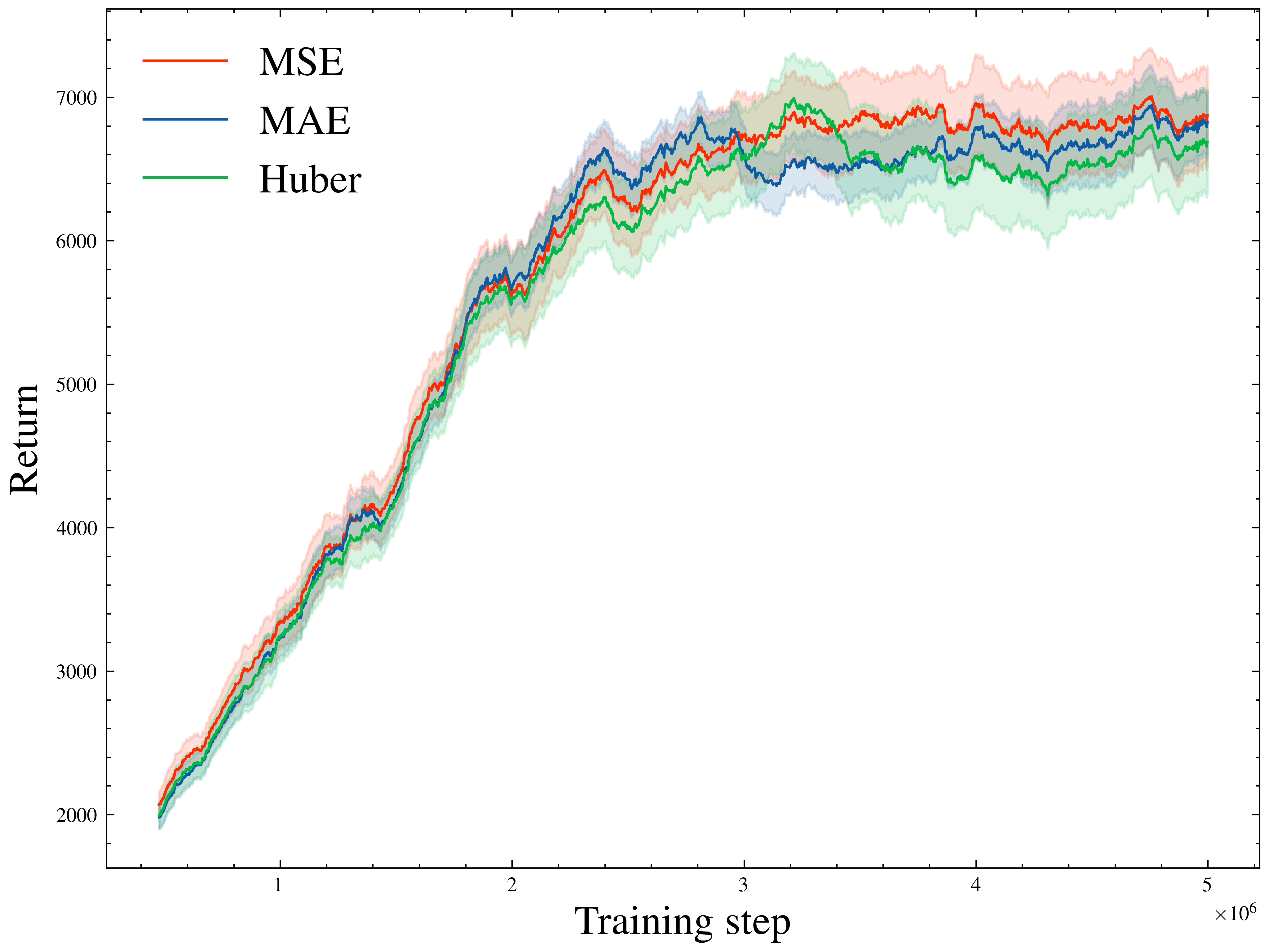}}
	\subfloat[HalfCheetah-v4]{\includegraphics[width=.31\textwidth]{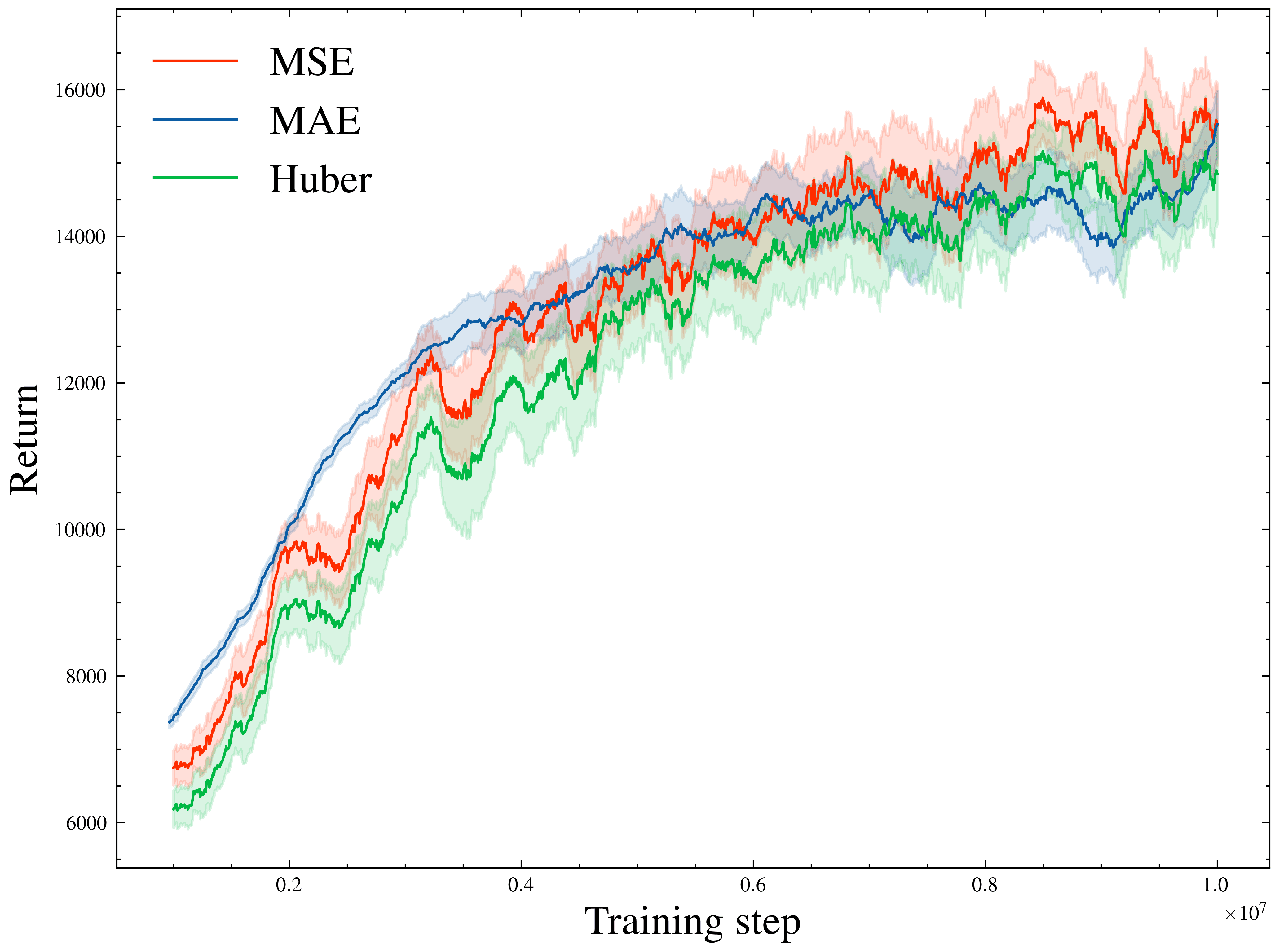}}
	\subfloat[Hopper-v4]{\includegraphics[width=.31\textwidth]{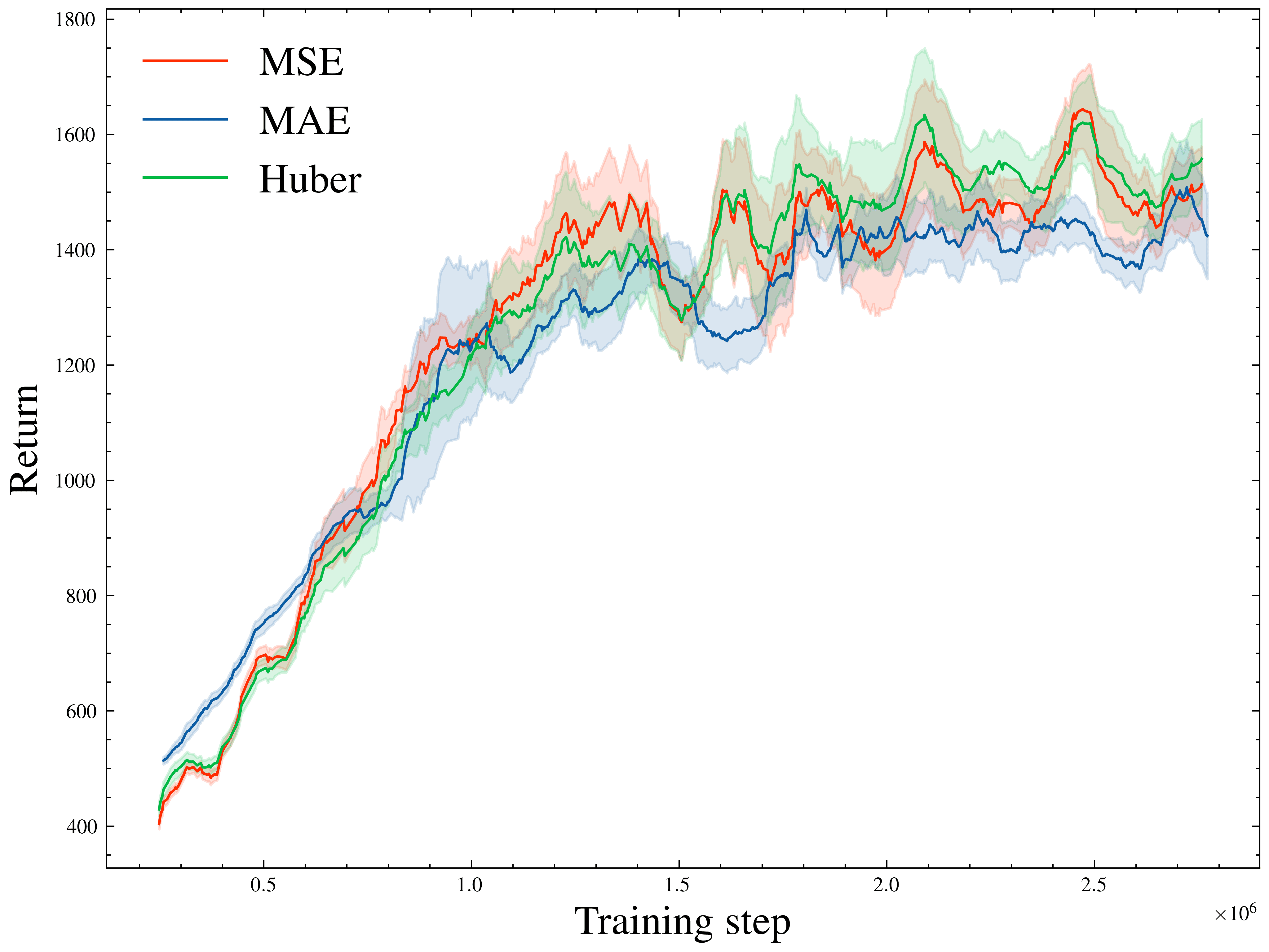}}
	\caption{
		Ablation study on loss selection for BIEV regularization.
	} \label{fig:abl-biev-loss}
\end{figure}

\begin{figure}[t]
	\centering
	\subfloat[Ant-v4]{\includegraphics[width=.31\textwidth]{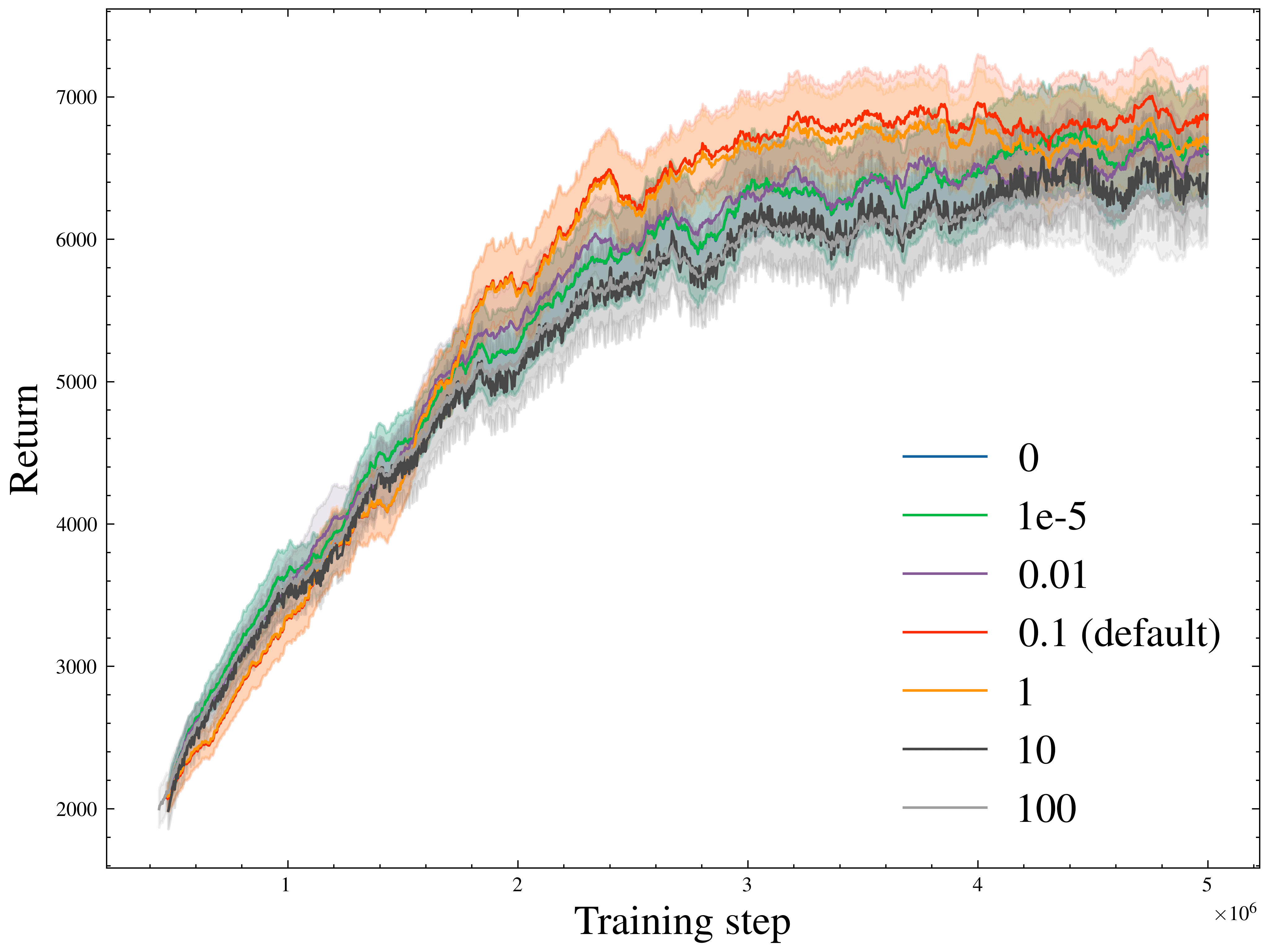}}
	\subfloat[HalfCheetah-v4]{\includegraphics[width=.31\textwidth]{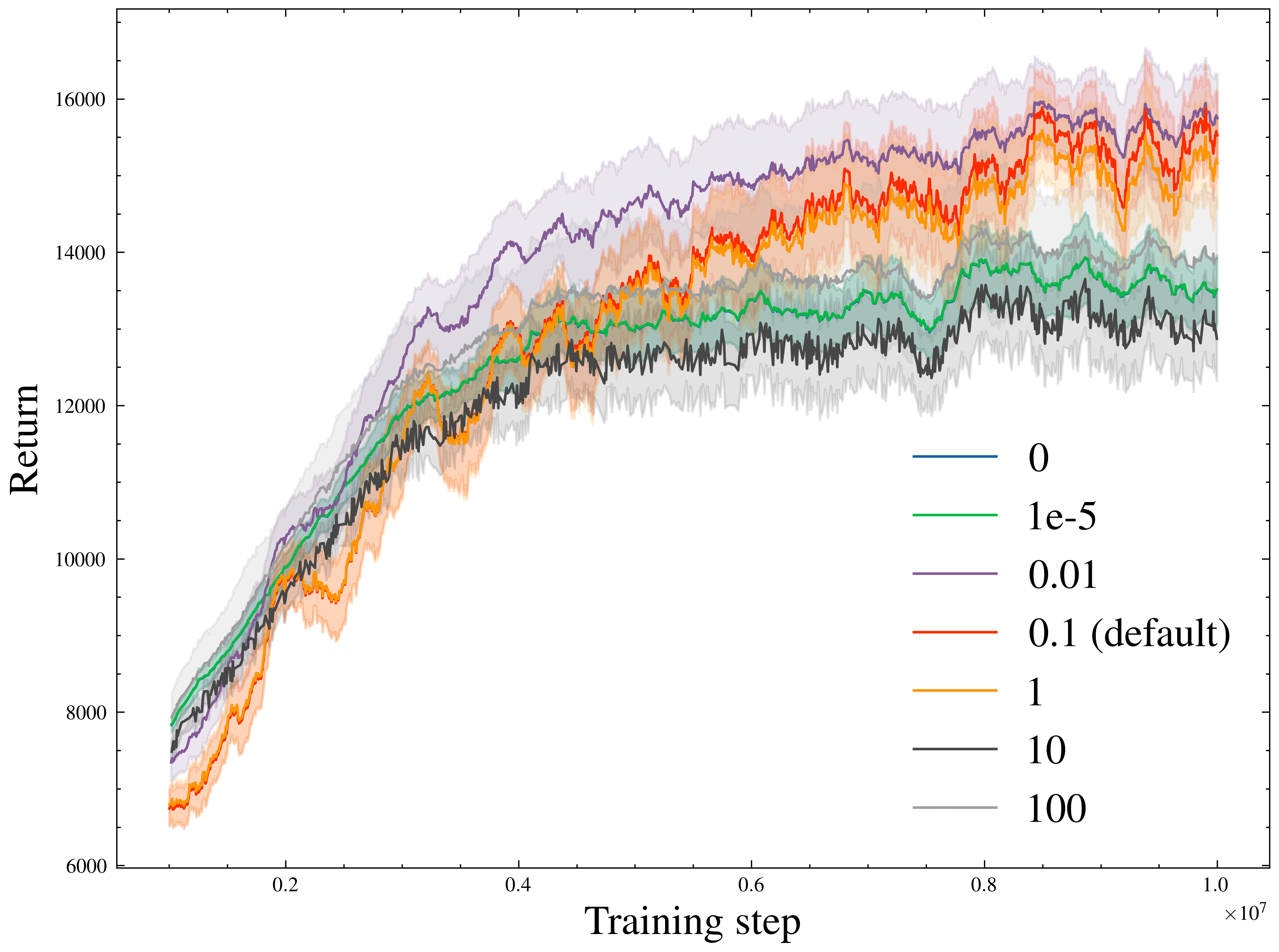}}
	\subfloat[Hopper-v4]{\includegraphics[width=.31\textwidth]{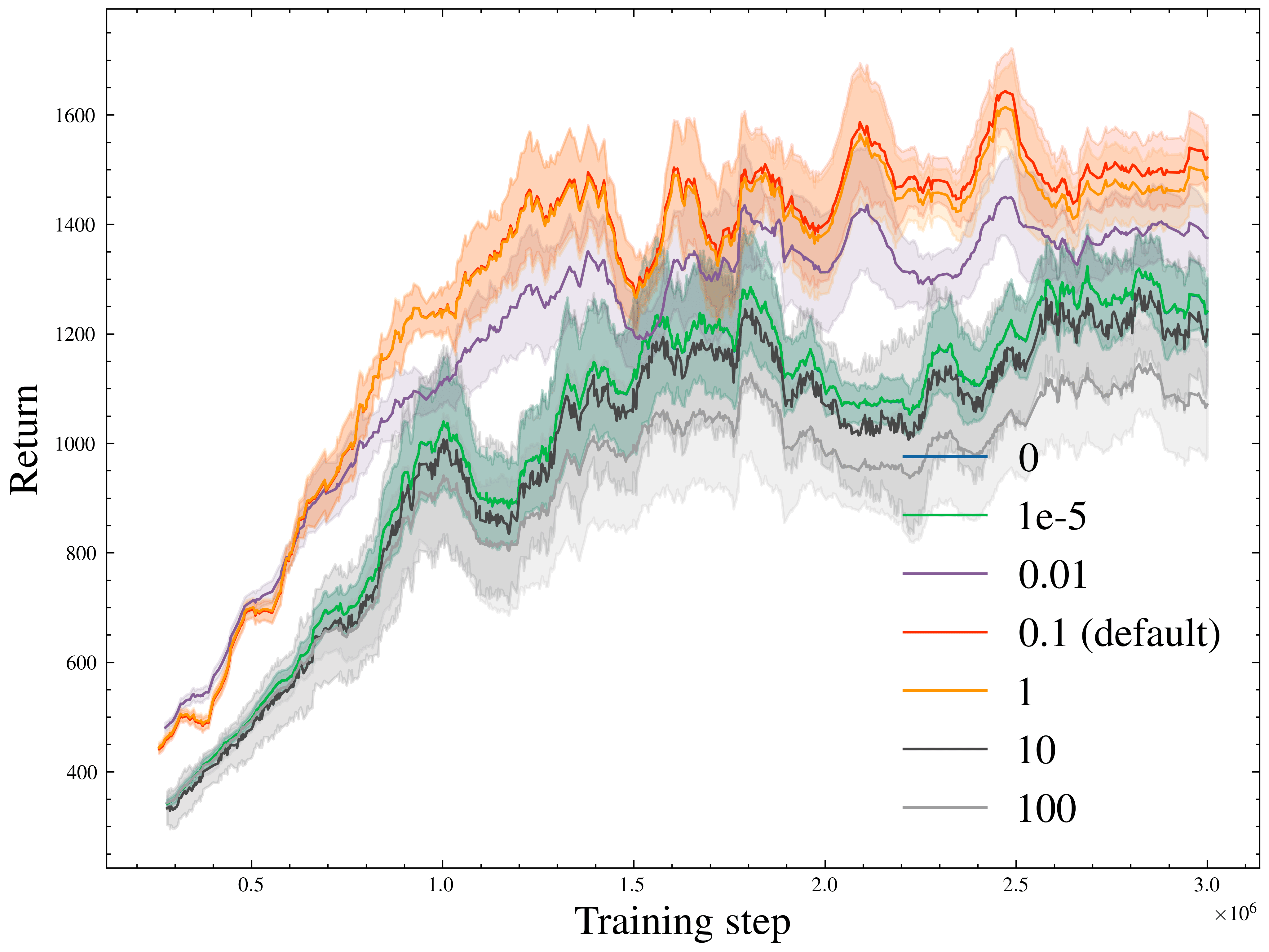}}
	\caption{
		Ablation study on regularizing temperature $\lambda$.
	} \label{fig:abl-temp}
\end{figure}

\begin{figure}[t]
	\centering
	\subfloat[Ant-v4]{\includegraphics[width=.31\textwidth]{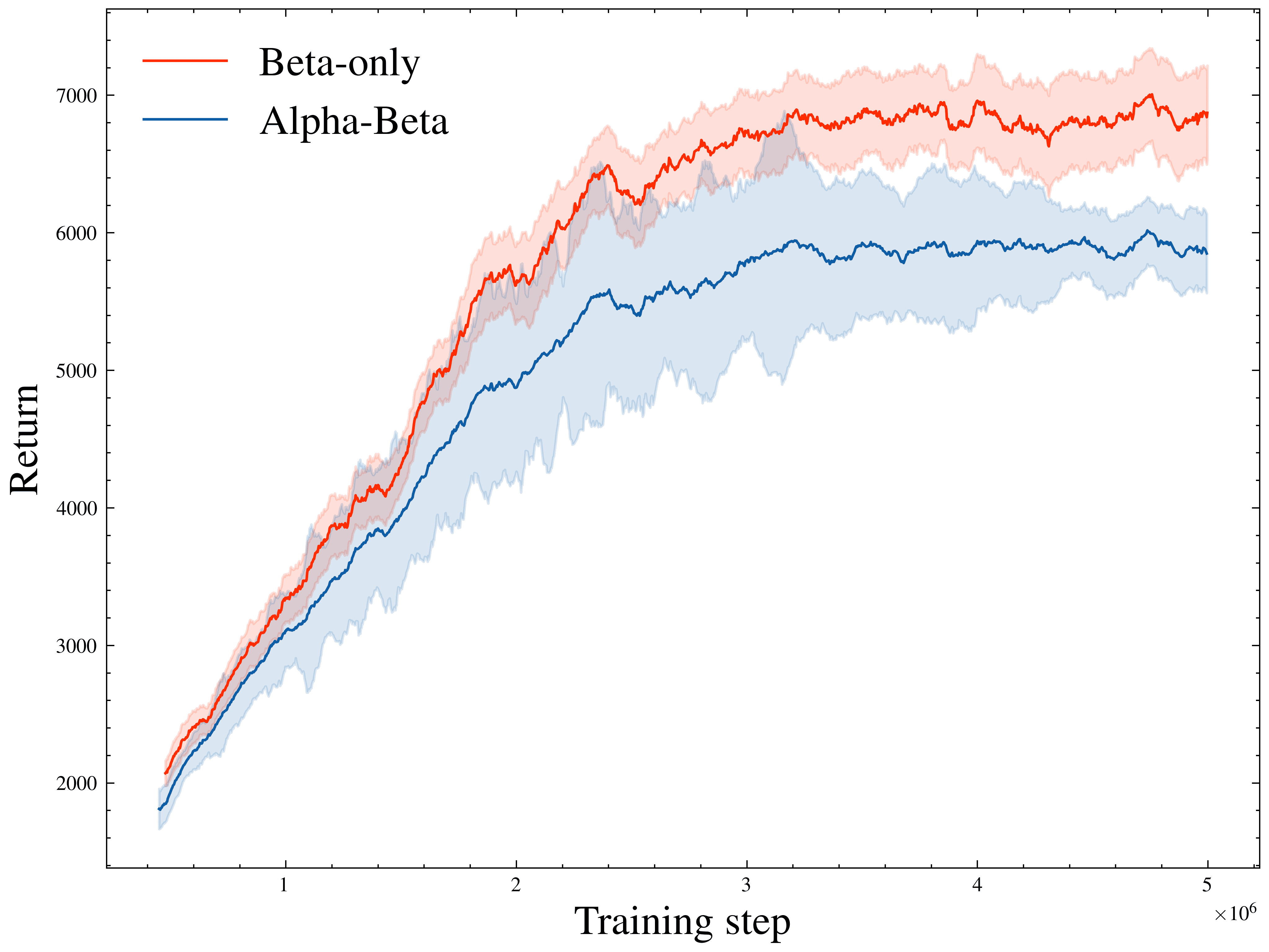}}
	\subfloat[HalfCheetah-v4]{\includegraphics[width=.31\textwidth]{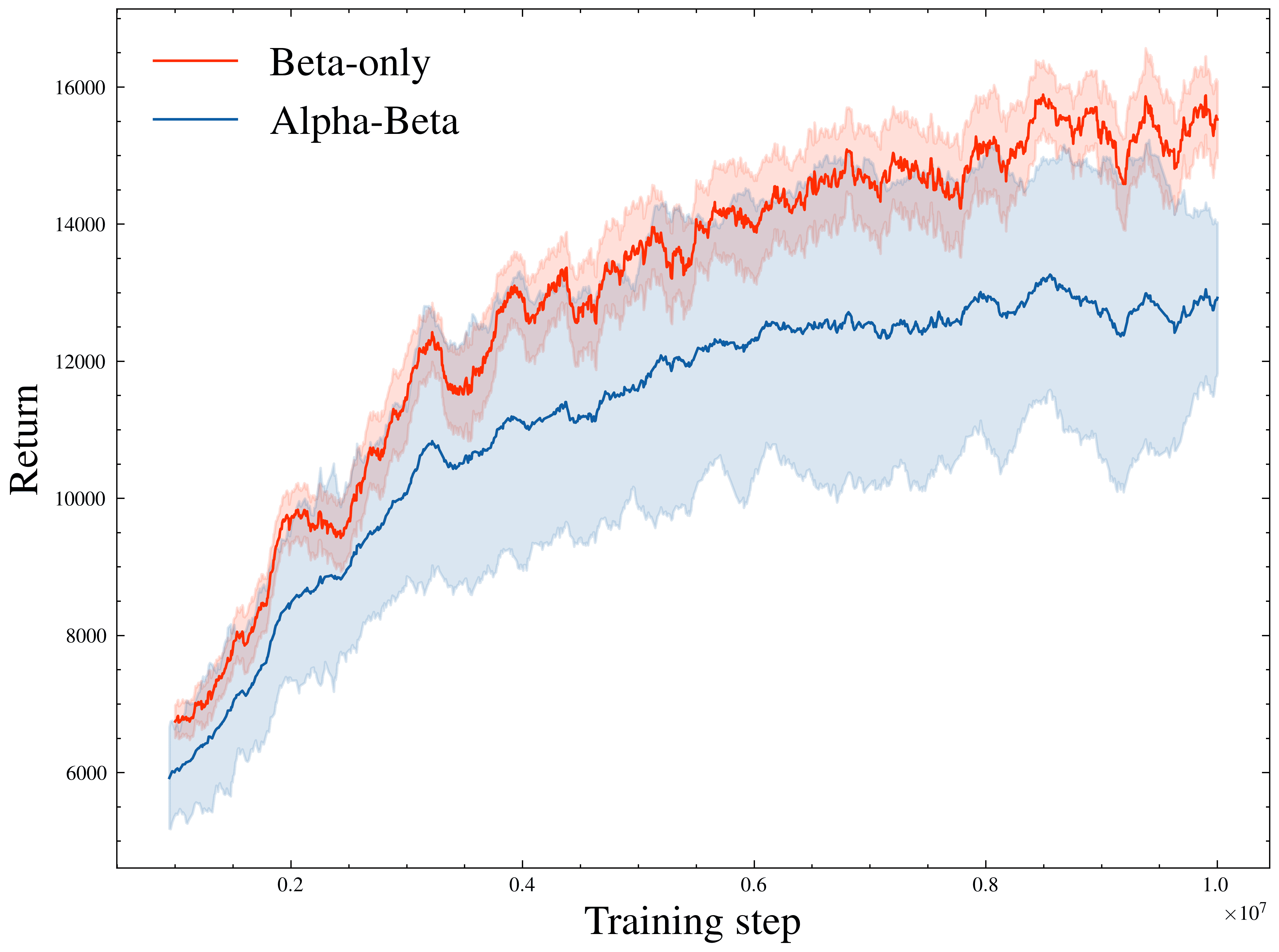}}
	\subfloat[Hopper-v4]{\includegraphics[width=.31\textwidth]{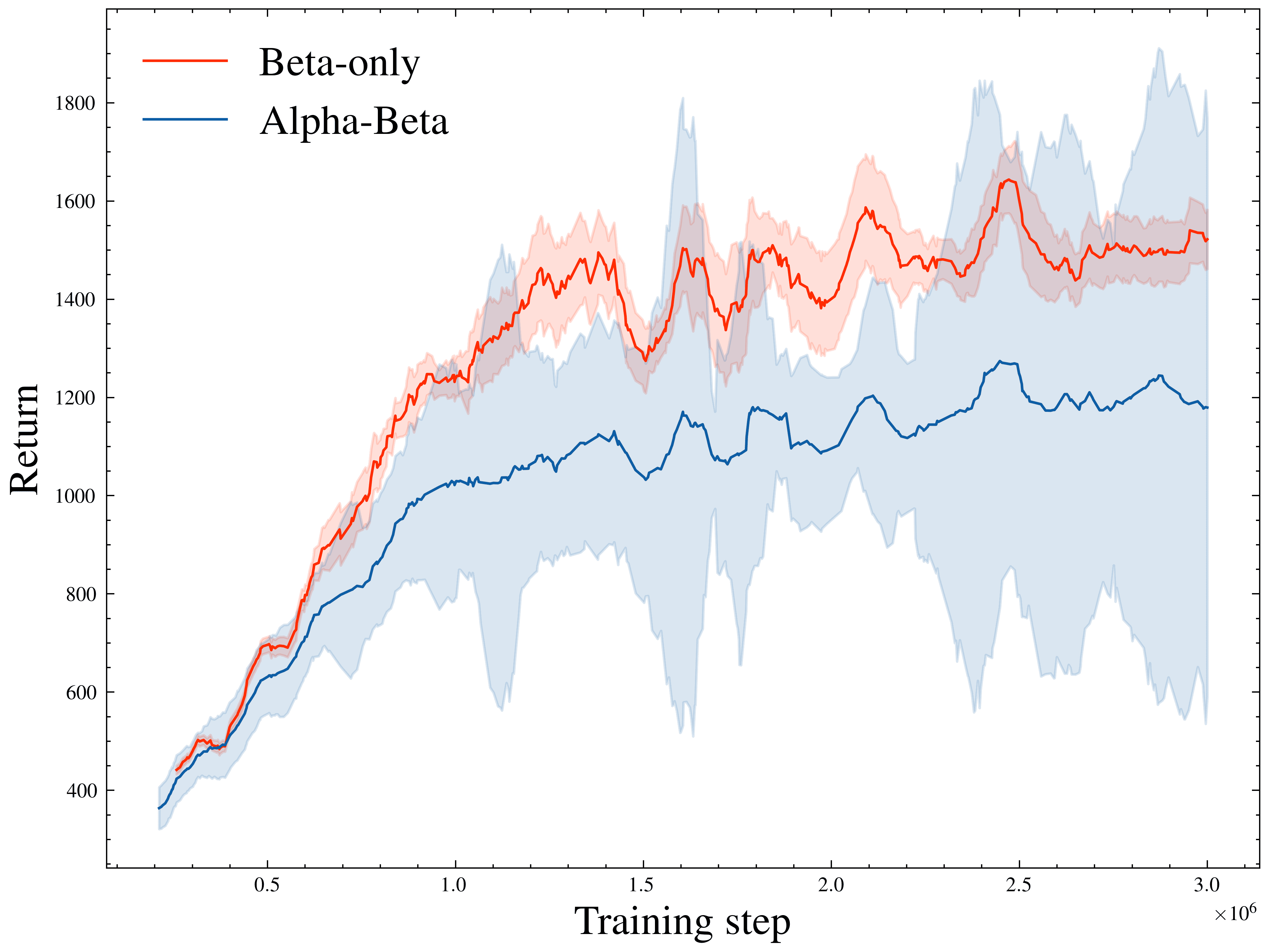}}
	\caption{
		Ablation study on alpha head.
	} \label{fig:abl-alpha}
\end{figure}

\begin{figure}[t]
	\centering
	\subfloat[120]{\includegraphics[width=.23\textwidth]{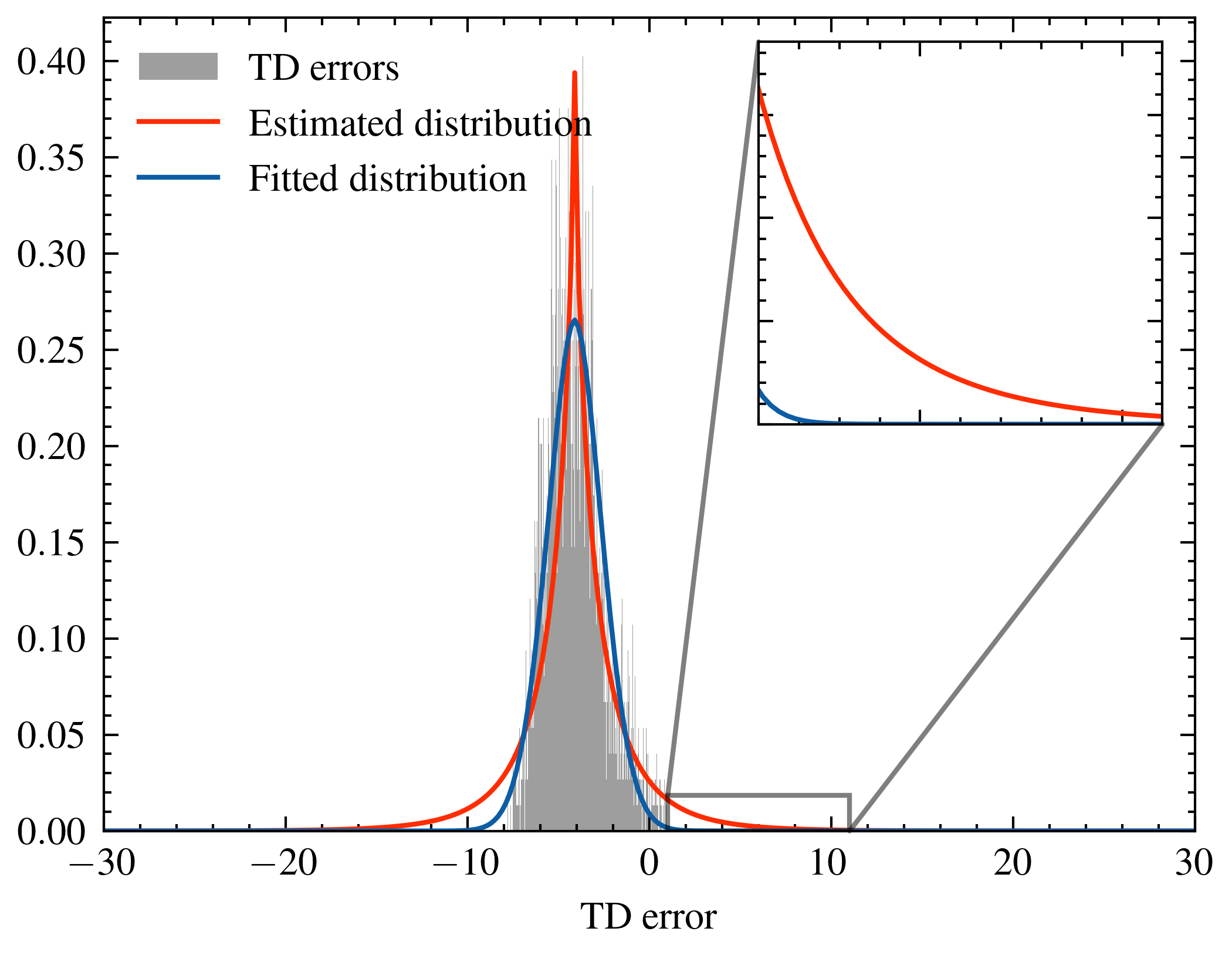}}
	\subfloat[1K]{\includegraphics[width=.23\textwidth]{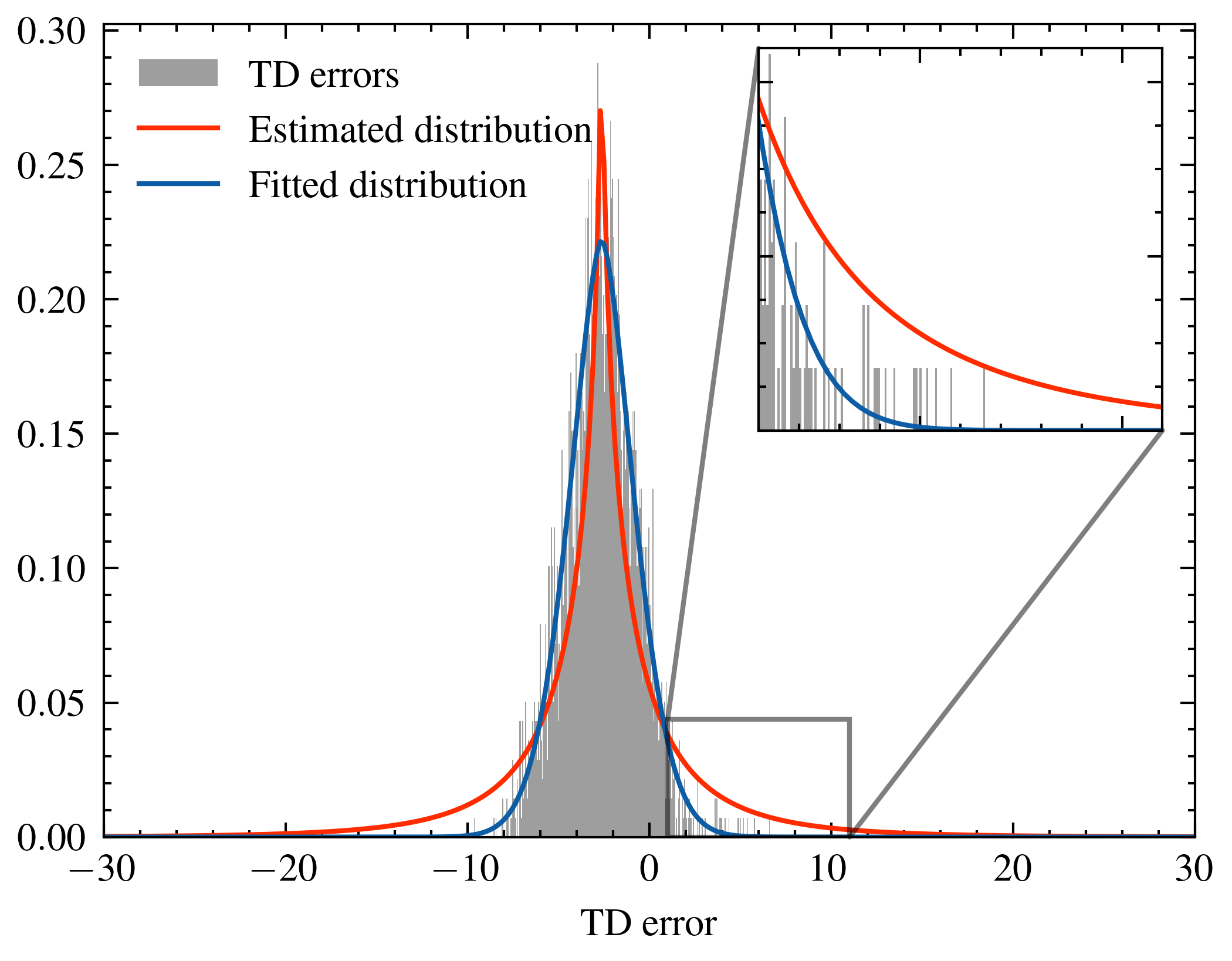}}
	\subfloat[100K]{\includegraphics[width=.23\textwidth]{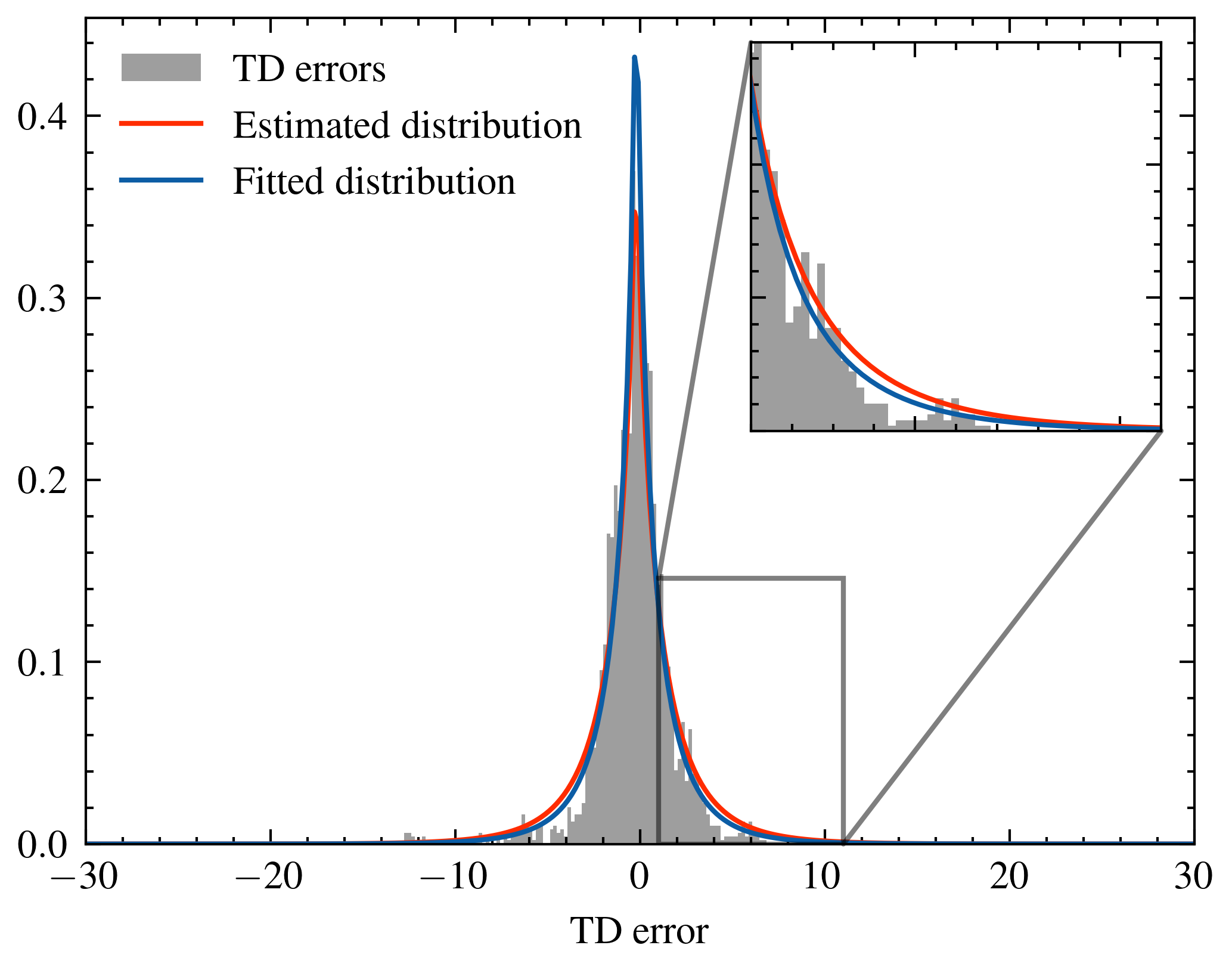}}
	\subfloat[5M]{\includegraphics[width=.23\textwidth]{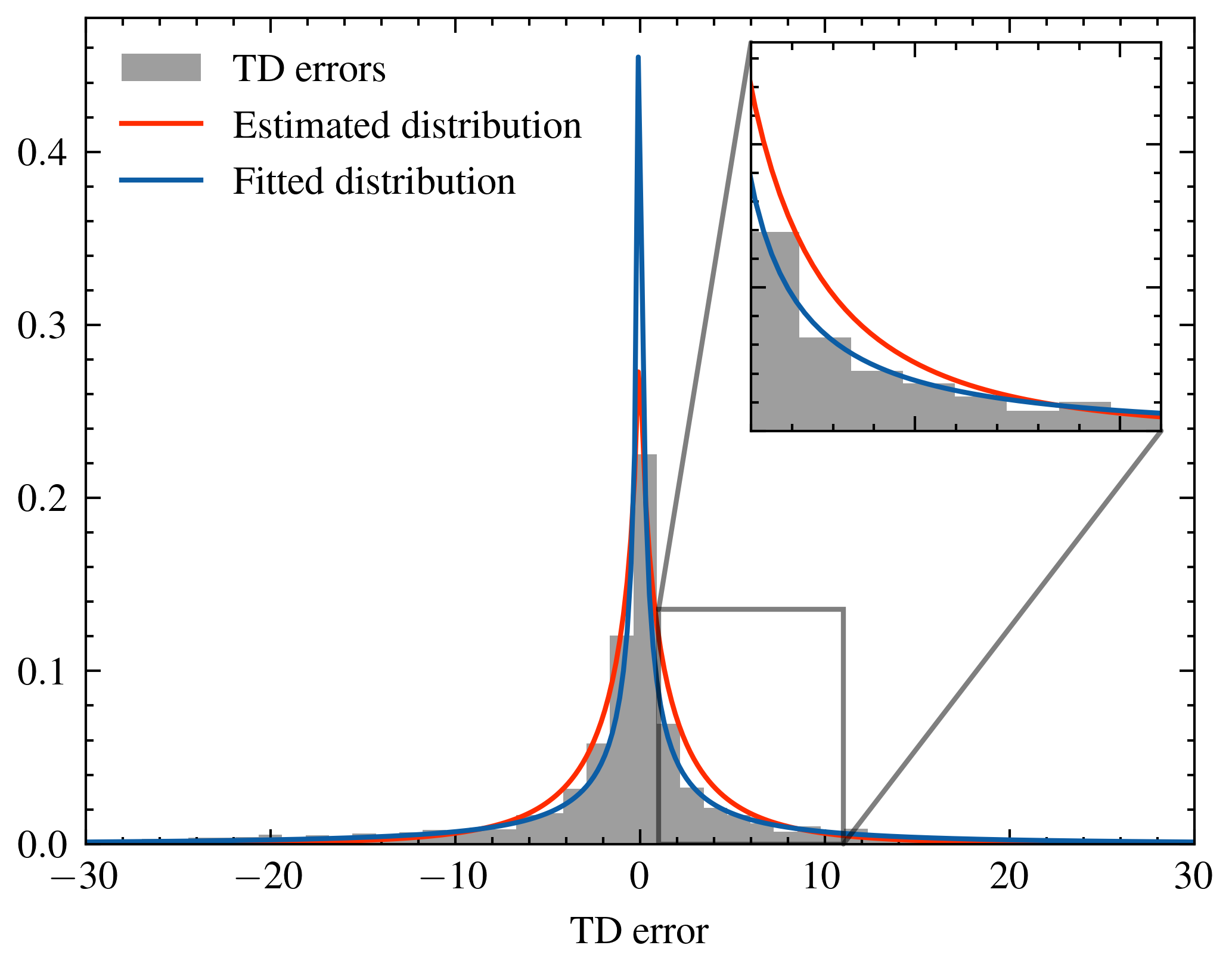}}
	\caption{
		TD error plots of SAC on the Ant-v4 environment at different training steps.
	} \label{fig:abl-beta-sufficiency}
\end{figure}

We conduct a series of ablation studies using the SAC algorithm across selected MuJoCo environments.


\subsection{On Risk-Averse Weighting} \label{apdx:abl:raw}


\cref{thm:ssd} motivates, but does not derive, a monotone preference in $\beta$ under our fixed-scale surrogate.
\cref{fig:abl-raw} compares the implemented normalized $\beta$ weighting with the unweighted surrogate and an inverse-$\beta$ negative control.
The chosen direction improves early learning in the plotted environments but does not consistently improve asymptotic performance.
The inverse control produces periodic non-monotone jumps; we did not measure an exploration statistic that would support a causal interpretation.


\subsection{On BIEV Regularization} \label{apdx:abl:biev}


The BIEV term is independent of the $\beta$ head: it uses variance and sample excess kurtosis across the critics' TD errors rather than a moment implied by the GGD parameterization.
As depicted in~\cref{fig:abl-biev}, BIEV is close to BIV in sample efficiency and improves some asymptotic curves, without uniform dominance.


The reported GGD regularized variants use absolute TD error.
\cref{fig:abl-biev-loss} compares that choice with squared error; differences are generally modest in these environments, and the MAE form is less sensitive to $\lambda$ in the plotted ablation (\cref{apdx:abl:lambda}).


\subsection{On Regularizing Temperature} \label{apdx:abl:lambda}


The loss function presented in~\cref{eq:loss} is influenced by the scale of the rewards, requiring the regularizing temperature to be adjusted accordingly for different environments.
As shown in~\cref{fig:abl-temp}, the regularizing temperature $\lambda$ affects sample efficiency, training stability, and asymptotic performance.
However, this sensitivity is notable only with large differences in scale, such as between 0.01 and 100.
Furthermore, as the regularizing temperature approaches zero, the model's performance converges to that of the GGD error modeling scheme without BIEV regularization, labeled as `0', as expected.


\subsection{On Alpha Head} \label{apdx:abl:alpha}


We use only the $\beta$ head as the one-parameter design discussed in~\cref{rem:beta-head}.
Incorporating the alpha head diminishes sample efficiency and leads to decreased asymptotic performance, particularly in Ant-v4, as shown in~\cref{fig:abl-alpha}.


To inspect this restriction, we compare pooled empirical TD errors with visualization-only GGDs using the model's aggregate $\beta$ estimates and with SciPy fits that optimize both $\alpha$ and $\beta$.
\cref{fig:abl-beta-sufficiency} shows that the one-parameter curves capture part of the pooled shape in Ant-v4, but this marginal visualization does not establish per-transition sufficiency or identifiability.
Together with~\cref{fig:abl-alpha}, it supports the reported engineering choice rather than a general statistical claim.


\section{Extension} \label{apdx:qlearn}

\begin{figure}[H]
	\centering
	\subfloat[CartPole-v1]{\includegraphics[width=.65\linewidth]{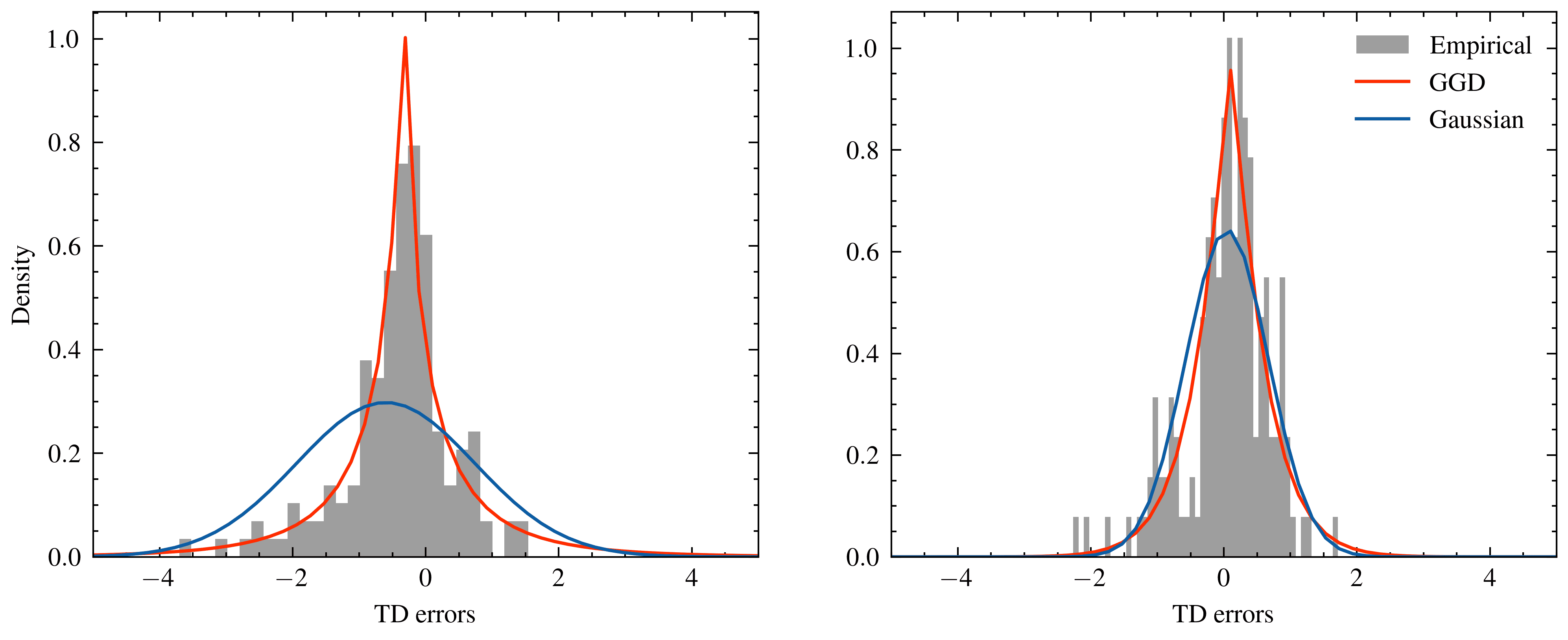}} \\
	\subfloat[MountainCar-v0]{\includegraphics[width=.65\textwidth]{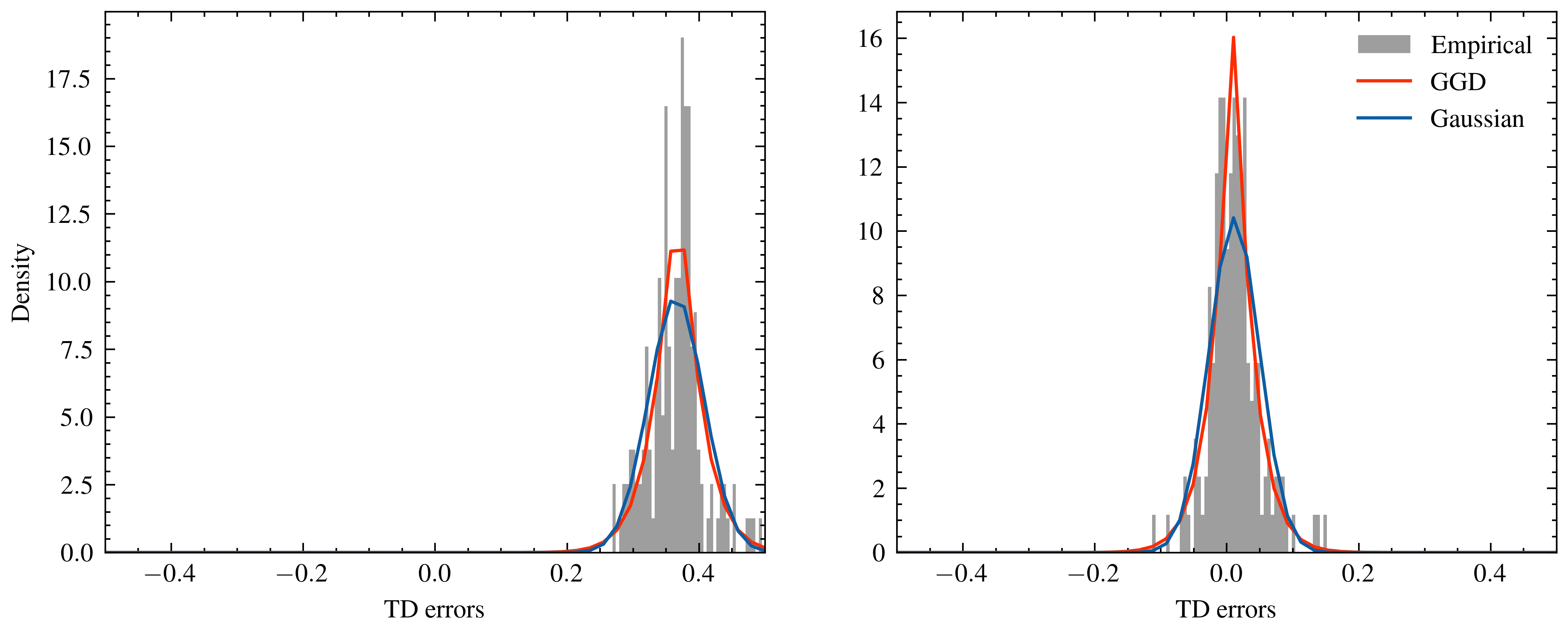}}
	\caption{
		TD error plots of DQN.
	} \label{fig:tde-dqn}
\end{figure}

As discussed in~\cref{sec:disc}, we expand the empirical analysis to $Q$-learning using DQN~\citep{mnih2015human}.
The pooled errors in~\cref{fig:tde-dqn} deviate visibly from the Gaussian overlays and are descriptively closer to the fitted GGD curves in these environments.
This extension separates the transferable GGD-inspired critic surrogate from the shape-weighting rule designed for stochastic policy-gradient algorithms with separate actors.

\begin{figure}[H]
	\centering
	\subfloat[CartPole-v1]{\includegraphics[width=.31\linewidth]{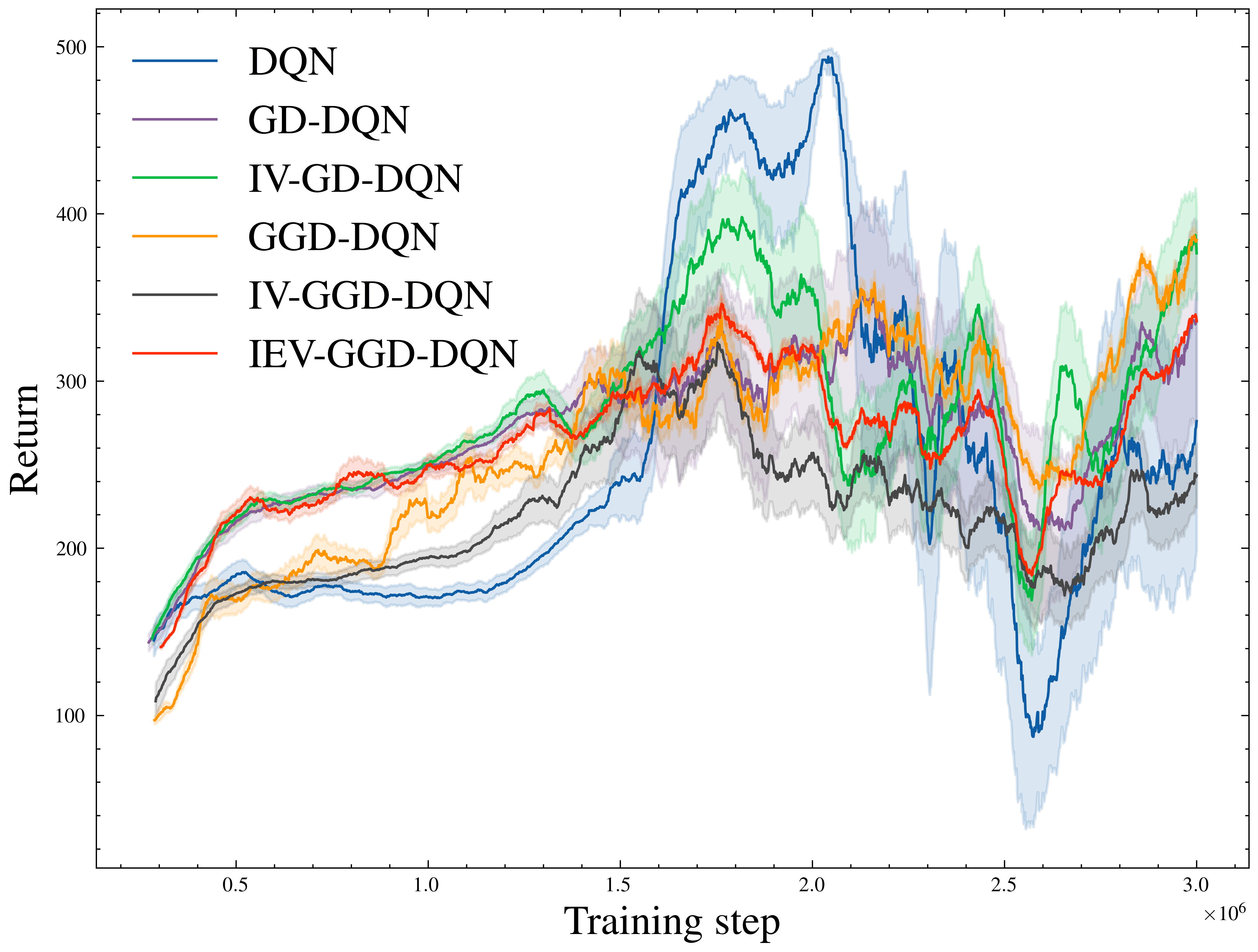}}
	\subfloat[LunarLander-v2]{\includegraphics[width=.31\textwidth]{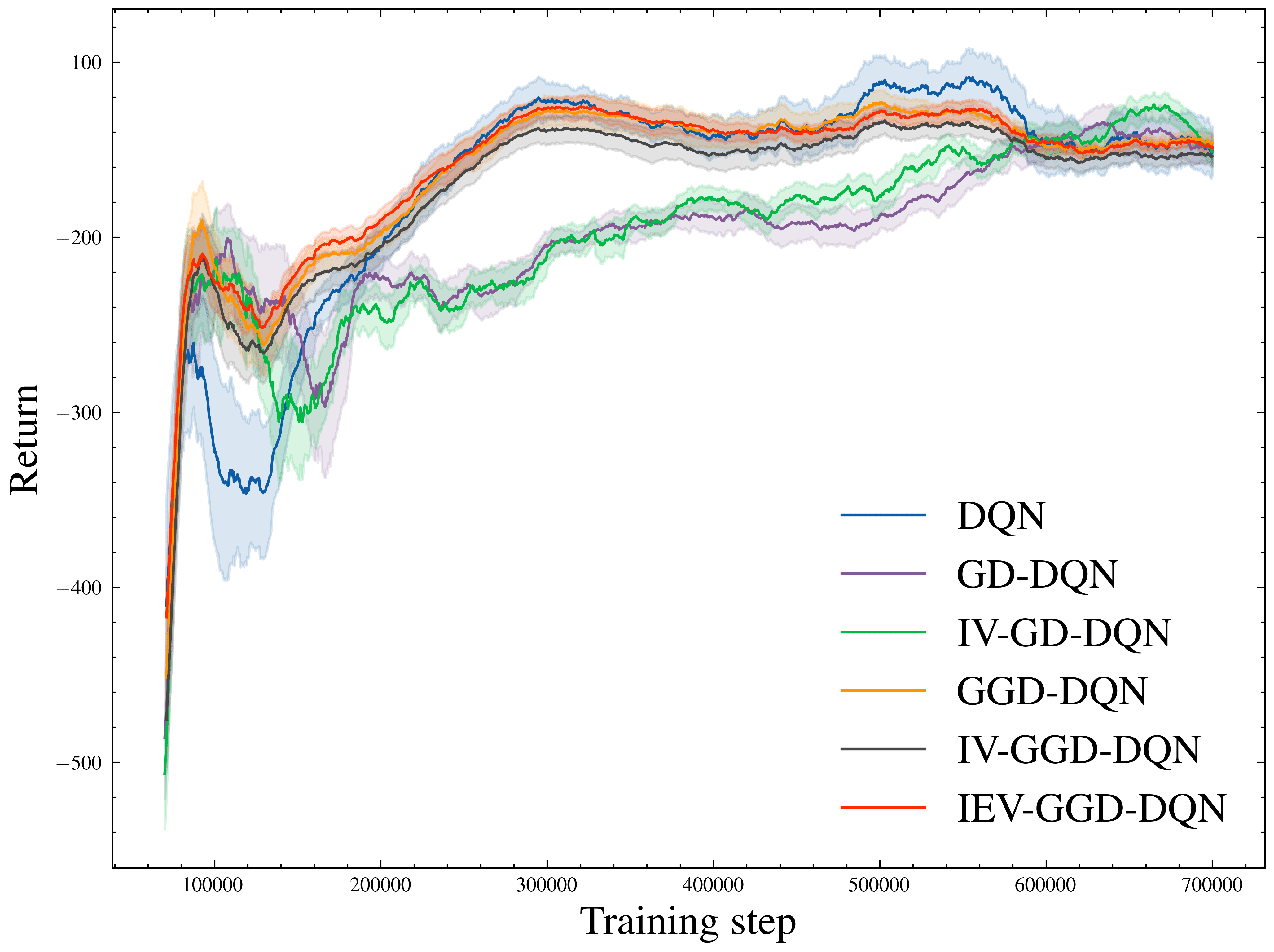}}
	\subfloat[MountainCar-v0]{\includegraphics[width=.31\textwidth]{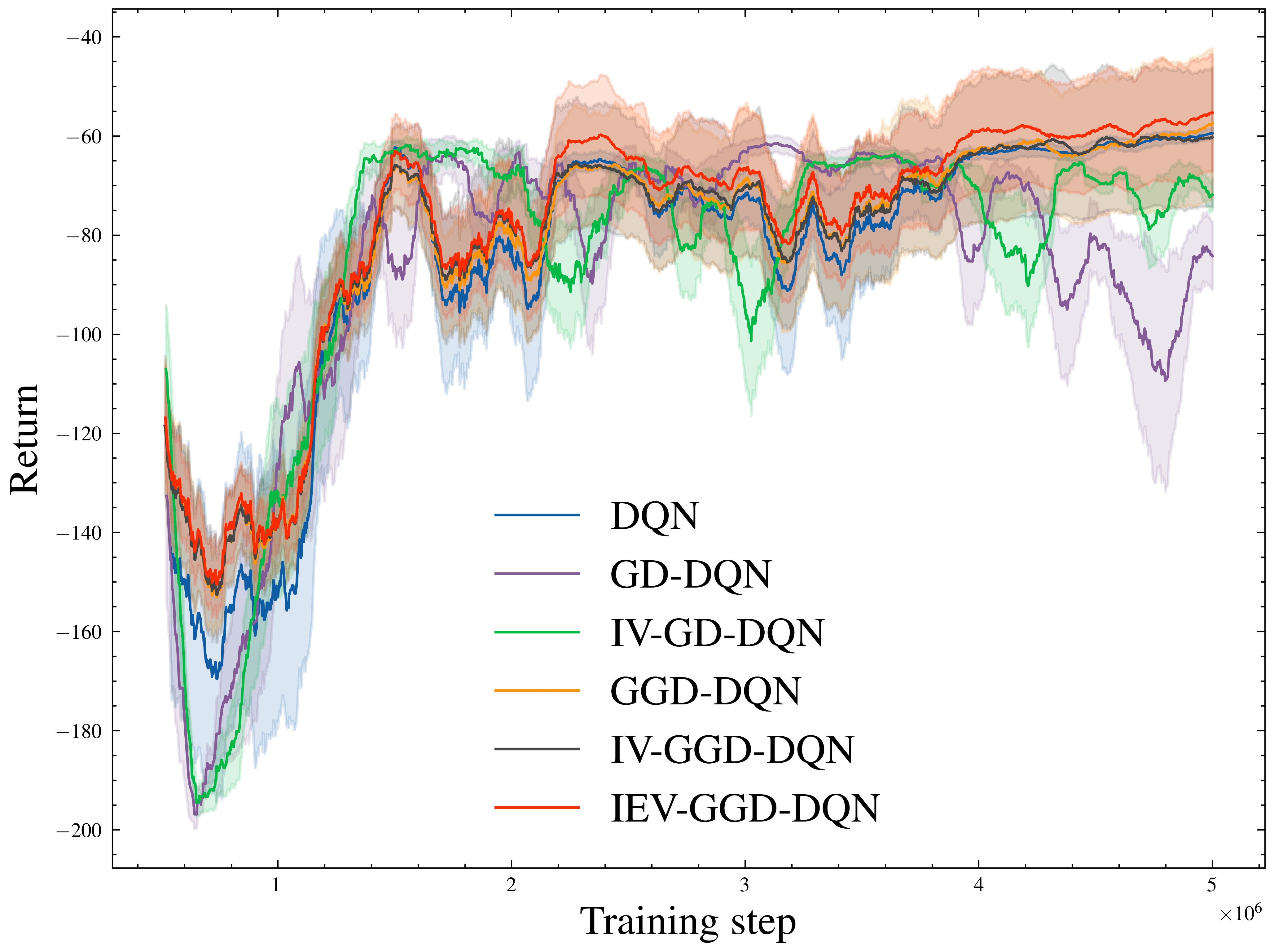}}
	\caption{
		Sample efficiency curves of DQN on noisy discrete control environments.
	} \label{fig:dqn}
\end{figure}

\begin{figure}[H]
	\centering
	\subfloat[CartPole-v1]{\includegraphics[width=.31\linewidth]{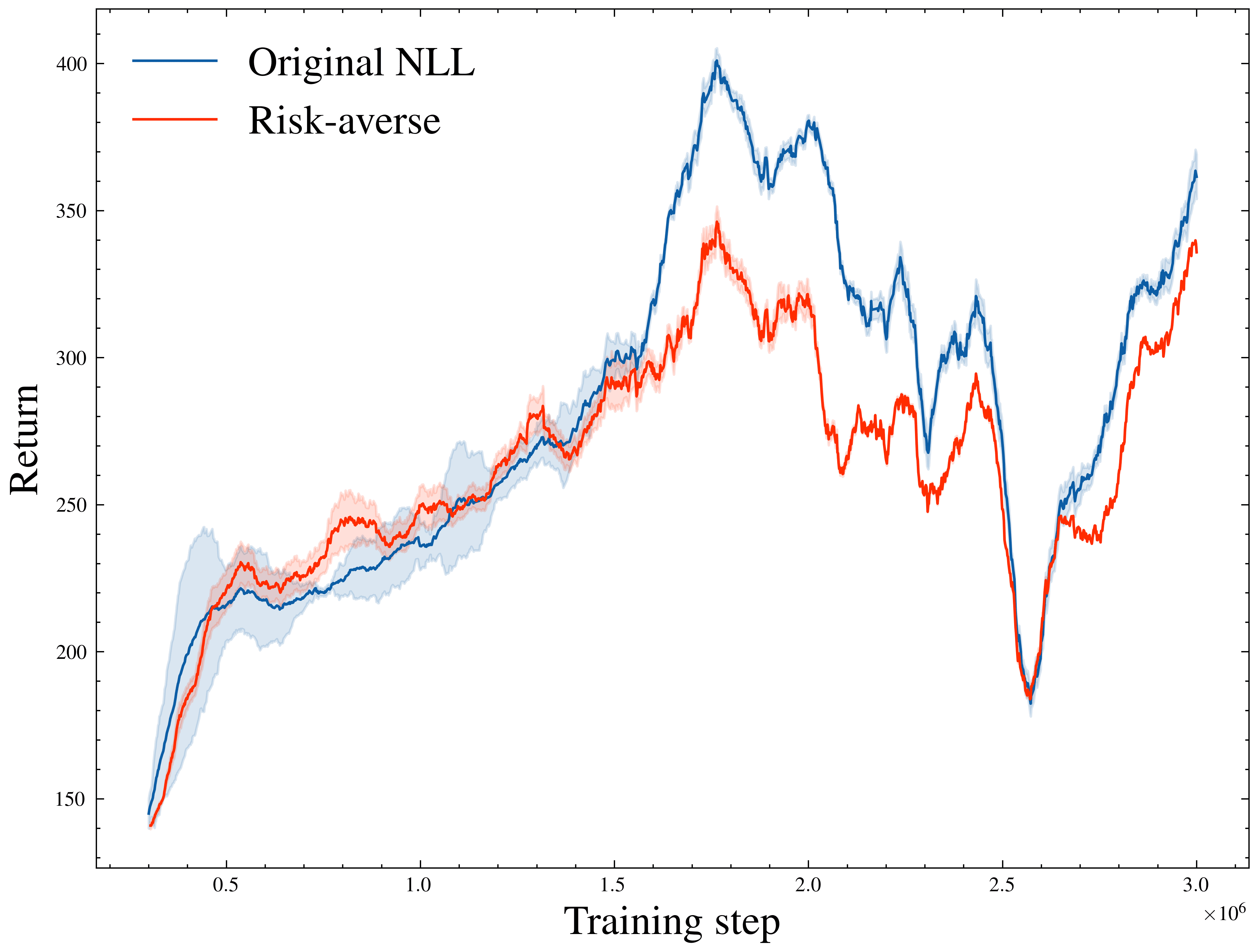}}
	\subfloat[LunarLander-v2]{\includegraphics[width=.31\textwidth]{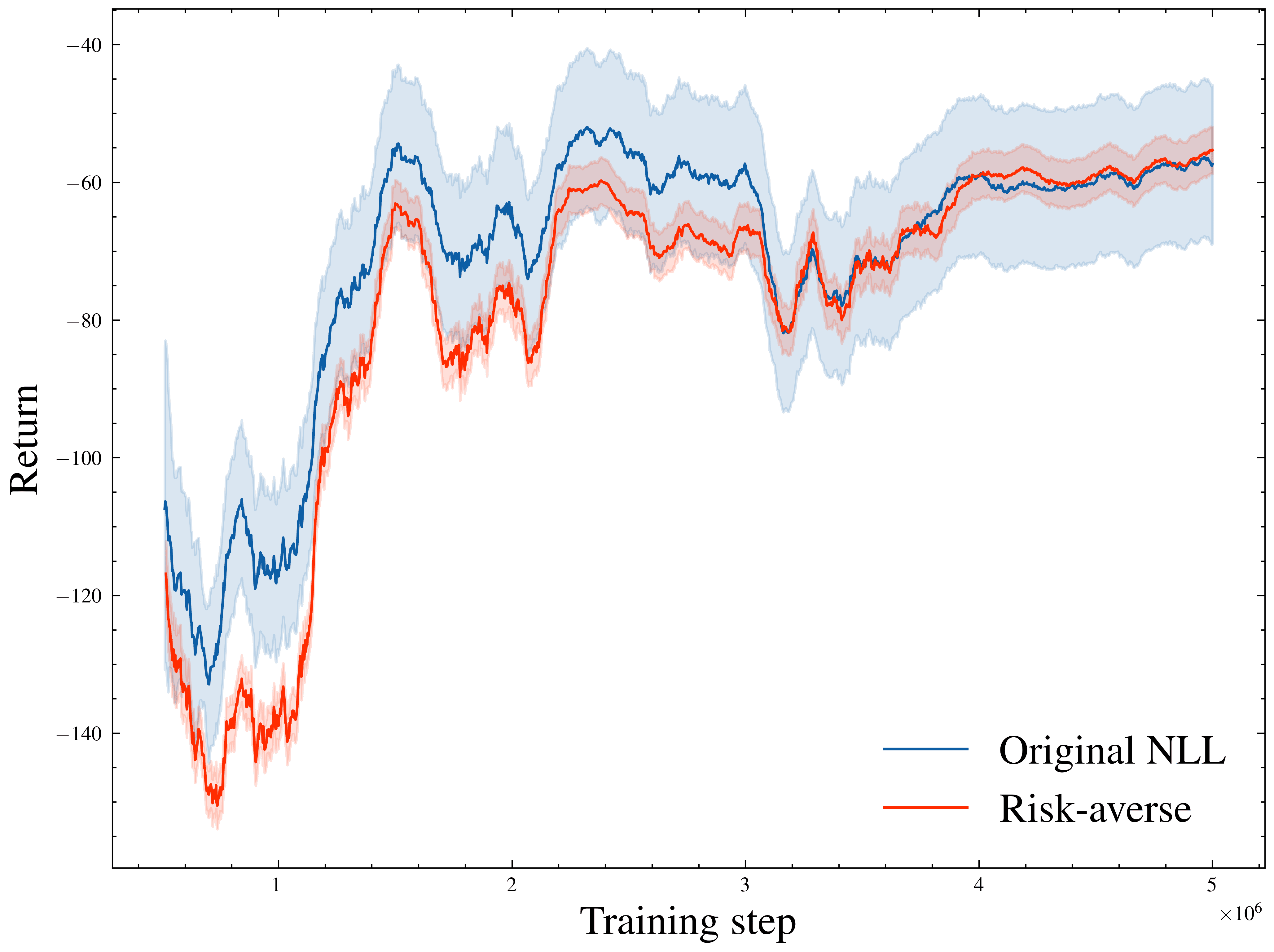}}
	\subfloat[MountainCar-v0]{\includegraphics[width=.31\textwidth]{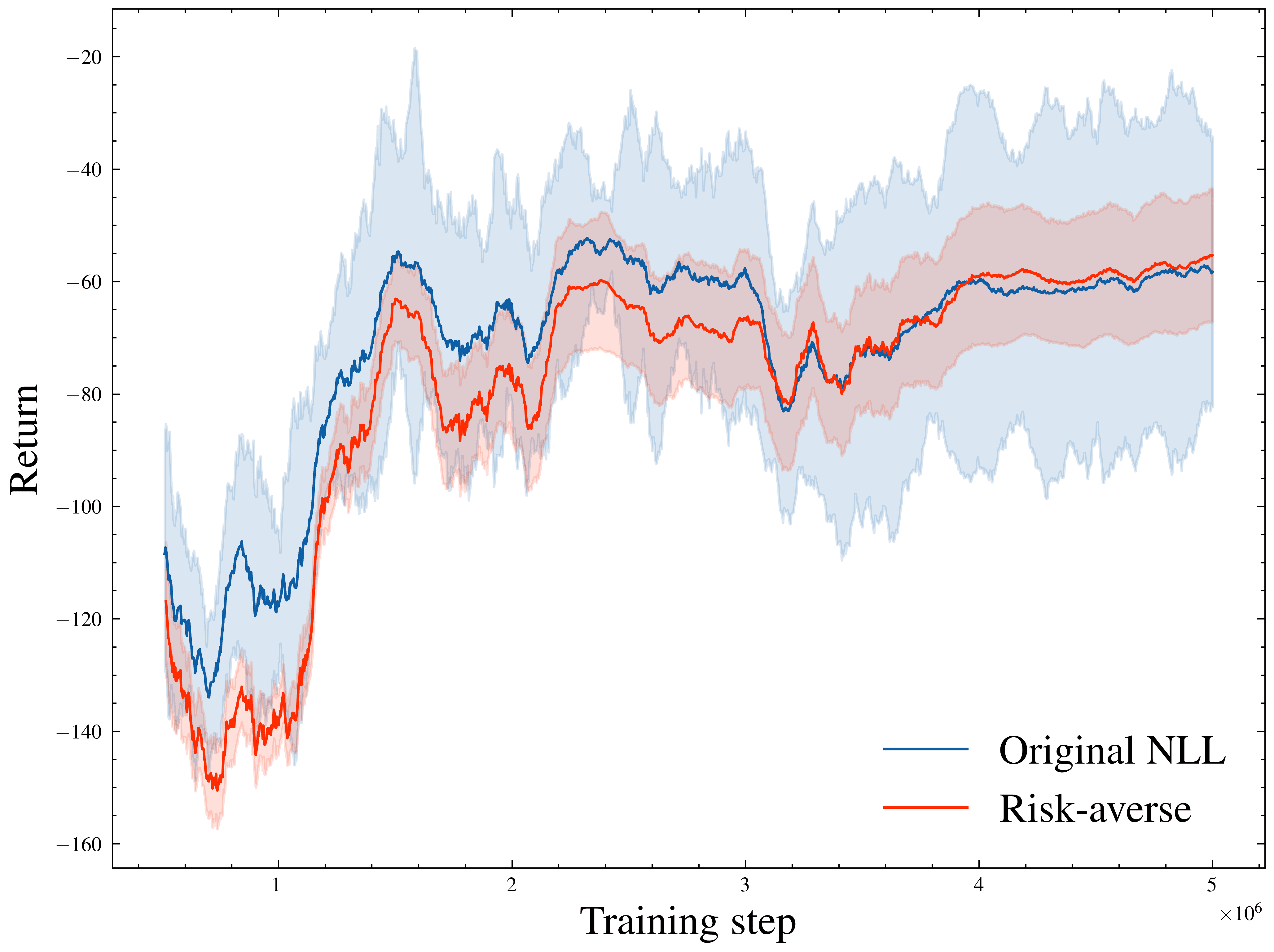}}
	\caption{
		Ablation study on risk-averse weighting for DQN.
	} \label{fig:dqn-weighting}
\end{figure}

The reported performance in~\cref{fig:dqn} does not show the same improvement from direct $\beta$ weighting as the policy-gradient experiments, and the regularized curves are close in several settings.
One plausible explanation is that critic reweighting interacts directly with action selection in the greedy DQN setup, whereas SAC and PPO use separate stochastic actors.
The experiments do not isolate this mechanism, so we treat it as a scope hypothesis rather than a confirmed causal account.
The contrast in~\cref{fig:dqn-weighting}, where the unweighted surrogate is slightly more sample-efficient than the $\beta$-weighted variant, motivates restricting our claim about the weighting rule to the evaluated policy-gradient setting.

\end{document}